\documentclass[11pt]{article}

\usepackage[final]{acl}

\usepackage{times,latexsym}
\usepackage[most]{tcolorbox}
\usepackage{etoc}
\usepackage{subcaption}
\usepackage{wrapfig}
\usepackage{enumitem}
\usepackage{cuted}
\usepackage[T1]{fontenc}

\usepackage[utf8]{inputenc}
\usepackage{microtype,inconsolata}
\usepackage{multirow}
\usepackage{graphicx,booktabs}
\usepackage[abs]{overpic}
\usepackage{hyperref, amsfonts, amsmath, enumitem}
\usepackage[capitalise]{cleveref}
\usepackage{dashrule}

\usepackage{url}

\newtcbox{\inlinebox}[1][]{
 box align=base,
 nobeforeafter,
 colback=white,
 colframe=yellow,
 size=small,
 left=0pt,
 right=0pt,
 boxsep=2pt,
 arc=0pt,outer arc=0pt
 #1}
 
\title{Understanding the Effects of Distractors on \\Reasoning Vision-Language Models}

\author{Jiyun Bae \quad Hyunjong Ok \quad Sangwoo Mo \quad Jaeho Lee\\
Pohang University of Science and Technology (POSTECH)\\
\tt\small \{jiyun.bae,hyunjong.ok,sangwoo.mo,jaeho.lee\}@postech.ac.kr}

\begin{document}
\maketitle
\etocdepthtag.toc{mtmain}
\begin{abstract} \label{sec:abstract}


How does irrelevant information (i.e., distractors) affect test-time scaling in vision-language models (VLMs)? Prior work on text-only language models has shown that textual distractors can intensify inverse scaling, causing models to reason longer but less effective reasoning traces. In this work, we investigate whether similar phenomena arise in multimodal settings. We introduce Idis (Images with distractors), a visual question-answering dataset that systematically varies distractors along semantic and numerical dimensions. Our analyses reveal that visual distractors affect reasoning VLMs in a fundamentally different way from textual distractors: although inverse scaling still emerges, visual distractors reduce accuracy without increasing reasoning length. We further show that attribute counts extracted from reasoning traces provide key insights into how distractors interact with reasoning length and accuracy. As a sanity check, we propose a simple prompting strategy that mitigates distractor-driven predictions in reasoning vision-language models.\footnote{Code: \url{https://github.com/effl-lab/Idis}.}
\vspace{-4mm}
\end{abstract}

\begin{figure}[t]
\centering
\begin{minipage}[c]{0.04\linewidth}
  \raggedleft (a)
  \label{fig:intro_a}
\end{minipage}%
\hspace{1mm}%
\begin{minipage}[c]{0.63\linewidth}
  \centering
  \includegraphics[width=\linewidth,trim=40 0 60 5,clip]%
    {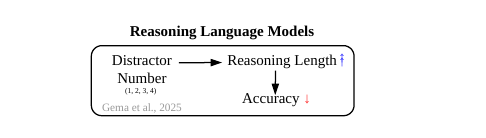}
\end{minipage}%
\hspace{1mm}%
\begin{minipage}[c]{0.23\linewidth}
  \centering
  \includegraphics[width=\linewidth,trim=10 3 10 0,clip]%
    {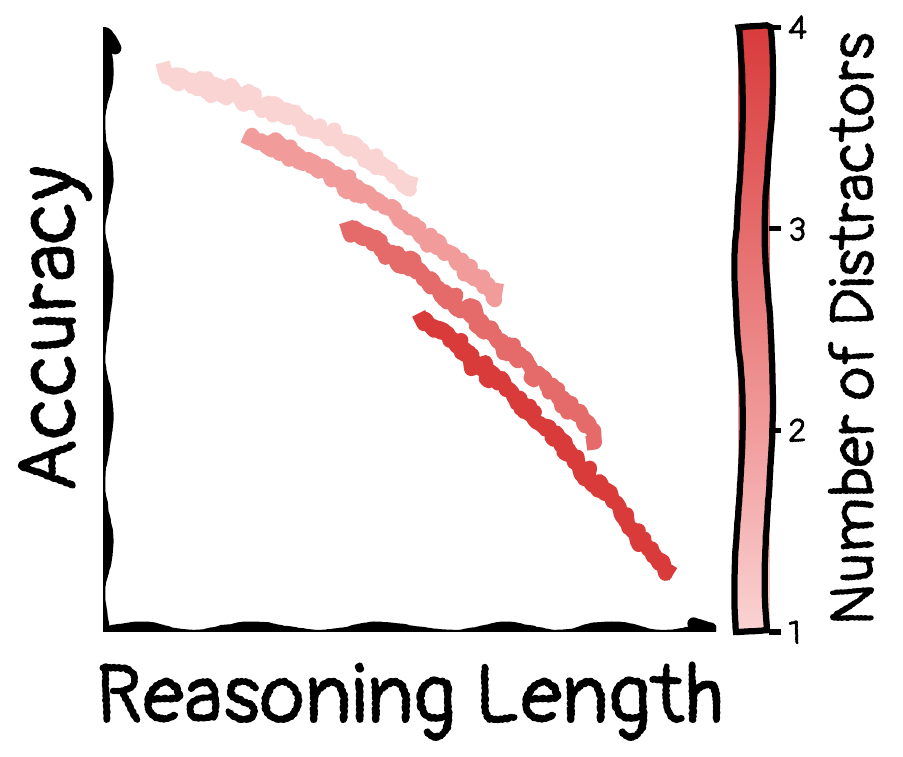}
\end{minipage}

\vspace{-2mm}
\noindent\hspace*{0.01\linewidth}%
\hdashrule{0.99\linewidth}{0.6pt}{3mm 1mm}
\vspace{-4mm}

\begin{minipage}[c]{0.04\linewidth}
  \raggedleft (b)
  \label{fig:intro_b}
\end{minipage}%
\hspace{1mm}%
\begin{minipage}[c]{0.63\linewidth}
  \centering
  \includegraphics[width=\linewidth,trim=40 0 40 5,clip]%
    {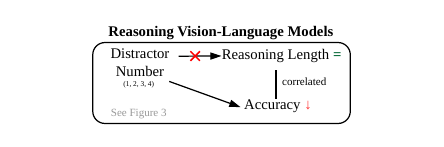}
\end{minipage}%
\hspace{1mm}%
\begin{minipage}[c]{0.23\linewidth}
  \centering
  \includegraphics[width=\linewidth,trim=10 3 13 0,clip]%
    {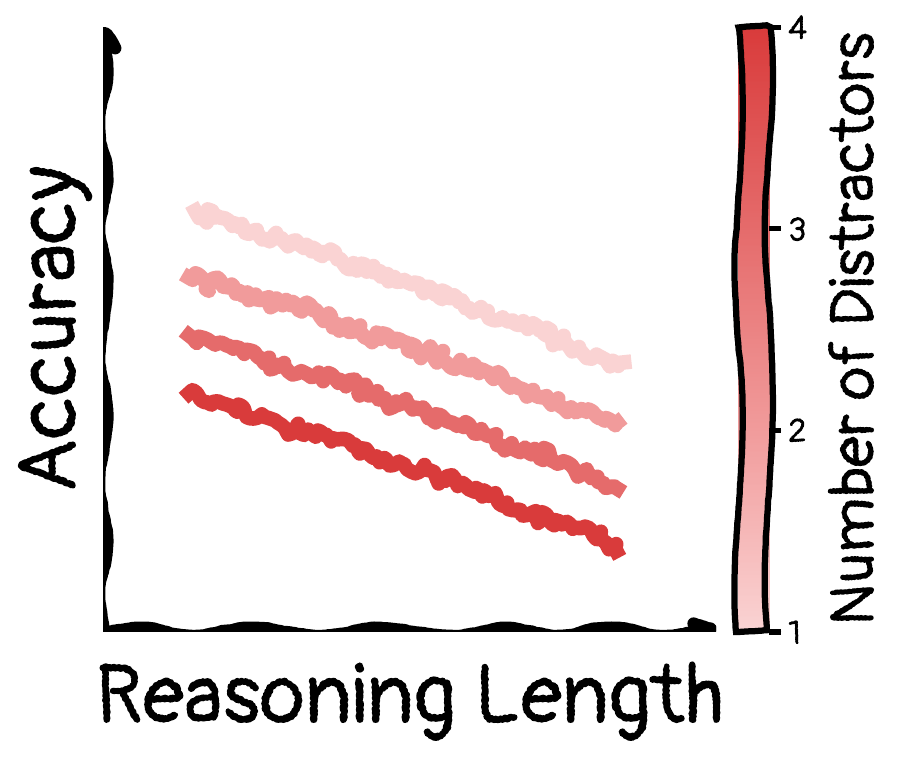}
\end{minipage}

\begin{minipage}[c]{0.04\linewidth}
  \raggedleft (c)
  \label{fig:intro_c}
\end{minipage}%
\hspace{1mm}%
\begin{minipage}[c]{0.63\linewidth}
  \centering
  \includegraphics[width=\linewidth,trim=35 0 40 5,clip]%
    {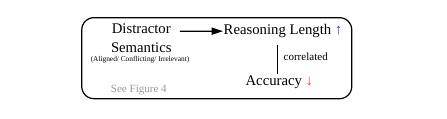}
\end{minipage}%
\hspace{1mm}%
\begin{minipage}[c]{0.23\linewidth}
  \centering
  \includegraphics[width=\linewidth]%
    {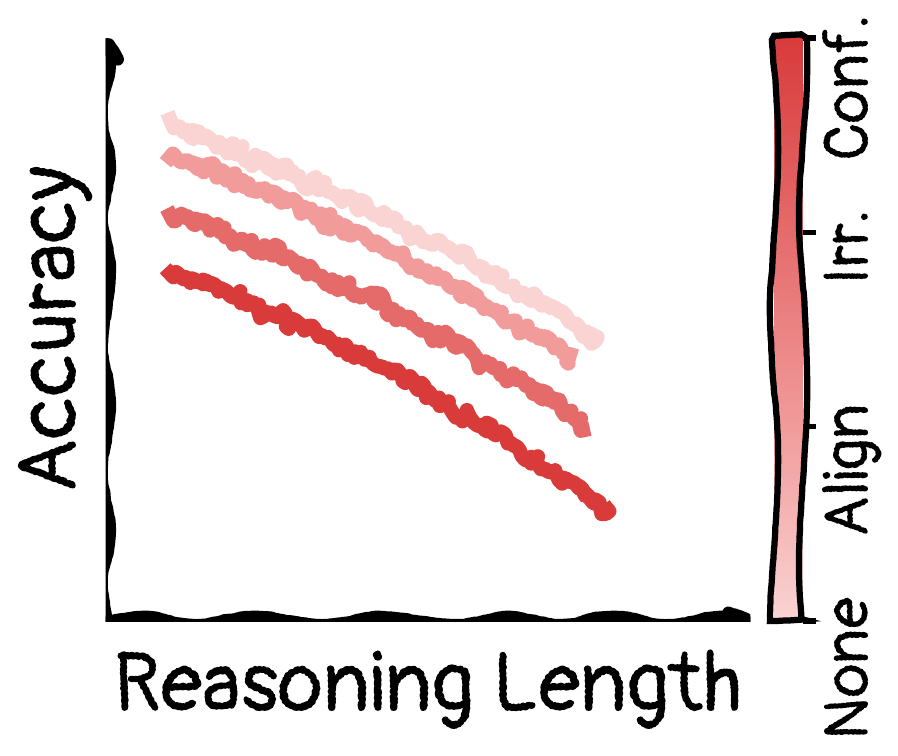}
\end{minipage}

\begin{minipage}[c]{0.04\linewidth}
  \raggedleft (d)
  \label{fig:intro_d}
\end{minipage}%
\hspace{1mm}%
\begin{minipage}[c]{0.63\linewidth}
  \centering
  \includegraphics[width=\linewidth,trim=35 0 60 5,clip]%
    {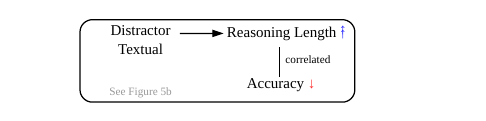}
\end{minipage}%
\hspace{1mm}%
\begin{minipage}[c]{0.23\linewidth}
  \centering
  \includegraphics[width=\linewidth,trim=10 3 10 0,clip]%
    {figure/illustration_inverse_scaling_top.pdf}
\end{minipage}
\caption{\textbf{Test-time scaling of reasoning VLMs.} 
Adding visual distractors decreases accuracy without increasing reasoning length, shifting the entire length--accuracy curve downward, unlike reasoning LMs. In contrast, textual distractors inserted into the prompt intensify the test-time inverse scaling pattern.
}
\vspace{-6mm}
\label{fig:intro}
\end{figure}
\section{Introduction}\label{sec:intro}

\vspace{-1mm}
Increasing test-time computation---e.g., generating more tokens during inference---has proven to be an effective strategy for improving the prediction quality of language models (LMs), and similar benefits have been observed for vision-language models (VLMs). In particular, reasoning VLMs equipped with long chain-of-thought traces have shown impressive performance across tasks requiring multimodal understanding, ranging from visual question-answering (VQA) to mathematical reasoning and embodied tasks \citep{qwen3vl,interns1,glm41vthinking}.

However, is longer reasoning always beneficial? Unfortunately, the answer is no. Reasoning models are prone to several failure modes: they may \textit{\textbf{overthink}}---producing lengthy reasoning traces without improving upon shorter, non-reasoning outputs \citep{chen2024not,sui2025stop}---or even exhibit \textit{\textbf{inverse scaling}}, where increased test-time computation consistently degrades output quality \citep{cuadron2025danger,hassid2025don,inverse_scaling_ttc}. These observations highlight the need for a clearer understanding of the factors that drive such scaling failures. Yet, systematic analyses of these phenomena remain limited.

Recent findings on text-only LMs provide an important clue regarding the cause of scaling failures \citep{inverse_scaling_ttc}. In particular, prior work highlights the role of \textit{\textbf{textual distractors}} in triggering inverse scaling, identifying two consistent patterns: First, the presence of irrelevant information in the context (i.e., distractors) consistently induces the scaling failure. Second, adding more distractors lengthens the reasoning process, which in turn reduces accuracy (\cref{fig:intro}a). These observations suggest that longer reasoning may amplify flawed heuristics introduced by distractors.

In this work, we study whether \textit{\textbf{distractors}} introduced through \textit{\textbf{visual}} and \textit{\textbf{textual}} channels induce analogous failure modes in reasoning VLMs. Rather than focusing only on a single distractor modality, we examine a broader class of distractors, including objects or rendered text inserted into the image and irrelevant information appended to the question prompt. Our motivation is twofold. First, real multimodal inputs often contain irrelevant information in both the visual and linguistic modalities, such as background clutter, contextual objects, or distracting descriptions. Second, VLMs may rely on visual and textual evidence in different ways, raising the question of whether distractors affect test-time scaling through their content itself or through the modality in which they are presented.


To systematically examine the effect of visuo-linguistic distractors on the test-time scaling behavior of reasoning VLMs, we construct Idis (\underline{I}mages with \underline{dis}tractors), a new VQA benchmark suite built from two task families: perception-centric object classification (Idis-perception) and reasoning-centric visual mathematics problems (Idis-math).
Idis comprises over 277k natural and synthetic images featuring target objects alongside one or more distractors, while preserving the original target information and answer. 

Using Idis, we reveal that the scaling pattern differs substantially depending on the modality of distraction. Visual distractors degrade accuracy without substantially increasing reasoning length, shifting the length-accuracy curve downward (\Cref{fig:intro}b,c). This effect further depends on distractor severity, including the number of distractors and their semantic relationship to the target information. In contrast, textual distractors appended to the question prompt more closely resemble the pattern reported in reasoning LMs \citep{inverse_scaling_ttc}: they increase reasoning length and thereby intensify test-time inverse scaling (\Cref{fig:intro}d).

To better understand the behavior of reasoning VLMs, we analyze the visual attributes verbalized in reasoning traces. We find that distractor attribute ratio is strongly associated with final accuracy, and attention-blocking experiments further suggest that attributes in the trace directly affect answer formation. Motivated by this finding, we test a simple prompt-based sanity check that encourages models to focus on the target object and reason from its own visual attributes. This reduces reliance on distractor-related cues and improves robustness on Idis. We further show that the same insight extends to Waterbirds \citep{sagawa2019distributionally}, where distractors arise from background context rather than localized inserted objects.

Our key contributions are as follows:
\begin{itemize}[leftmargin=*,noitemsep, topsep=0pt]
\item We introduce \textbf{Idis}, a VQA benchmark suite that systematically varies distractor modality, number, and semantic relationship across both perception-centric and reasoning-centric tasks.
\item We identify the \textbf{modality-dependent effect of distractors} on scaling that departs from prior findings on reasoning LMs \citep{inverse_scaling_ttc}.
\item Our \textbf{attribute-centric analysis} reveals that the distractor attribute ratio---not reasoning length---drives accuracy for visual distractors, motivating an attribute-guiding prompt strategy that improves robustness on Idis and Waterbirds.
\end{itemize}


\begin{figure*}[t]
\begin{center}
\begin{overpic}[width=1\linewidth,]{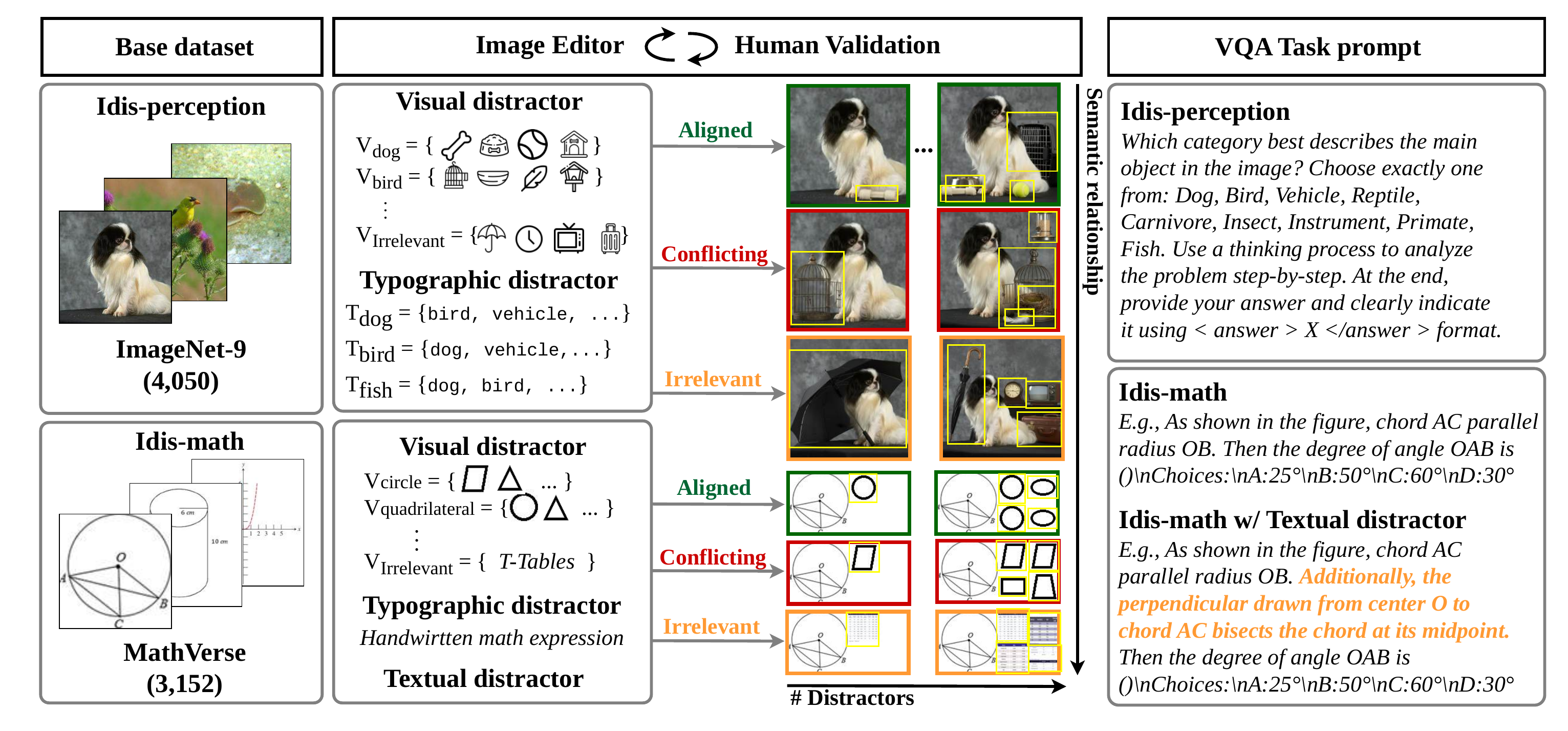}
\end{overpic}
\end{center}
\vspace{-1em}
\caption{\textbf{Idis construction pipeline and VQA task prompt.} {
Idis is a VQA benchmark suite that aims to evaluate how visual and linguistic distractors affect the test-time scaling behavior of reasoning VLMs. Distractor-augmented images are produced through generative editing or deterministic compositing and manually validated through a human-in-the-loop process. \inlinebox{Yellow boxes} indicate distractor regions (not included in original images).}
}
\label{fig:datapipe}
\end{figure*}

\section{Related work}\label{sec:related_work}

\paragraph{Test-time scaling and distractors.} Recent work shows that reasoning models often overthink, spending excessive tokens for only marginal accuracy gains \citep{sui2025stop,chen2024not,yeo2025demystifying,luo2025o1}. More recent studies further report \textit{inverse scaling}, where accuracy decreases as more test-time token usage increases. \citet{cuadron2025danger} observe this phenomenon in interactive settings, such as agentic tasks, with related trends reported in subsequent studies \citep{hassid2025don,marjanovic2025deepseek,pham2025sealqa}. Most closely related to our work, \citet{inverse_scaling_ttc} show that inverse scaling can arise reliably in the presence of distractors, such as irrelevant text or code. We extend this line of work to reasoning VLMs by formalizing the notion of visual and textual distractors in VQA and studying their impact on test-time scaling.

\paragraph{Distractors in VLMs.}
Prior work has shown that reasoning VLMs are vulnerable to diverse distractors, including GUI pop-ups \citep{ma-etal-2025-caution}, textual overlays \citep{deng2025words}, and unrelated images \citep{cai2025diagnosing}, which can substantially degrade accuracy. Our work differs in three ways. First, we study how distractors affect scaling behavior, rather than accuracy alone. Second, we systematically vary distractor severity along multiple dimensions---including semantic type, number, and text placement---and compare text distractors embedded in the image with those appended to the prompt.
Third, we extend the analysis to reasoning-centric mathematical tasks, providing, to the best of our knowledge, the first systematic study of how distractors affect visual mathematical reasoning.


\section{Idis: Images with distractors}
To study how distractors affect test-time scaling in reasoning VLMs, we introduce \textit{\textbf{Idis}} (\underline{I}mages with \underline{dis}tractors), a VQA benchmark that evaluates how visual and linguistic distractors shape the scaling trend. Each sample requires a model to answer a visual question while exposed to distractors to the original problem. Idis comprises two task families: \textit{Idis-perception}, an object classification task based on ImageNet-9 \citep{xiao2020role}, and \textit{Idis-math}, a visual math reasoning task based on MathVerse \citep{zhang2024mathverse}. Together, Idis allows us to study distractor effects in both perception-centric and reasoning-centric settings.

\subsection{Distractor design} \label{sec:distracting_objects}
We design distractors along two broad axes: visual and linguistic. Visual distractors are additional objects inserted into the image.
Linguistic distractors, in contrast, introduce distracting linguistic information embedded as an image or as text. We distinguish between typographic distractor, where text is rendered in the image, and textual distractor, where text is inserted into the question prompt.

\paragraph{Semantics of visual distractors.} We categorize visual distractors as \textit{aligned}, \textit{conflicting}, or \textit{irrelevant}, according to their semantic relationship with the target information needed to answer the question. 
For Idis-perception, this relationship is defined at class level: we follow the debiasing literature's notion of correlated features \citep{nam2020learning} and consider whether the distractor is spuriously associated with the target object class. For Idis-math, we use the same terminology analogously, but at the concept level, describing the relationship between the distractor and the mathematical concept required by the original problem.

\begin{itemize}[leftmargin=*,noitemsep, topsep=0pt]
\item \textit{Aligned.} The distractor is positively associated with the target class or belongs to the same concept family as the original problem: In Idis-perception, a bird cage for ``bird''; in Idis-math, another circle figure for a circle-related problem.
\item \textit{Conflicting.} The distractor is associated with a different class or lies outside the relevant mathematical concept space: In Idis-perception, a bird cage for ``vehicle''; in Idis-math, or a quadrilateral for a circle-related problem.
\item \textit{Irrelevant.} The distractor has no strong association with the target class or concept: In Idis-perception, a TV; In Idis-math, a table.
\end{itemize}
For Idis-perception, aligned distractors have the positive semantic relationship with the target, while conflicting distractors have the negative one. Thus, aligned distractors may reinforce the model’s correct decision, whereas conflicting distractors may bias the model toward the distractor-associated class. Irrelevant distractors, in contrast, test robustness to semantically unrelated visual information.

For Idis-math, however, aligned distractors are not necessarily helpful: because distractors are not part of the original problem, they are unnecessary for deriving the correct answer. Here, the aligned/conflicting/irrelevant taxonomy instead captures the concept-level relationship between the distractor and the mathematical concept needed by the problem. In particular, irrelevant distractors are drawn from the table subset of LogicVista \citep{xiao2024logicvista}, lying outside the geometry-focused problem space of MathVerse \citep{zhang2024mathverse}.

\paragraph{Number.} We add one to four distractors to each sample. When multiple distractors are included, they are drawn from the same semantic category (e.g., aligned), allowing us to isolate the effect of distractor quantity while holding the semantic relationship fixed. The construction varies by task. 
For Idis-perception, multiple distractors differ in fine-grained class; for example, an image containing a dog may be augmented with two aligned distractors such as a dog bowl and a wooden kennel. 

\paragraph{Linguistic distractors.}
Linguistic distractors fall into two types: typographic and textual. 
\begin{itemize}[leftmargin=*,noitemsep, topsep=0pt]
\item \textit{Typographic.} A text segment is rendered directly into the image, making the distracting information part of the visual input. In Idis-perception, we insert the class name of non-target classes; in Idis-math, we insert handwritten math expressions \citep{humynlabs_english_handwritten_math_notes, gervais2025mathwriting}.

\item \textit{Textual.} A text segment, generated by Sonnet 4.5 \citep{anthropic2025claudesonnet45}, is inserted into the question prompt as natural language. This is used only for Idis-math, as the textual prompt is fixed for all samples in Idis-perception.
\end{itemize}

\subsection{Dataset construction pipeline} \label{sec:3-2}
\label{ssec:data_generation_pipeline}

We construct Idis by systematically introducing distractors into the same underlying VQA examples. 
Starting from task-specific base datasets, we create distractor-conditioned variants by inserting controlled visual or linguistic distractors while preserving the original target information and answer. Throughout this process, we incorporate human verification to ensure that the target remains unchanged, the answer is preserved, and each distractor matches its intended semantic category.

Crucially, we generate multiple distractor-conditioned samples from each base example. The target object or original math problem is kept fixed, while only the distractor type, semantic relationship, or number is varied. This controlled intervention enables direct comparison between the original and distractor-augmented examples, allowing changes in model predictions and reasoning behavior to be attributed to the inserted distractors.




\paragraph{Idis-perception.} We use the ``original'' split of ImageNet-9 \citep{xiao2020role} as the base dataset, which contains 4,050 curated images from ImageNet classification benchmark \citep{5206848}. These images contain a single salient foreground object with minimal background clutter, enabling precise control over the number and type of distracting objects inserted into each image.

Na\"{i}vely compositing distractors into the base image, such as by direct overlay, can introduce artifacts that VLMs detect instead of reasoning about the image content, leading to traces such as \textit{``the image might be a composite or edited photo.''} To produce realistic edits while preserving the original target object, we use Gemini 2.5 Flash Image \citep{Gemini25FlashImage} for generative editing. For each image, we create a distractor-conditioned variant by sampling $k$ distinct distractor classes from one semantic category: aligned, conflicting, or irrelevant. 
We then prompt Gemini 2.5 Flash Image to insert the selected distractors while preserving the identity, location, and appearance of the target object.
For typographic distractors, we instead overlay the distractor class name directly onto the image. 

\paragraph{Idis-math.} 
We construct \textit{Idis-math} by adding controlled distractors to visual math problems from the MathVerse \citep{zhang2024mathverse}, while preserving the original problem structure and answer.


\begin{figure*}[t]
  \centering

  \begin{subfigure}{0.24\linewidth}
    \includegraphics[width=\linewidth]{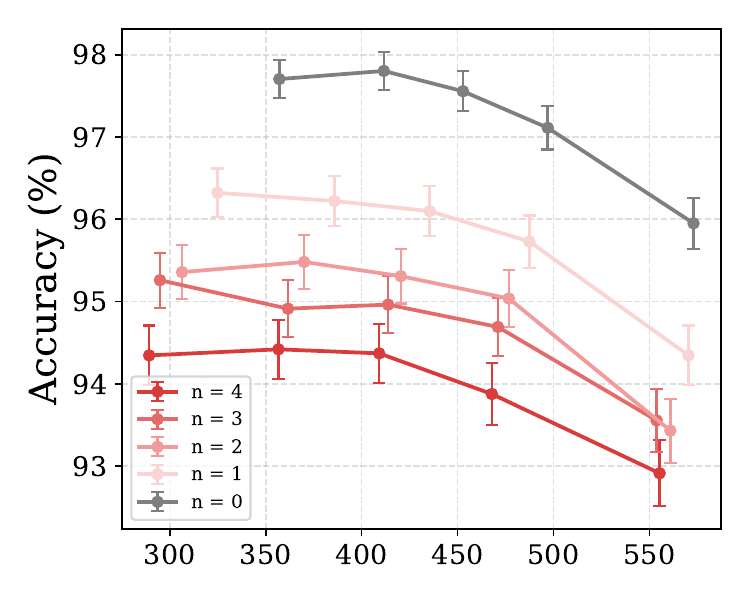}
  \end{subfigure}\hfill
  \begin{subfigure}{0.24\linewidth}
    \includegraphics[width=\linewidth]{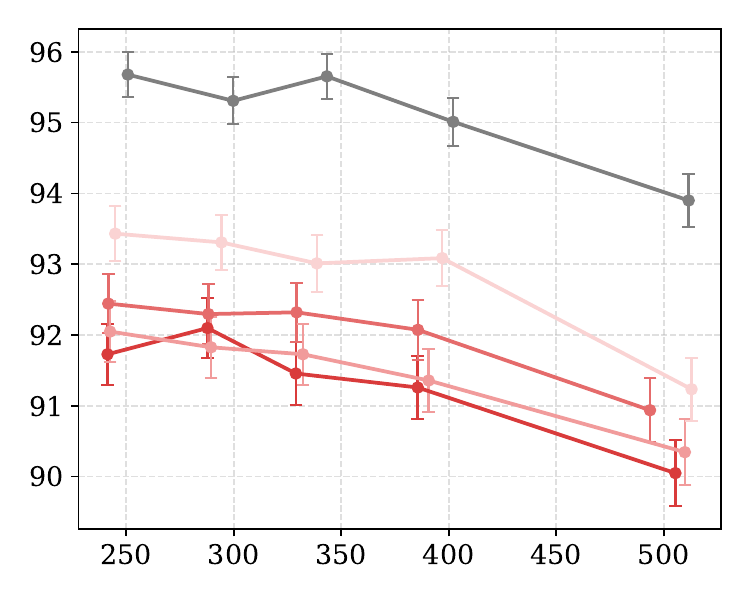}
  \end{subfigure}\hfill
  \begin{subfigure}{0.24\linewidth}
    \includegraphics[width=\linewidth]{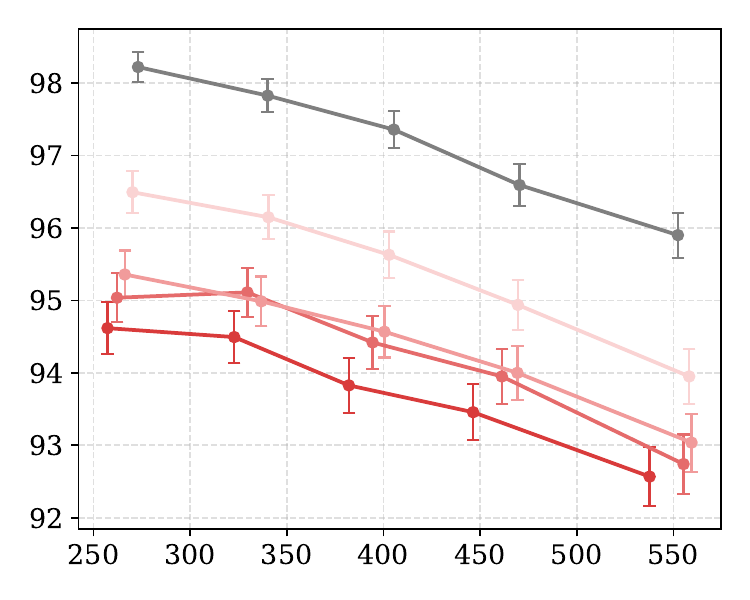}
  \end{subfigure}\hfill
  \begin{subfigure}{0.24\linewidth}
    \includegraphics[width=\linewidth]{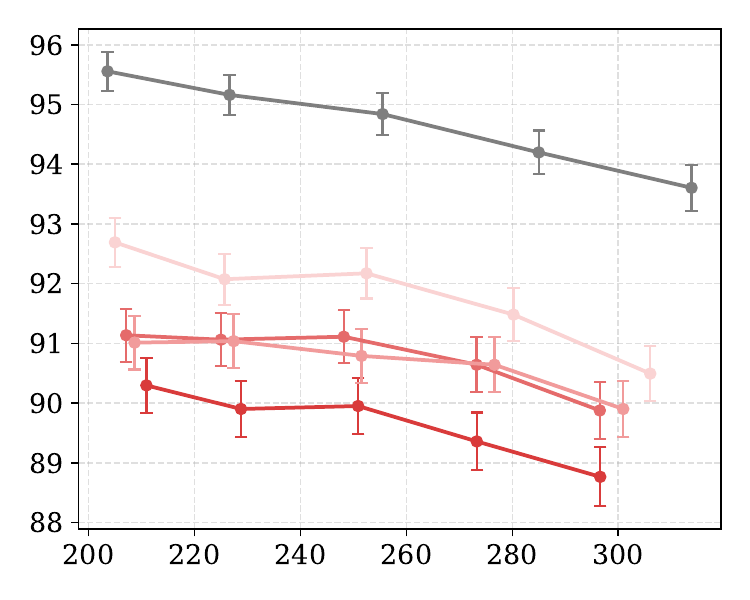}
  \end{subfigure}

  \vspace{0.5em}

  \begin{subfigure}{0.235\linewidth}
    \includegraphics[width=\linewidth]{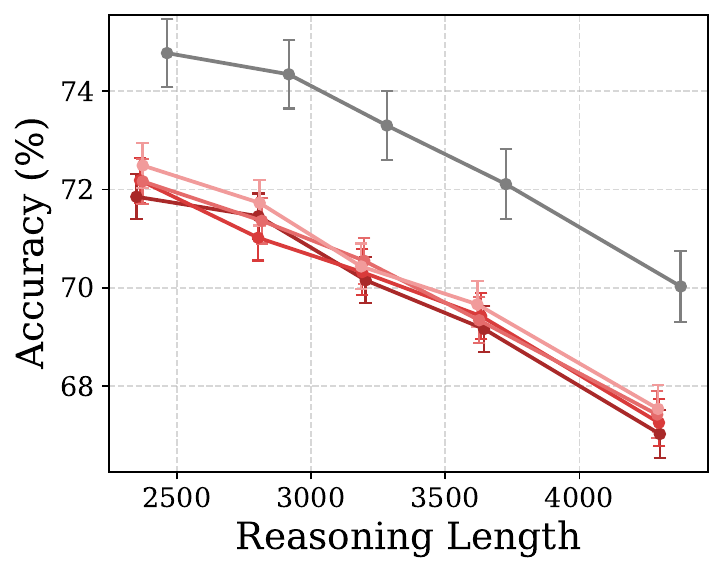}
    \subcaption{Qwen3-VL-Thinking}\label{fig:number_e}
  \end{subfigure}\hfill
  \begin{subfigure}{0.235\linewidth}
    \includegraphics[width=\linewidth]{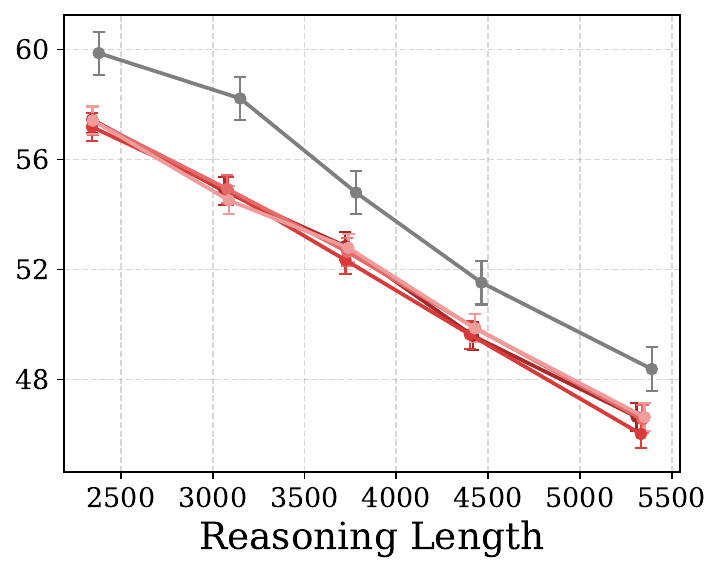}
    \subcaption{Intern-S1-mini}\label{fig:number_f}
  \end{subfigure}\hfill
  \begin{subfigure}{0.235\linewidth}
    \includegraphics[width=\linewidth]{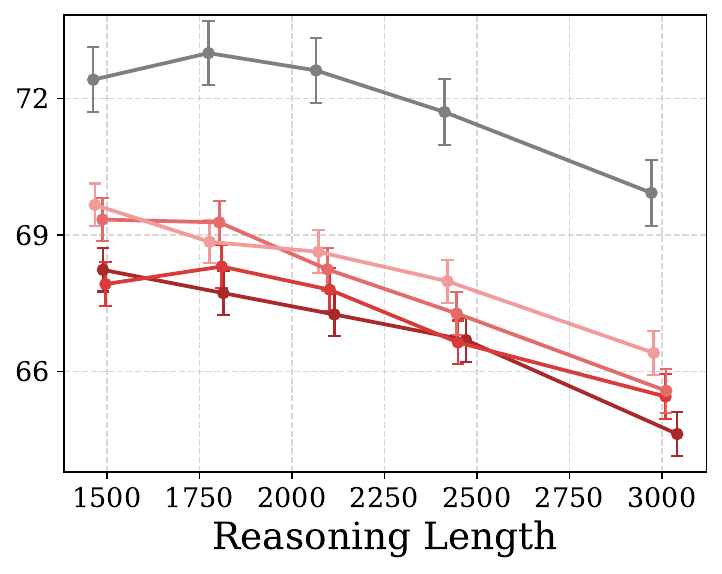}
    \subcaption{GLM-4.1V-Thinking}\label{fig:number_g}
  \end{subfigure}\hfill
  \begin{subfigure}{0.235\linewidth}
    \includegraphics[width=\linewidth]{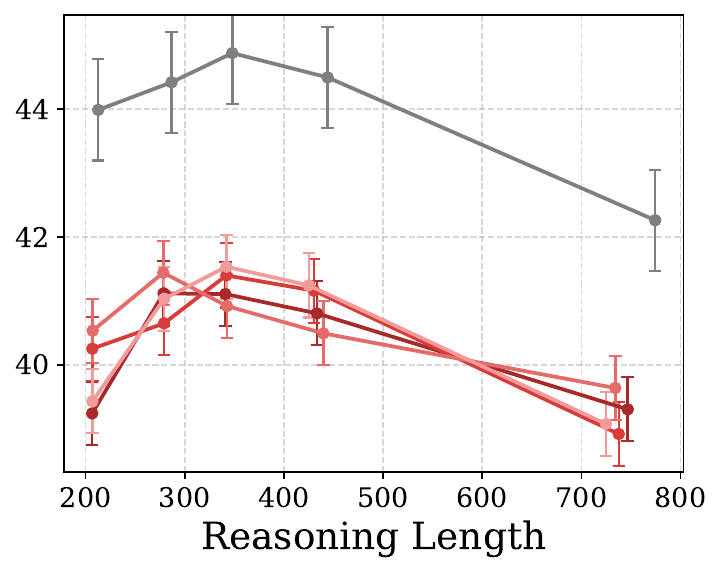}
    \subcaption{R1-OneVision}\label{fig:number_h}
  \end{subfigure}

  \caption{\textbf{Test-time scaling vs. \#distractors.}
  Inserting more visual distractors does not extend the reasoning length of reasoning VLMs, but instead shifts the length-accuracy curve downward, in contrast to reasoning LMs. The first and second rows report results on Idis-perception and Idis-math, respectively.}
  \label{fig:distractor-num}
\end{figure*}

For visual distractors, we deterministically composite distractors around the target diagram on a fixed canvas using PIL, rather than relying on generative image editing. We preserve the original resolution and aspect ratio, and place distractors to avoid overlap with the target figure.

For typographic distractors, we use the same rendering protocol, but insert handwritten math expressions \citep{humynlabs_english_handwritten_math_notes, gervais2025mathwriting} instead of geometric figures. These expressions are placed around the target diagram as visually salient but answer-irrelevant distractors.

For textual distractors, we leave the diagram unchanged and insert mathematically plausible but answer-irrelevant sentences into the question prompt. These sentences are constructed to be consistent with the diagram, but not necessary for deriving the correct answer.

\paragraph{Human verification.} 
To ensure dataset quality, we perform iterative human verification until all samples in the benchmark satisfy subtask-specific acceptance criteria. Samples that fail verification are either regenerated or excluded. \Cref{sec:suppl_B} details the verification protocol, including selection criteria and filtering statistics.

\paragraph{Other details.} We provide omitted details, including the dataset statistics, prompts, and human verification details in \Cref{sec:suppl_B}.


\section{Experimental setup}\label{sec:setup}

Following \citet{inverse_scaling_ttc}, we study \textit{sequential scaling}, where models scale by producing longer reasoning trace rather than aggregating parallel reasoning traces. For each question, we sample five responses, rank them by reasoning length, and compute the accuracy of each rank across all questions. Each rank is plotted at its average reasoning length across questions. This repeated sampling controls for question difficulty and image complexity. Thus, the resulting length-accuracy curves reflect within-question variation in reasoning length, rather than across-question differences in problem hardness that could otherwise confound the analysis.

\paragraph{Models.} We evaluate four open-weight frontier reasoning VLMs in the 7--9B parameter range, spanning diverse architectures and training recipes: Qwen3-VL-8B-Thinking \citep{qwen3vl}, GLM-4.1V-9B-Thinking \citep{glm41vthinking}, Intern-S1-mini \citep{interns1}, and R1-OneVision-7B-RL \citep{yang2025r1}. For brevity, we omit parameter counts when referring to these models hereafter.

\paragraph{Reasoning budget.} Unlike text-only reasoning models, reasoning VLMs typically lack an explicit mechanism to control the reasoning budget.\footnote{We have also attempted controlling the reasoning length via prompting, which turned out to be ineffective; see \Cref{sec:suppl_E} for details.
} Thus, we mainly focus on the setting of ``natural overthinking,'' i.e., the models naturally generating extended reasoning \citep{inverse_scaling_ttc}.


\paragraph{Metrics.} Following \citet{inverse_scaling_ttc}, we primarily focus on the interplay between three elements: (1) the accuracy; (2) the reasoning length; (3) the number of distractors added to the sample.



\section{Results} \label{sec:analysis}

We summarize our main findings as follows:
\begin{itemize}[leftmargin=*,noitemsep, topsep=0pt]
\item \textit{Visual distractors} lower the length-accuracy curves without substantially increasing reasoning length; the effect depends on the number of distractors and their semantics (\Cref{sec:5-A}).
\item \textit{Linguistic distractors} have modality-dependent effects: typographic distractors mainly shift the curve downward (as visual distractors do), whereas textual distractors reduce accuracy by increasing reasoning length (\Cref{sec:5-B}).
\end{itemize}

\subsection{Scaling vs. visual distractors} \label{sec:5-A}
\paragraph{Number of distractors.} 
We first put particular focus on how the number of the distractors affects the scaling trend.
In reasoning LMs \citep{inverse_scaling_ttc}, adding more textual distractors intensifies test-time inverse scaling. In contrast, \cref{fig:distractor-num} shows that additional visual distractors consistently shift the length-accuracy curve downward without substantially changing the range of reasoning lengths. On Idis-math (bottom row), we observe similar trends, except that the marginal effect of having more than one distractor is notably smaller. We also draw particular attention to the curves for R1-OneVision. Here, the overall shape of the curve is maintained despite the shifts.

\begin{figure}[t]
  \centering

  \begin{subfigure}{0.5\linewidth}
    \centering
    \includegraphics[width=\linewidth]{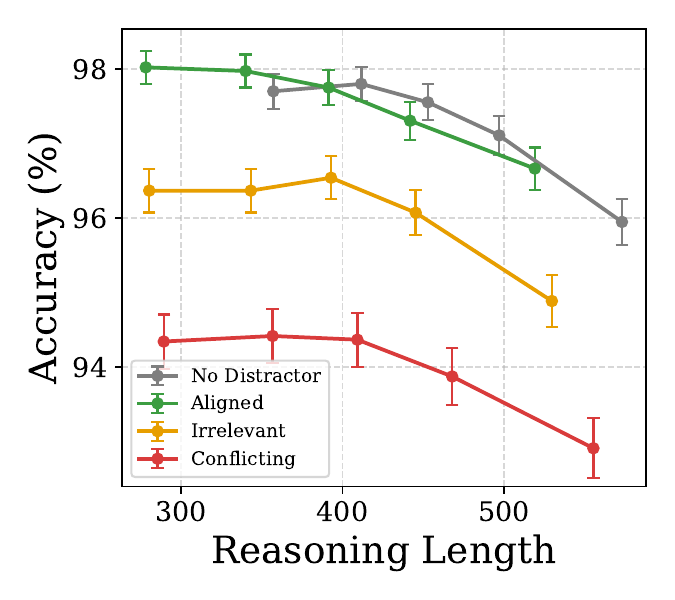}
    \subcaption{Idis-perception}
    \label{fig:semantic_a}
  \end{subfigure}\hfill
  \begin{subfigure}{0.485\linewidth}
    \centering
    \includegraphics[width=\linewidth]{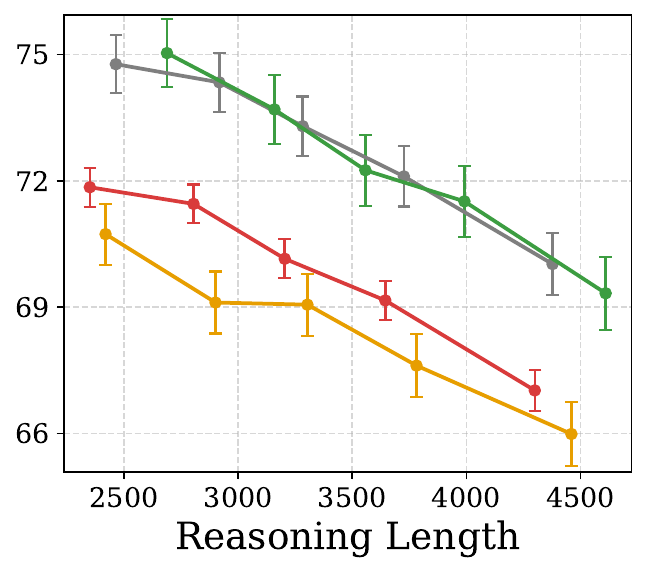}
    \subcaption{Idis-math}
    \label{fig:semantic_b}
  \end{subfigure}

  \caption{\textbf{Test-time scaling vs. distractor semantics.}
  Conflicting distractors cause the largest downward shift, indicating that reasoning VLMs are vulnerable to distractors that semantically conflict with the target object. Results are on Qwen3, when we add four distractors at a time; see \cref{sec:suppl_F-1} for other VLMs.}
  \label{fig:semantic}
\end{figure}

\paragraph{Distractor semantics.} 
Next, we study the interplay between the scaling behavior and the semantic properties of the distractors. In \cref{fig:semantic}, we observe that distractor semantics also tend to shift the scaling curve downward along the $y$-axis, similar to the effect of increasing the number of distractors. For Idis-perception, the downward shift is large when the distractors are semantically conflicting with the target, while aligned distractors cause little to no degradation compared to the no-distractor baseline. For Idis-math, interestingly, irrelevant samples give the largest drop; this may be due to the fact that ``irrelevant'' distractors are much OOD-ish for Idis-math, leading to the largest confusion.

\paragraph{Discussion: Severity of distraction.} It is noteworthy that both aspects (numerics and semantics) have a very similar effect; the severity of distraction increases as we have greater number of distractors or more confusing ones. In \cref{ssec:idis_manual}, we show that ``size'' also has a similar effect, where the curve shifts down more for larger distractors.

\subsection{Scaling vs. linguistic distractors} \label{sec:5-B}
We next examine linguistic distractors for two reasons. First, they connect directly to prior findings in reasoning LMs, where irrelevant text amplifies test-time inverse scaling \citep{inverse_scaling_ttc}. Second, in multimodal tasks, the same distraction can be introduced visually, by rendering text in the image, or linguistically, by inserting it into the prompt. This lets us test whether reasoning VLMs respond differently depending on the distractor modality.

\begin{figure}[t]
  \centering

  \begin{subfigure}{0.5\linewidth}
    \centering
    \includegraphics[width=\linewidth]{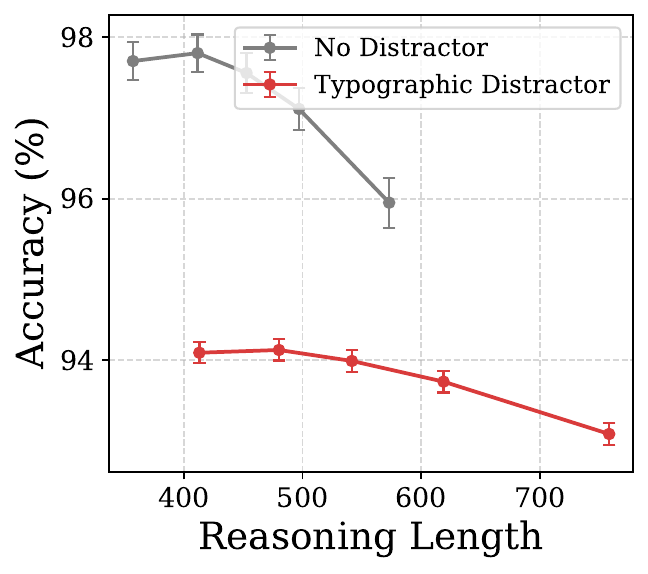}
    \subcaption{Idis-perception}
    \label{fig:text_a}
  \end{subfigure}\hfill
  \begin{subfigure}{0.5\linewidth}
    \centering
    \includegraphics[width=\linewidth]{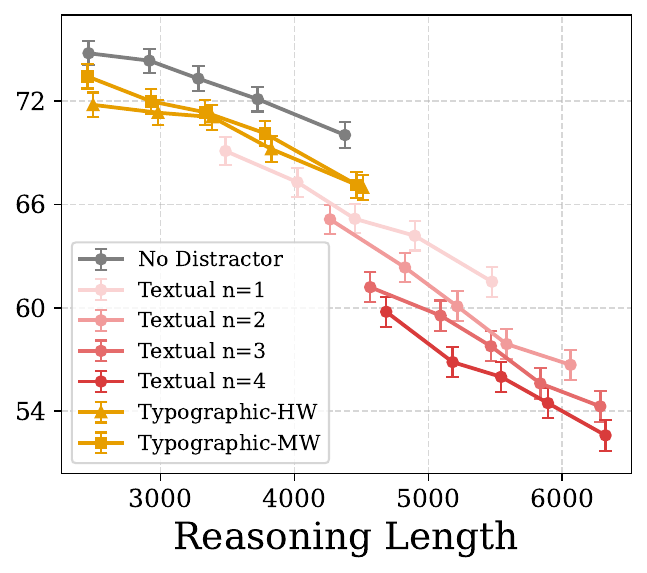}
    \subcaption{Idis-math}
    \label{fig:text_b}
  \end{subfigure}

  \caption{\textbf{Test-time scaling vs. linguistic distractors.} Typographic distractors, rendered as an image, has a similar effect with visual distractors of shifting the curve downward with limited increase in reasoning length. Textual distractors, inserted in the question prompt, also reduces the accuracy by increasing the reasoning length. Results are on Qwen3; see \cref{sec:suppl_F-1} for other VLMs.}
  \label{fig:textual}
\end{figure}

\paragraph{Typographic distractor.} 
As shown in \Cref{fig:textual}, typographic distractors mirror the pattern observed for visual distractors: they reduce accuracy without substantially expanding the range of reasoning lengths. This trend is less pronounced on Idis-perception, but becomes clearer on Idis-math, where we can compare against textual distractors.

\paragraph{Textual distractor.} 
This pattern does not hold for textual distractors inserted into the question prompt. As shown in \Cref{fig:text_b}, increasing the number of textual distractors induces a clear test-time inverse-scaling pattern, similar to that observed by \citet{inverse_scaling_ttc} in text-only reasoning LMs: much accuracy drop is driven by longer reasoning traces.

\paragraph{Discussion: Modality matters.} Together, these observations suggest that the ``modality of distraction'' has a dominant effect in test-time scaling, instead of whether the distractor has been a visual object or linguistic information.

\begin{figure}[t]
  \centering

  \begin{subfigure}{0.5\linewidth}
    \centering
    \includegraphics[width=\linewidth]{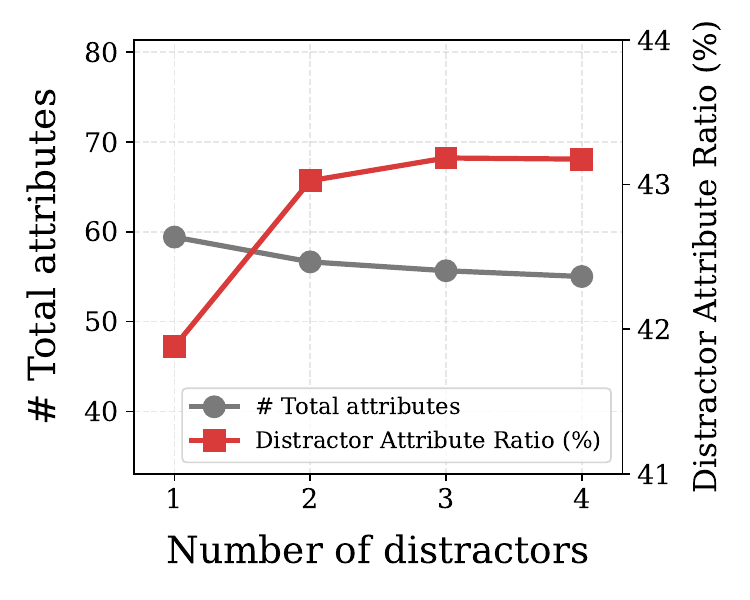}
    \subcaption{Qwen3-VL-Thinking}
  \end{subfigure}\hfill
  \begin{subfigure}{0.5\linewidth}
    \centering
    \includegraphics[width=\linewidth]{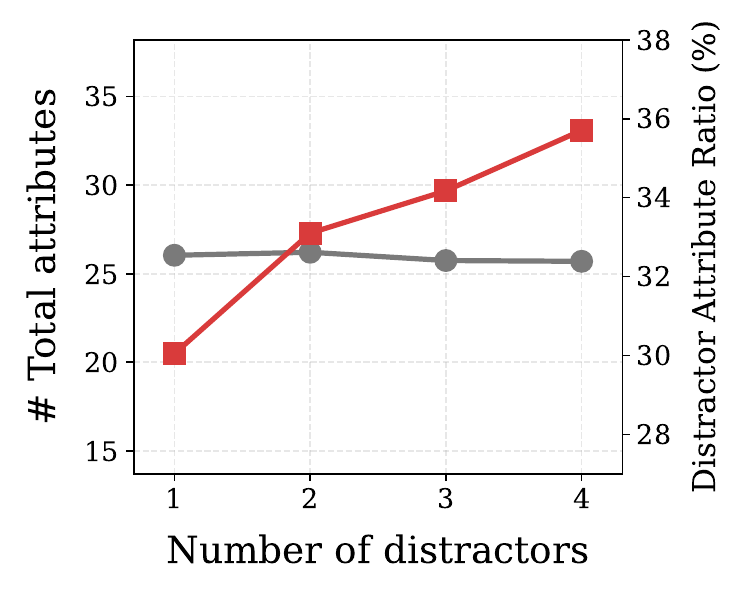}
    \subcaption{R1-Onevision}
  \end{subfigure}

  \vspace{0.5em}

  \begin{subfigure}{0.49\linewidth}
    \centering
    \includegraphics[width=\linewidth]{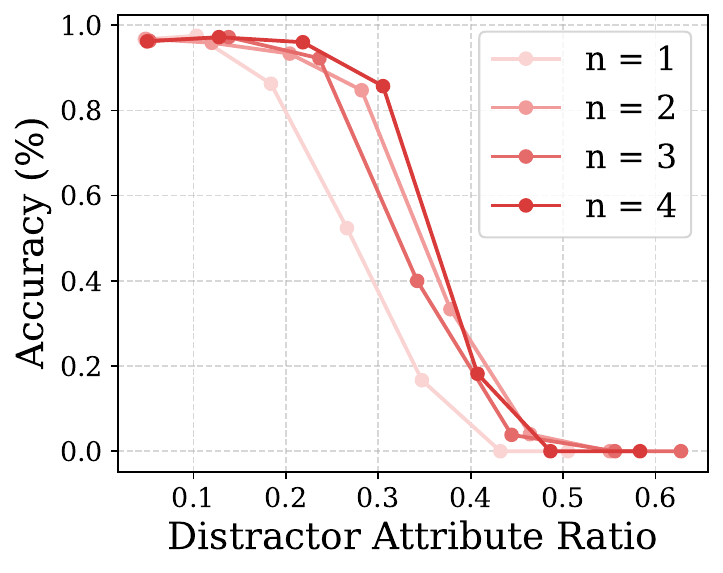}
    \subcaption{Idis-perception}
    \label{fig:sec5-ob1-a}
  \end{subfigure}\hfill
  \begin{subfigure}{0.49\linewidth}
    \centering
    \includegraphics[width=\linewidth]{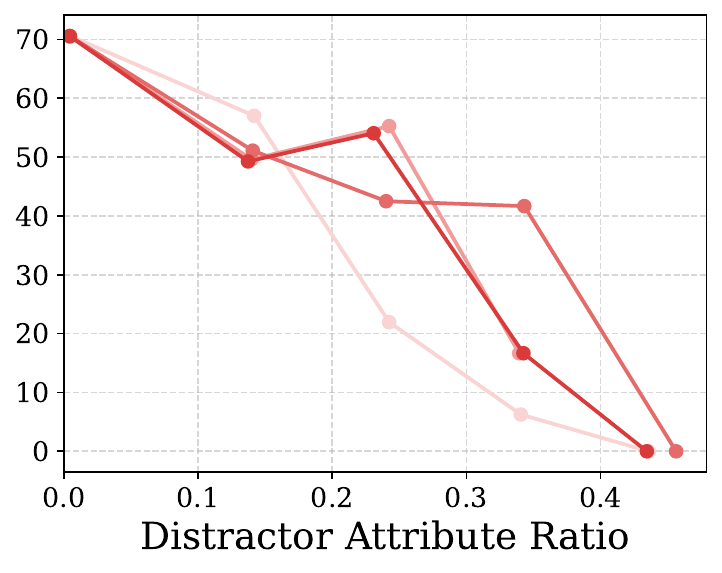}
    \subcaption{Idis-math}
    \label{fig:sec5-ob1-b}
  \end{subfigure}

  \caption{\textbf{Distractor attributes as a meaningful indicator.} In the top row, we find that as we increase the number of distractors, the number of total attributes remain similar but the fraction of distractor attributes consistently increases; see \cref{sec:suppl_F-1} for more plots. The bottom row shows that the distractor attribute ratio is strongly correlated with the model accuracy. The plots are for Qwen 3, when we add conflicting distractors. Plots on other models are given in \cref{sec:suppl_F-1}.}
  \label{fig:sec5-ob1}
\end{figure}

\section{Analysis} \label{sec:6}
To better understand this behavior, we conduct an \textit{\textbf{attribute-centric}} analysis of reasoning traces \citep{deng2025words, shojaee2025illusion, qiandemystifying}. Our analysis reveals a potential limitation of reasoning VLMs: \textit{Final predictions are strongly influenced by the raw fraction of distractor attributes in the trace, which itself increases with the number and size of distractors in the image} (\Cref{sec:6-A}).

This observation suggests that the distractor-robust VLM reasoning may require mechanisms that suppress distractor attributes during trace generation. As a sanity check on this insight, we also explore a prompt-based mitigation strategy that can be applied at the inference time.

\subsection{Tools}

For our analysis, we utilize the following tools.

\paragraph{Attribute extraction.} We parse visual attributes mentioned in each trace and link them to the corresponding target or distractor object. We use DeepSeek-V3.2-Exp \citep{deepseekai2024deepseekv32} with structured instructions; see \cref{sec:suppl_B-1} for procedural details and \cref{sec:suppl_B-parser} for the validation of this extraction process.

\paragraph{Object area.} We measure the pixel area of target and distractor objects and analyze its relation to accuracy and distractor attributes. We obtain masks with LangSAM \citep{langsam}, using each object description as the prompt, and compute area by counting masked pixels.

\paragraph{Detailed formula.} see \cref{ssec:suppl_B_metric}
\subsection{The role of distractor attributes}\label{sec:6-A}

The upper row of \Cref{fig:sec5-ob1} shows that (i) the number of total attributes in each trace remains similar as we vary the number of distractors;\footnote{This is well-expected since the reasoning length remains similar (\cref{fig:distractor-num}). Indeed, the number of attributes is tightly correlated with the reasoning length; see \cref{sec:suppl_F-1}} (ii) the fraction of distractor attributes, on the other hand, steadily grows as we increase the number of distractors.

The lower row of \Cref{fig:sec5-ob1} shows that the distractor attribute ratio is strongly correlated with the final prediction. The accuracy drops to near zero whenever the distractor attribute ratio exceeds 50\%, yet remains above 97\% whenever the ratio is below 20\%. We observe a similar pattern in Idis-math, where accuracy drops to zero when the distractor attribute ratio exceeds 0.4.

\begin{figure}[t]
  \centering
  \begin{subfigure}{0.48\linewidth}
    \centering
    \includegraphics[width=\linewidth]{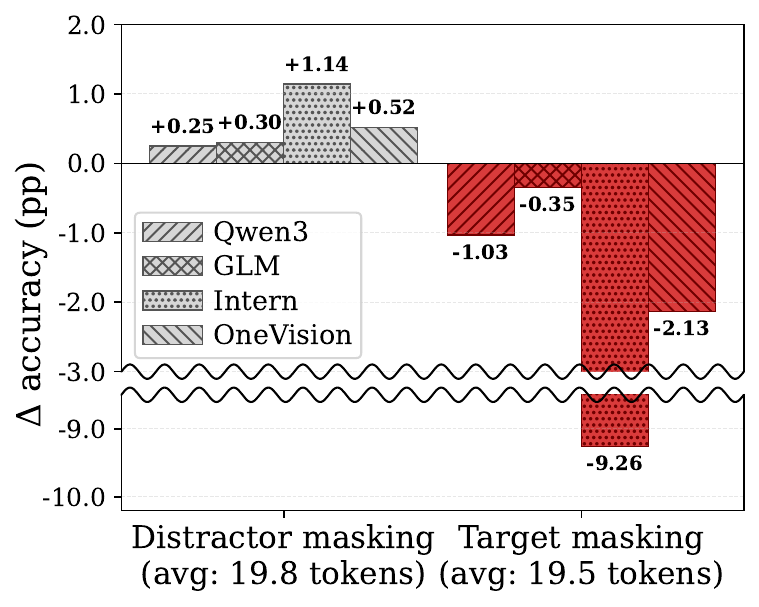}
    \subcaption{Token attention blocking }\label{fig:sec5-ob3-a}
  \end{subfigure}\hfill
  \begin{subfigure}{0.48\linewidth}
    \centering
    \includegraphics[width=\linewidth]{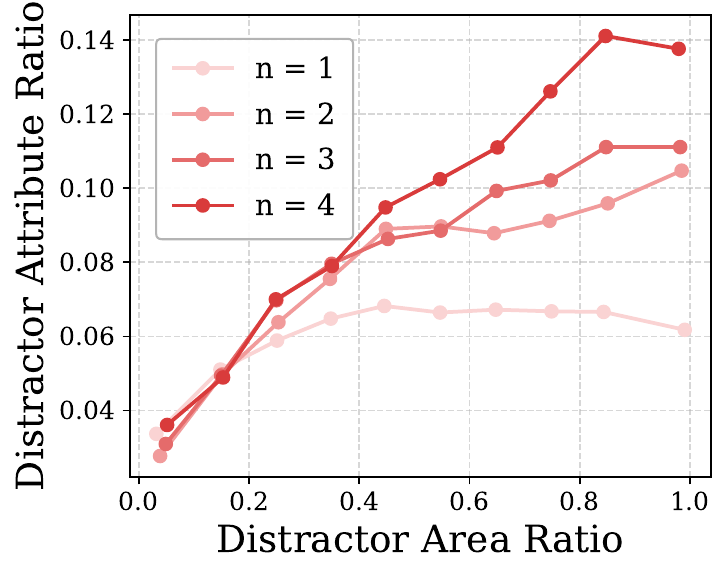}
    \subcaption{Distractor Area Ratio}\label{fig:sec5-ob3-b}
  \end{subfigure}

  \caption{\textbf{Closer look at the distractor attributes.} (a) Answers are sensitive to direct access to reasoning-trace attributes: blocking the attention of the final answer tokens to distractor-related attributes improves accuracy, while blocking target-related attributes degrades it.
  (b) Distractor attributes increase with distractor area and saturate at a higher ratio when a greater number of distractors are present in the image.}
  \label{fig:sec5-ob3}
\end{figure}

\paragraph{Discussion: Which causes which?} Together, the observations suggest that the distractor attribute ratio is highly correlated with the final prediction. Is final prediction guiding the reasoning trace, or is the reasoning trace guiding the final prediction? The attention blocking experiment (\cref{fig:sec5-ob3-a}) suggests that the latter is more plausible than the former. By blocking the attention going to the distractor (or target) attribute tokens, one can dramatically change the model accuracy.

\paragraph{What affects distractor attribute ratio?} \Cref{fig:sec5-ob3-b} suggests that the distractor attribute ratio is correlated with not only the number of distractors, but also the distractor area ratio. In \cref{ssec:idis_manual}, we supplement this observation with the experiments on a dataset where we manually control the size of the distractors (dubbed \textit{Idis-manual}), which confirm the effect of distractor sizes on the distractor attribute ratio.



\subsection{Sanity check: Prompt-based mitigation}

This observation suggests that the VLM reasoning under distractor-prone environments may benefit from the mechanisms that suppress distractor attributes and facilitate target attributes during the reasoning trace generation.

To validate this idea, we design a simple sanity check based on \textbf{\textit{prompting.}} Concretely, we prompt the model in two ways: (i) the original prompts, and (ii) a prompt that explicitly instructs the reasoning model to base its prediction on the target object itself rather than distractor or background. For example, to avoid focusing on non-target objects, one can add sentences like ``First identify the main
object. Base your reasoning only on that object’s own visual attributes.'' to the original prompt. 

\paragraph{Experimental setup.} We test this idea on two datasets: Idis-perception, and Waterbirds \citep{sagawa2019distributionally}. Waterbirds is a standard benchmark in the debiasing literature, where the task is to correctly classify waterbirds from landbirds, where the background may be either the water or the land; the model predictions are often overly reliant on the background cues, making erroneous predictions on images with bias-conflicting backgrounds, e.g., waterbirds on land. We design prompts that instructs the model to focus on the target (or foreground) objects for each of these tasks; see \cref{ssec:suppl_B-3} for the actual prompts.

\begin{figure}[t]
\centering
  \begin{subfigure}{0.48\linewidth}
    \includegraphics[width=\linewidth]{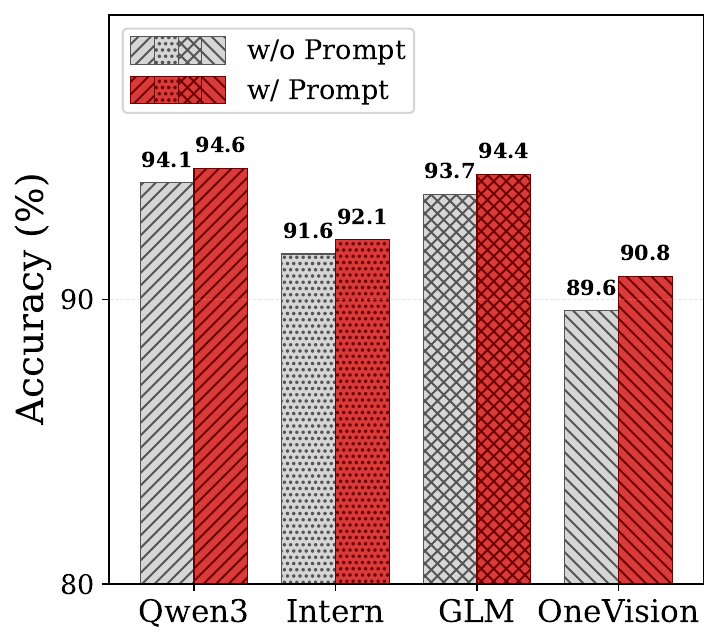}
  \end{subfigure}\hfill
  \begin{subfigure}{0.48\linewidth}
    \includegraphics[width=\linewidth]{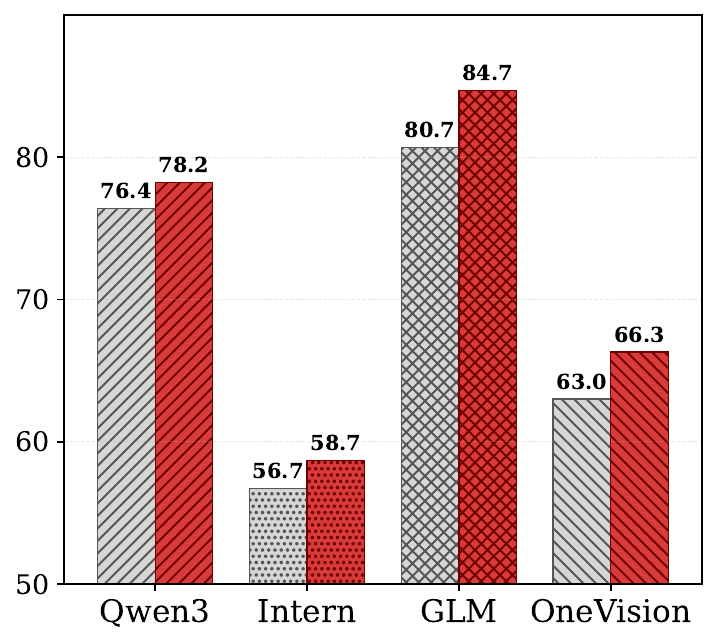}
  \end{subfigure}

  \vspace{1em} 

  \begin{subfigure}{0.48\linewidth}
    \includegraphics[width=\linewidth]{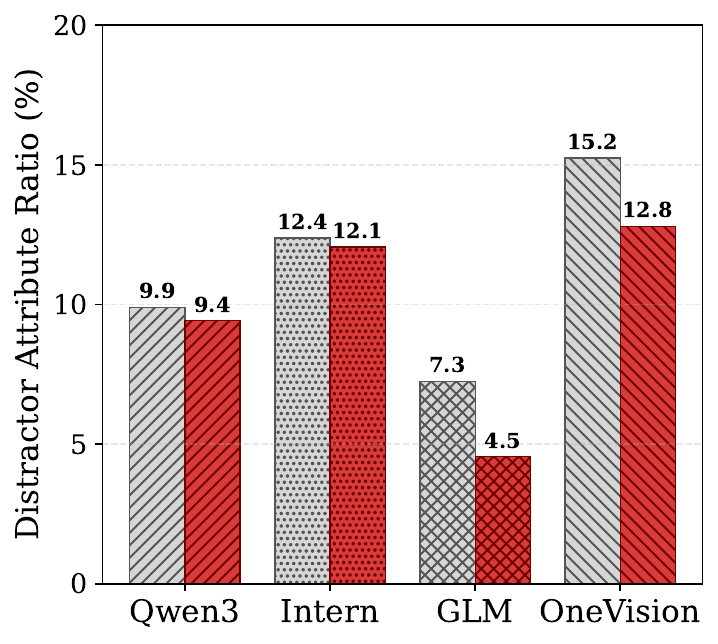}
    \caption{Idis-perception}
    \label{fig:prompt_Idis}
  \end{subfigure}\hfill
  \begin{subfigure}{0.48\linewidth}
    \includegraphics[width=\linewidth]{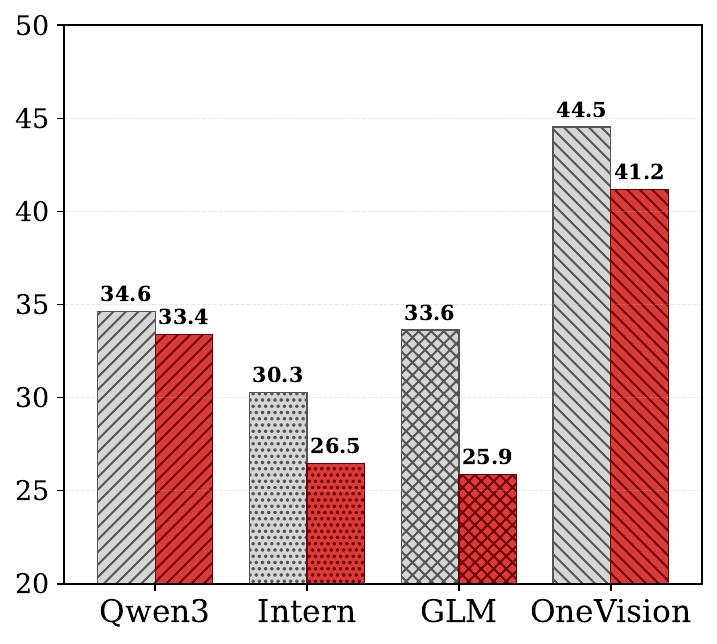}
        \caption{Waterbirds}
    \label{fig:prompt_waterbirds}
  \end{subfigure}


  \caption{\textbf{Specialized prompts can improve accuracy and reduce focusing on distractors.} Top row: The prompting strategy consistently improves the accuracy across all models, on both Idis-perception and Waterbirds. Bottom row: The prompt decreases the fraction of distractor-related attributes in the reasoning trace, which is likely to have caused the accuracy gain.
  }
  \label{fig:combined_results}
\end{figure}

\paragraph{Results.} The experimental results are given in \Cref{fig:combined_results}. The top row provides comparisons on the the accuracy of reasoning VLMs with and without specialized prompts. On the Idis-perception dataset, we observe a slight yet consistent increase in accuracy, ranging from 0.5\%p to 1.2\%p. The accuracy gain is more pronounced on the Waterbirds dataset, ranging from 1.8\%p to 4.0\%p.

The results on the bottom row of \Cref{fig:combined_results} demonstrates that such an accuracy gain is well-correlated with the decrease in the fraction of distractor-related attributes inside the reasoning trace. The distractor attribute ratio decreases quite consistently over all models and on both datasets.

\paragraph{Takeaway.} The proposed prompting strategy \textit{per se} is somewhat limited in its ability to improve the model accuracy. However, the results suggest that it might be possible to mitigate the reasoning model's vulnerability toward blindly extracting and distractor-related attributes in reasoning traces.

\section{Conclusion}

We study the test-time scaling behavior of reasoning VLMs by introducing Idis, a VQA benchmark suite for analyzing visual and linguistic distractors across perception-centric and reasoning-centric tasks. We show that distractor modality shapes scaling behavior: visual and typographic distractors reduce accuracy without substantially increasing reasoning length, whereas textual distractors intensify the test-time inverse scaling. Our attribute-level analysis further shows that reasoning VLMs are influenced by distractor-related attributes verbalized in their traces. Based on this insight, we explore a attribute-guide prompt that improves robustness on both Idis and Waterbirds. We hope this work informs future efforts to build distractor-robust reasoning VLMs.



\section*{Limitations}
Our analyses focus on a VQA domain, which provides a controlled setup to analyze the effects of visual and textual distractors. Extending this framework to more complex reasoning-heavy tasks, such as agentic decision making or multi-step planning, remains challenging and is an essential next step. Furthermore, our findings suggest several opportunities for designing mitigation strategies that guide models toward task-related evidence, but developing such methods in a more systematic and generalizable way remains future work.
\section*{Acknowledgments}

This work was supported by the National Research Foundation of Korea (NRF) grant funded by the Korea government (MSIT) (No. RS-2026-25494004, 40\%), the Institute of Information \& communications Technology Planning \& Evaluation(IITP) under the Leading Generative AI Human Resources Development(IITP-2026-RS-2026-25546560) grant funded by the Korea government(MSIT) (30\%), the Institute of Information \& Communications Technology Planning \& Evaluation (IITP) grant funded by the Korea government (MSIT) (No. RS-2024-00457882, AI Research Hub Project, 10\%; No.RS-2019-II191906, Artificial Intelligence Graduate School Program(POSTECH), 10\%), and partly by the Institute of Information \& communications Technology Planning \& Evaluation (IITP, AI Computing Support Project for R\&D) grant funded by the Korea government(MSIT) (No. RS-2026-25505492, High-Performance Research AI Computing Infrastructure Support at the 2 PFLOPS Scale, 10\%)

\bibliography{custom}

\clearpage
\appendix










\newtcolorbox{prompt_for_data}[1]{
  enhanced,
  breakable,
  colback=white,
  colframe=black!15,
  colbacktitle=black!15,
  coltitle=black,
  fonttitle=\bfseries,
  title=#1,
  boxrule=0.4pt,
  arc=0pt,
  outer arc=0pt,
  left=5pt,
  right=5pt,
  top=4pt,
  bottom=4pt,
  titlerule=0pt,
  before skip=6pt,
  after skip=6pt
}

\begin{table}[!ht]
\centering
\small
\resizebox{\linewidth}{!}{%
\begin{tabular}{llcccc}
\toprule
\multirow{2}{*}{$N$} & \multirow{2}{*}{Resolution} 
& \multicolumn{4}{c}{Images per semantic category} \\
\cmidrule(lr){3-6}
 &  & Conflicting & Irrelevant & Aligned & Total \\
\midrule
0 & $224{\times}224$   
& -- & -- & -- & 4,050 \\
\midrule
\multicolumn{6}{c}{\textit{Visual distractor}} \\
\midrule
1 & $1024{\times}1024$ & 4,050 & 4,050 & 4,050 & 12,150 \\
2 & $1024{\times}1024$ & 4,050 & 4,050 & 4,050 & 12,150 \\
3 & $1024{\times}1024$ & 4,050 & 4,050 & 4,050 & 12,150 \\
4 & $1024{\times}1024$ & 4,050 & 4,050 & 4,050 & 12,150 \\
\midrule
\multicolumn{6}{c}{\textit{Linguistic distractor: Typographic}} \\
\midrule
1 & $1024{\times}1024$ & 32,400 & -- & -- & 32,400 \\
2 & $1024{\times}1024$ & 32,400 & -- & -- & 32,400 \\
3 & $1024{\times}1024$ & 32,400 & -- & -- & 32,400 \\
4 & $1024{\times}1024$ & 32,400 & -- & -- & 32,400 \\
\bottomrule
\end{tabular}%
}
\caption{\textbf{Dataset statistics of \textit{Idis-perception}}. The no-distractor setting corresponds to the original ImageNet-9 split. Visual distractors are generated for each semantic category and distractor number. Typographic distractors are constructed by rendering the class names of non-target categories directly into the image.}
\label{tab:Idis-perception_statistic}
\end{table}

\begin{table}[!ht]
\centering
\small
\resizebox{\linewidth}{!}{%
\begin{tabular}{lccccccc}
\toprule
\multirow{2}{*}{$N$}
& \multicolumn{6}{c}{Samples per distractor category}
& \multirow{2}{*}{Total} \\
\cmidrule(lr){2-7}
& Aligned & Conflicting & Irrelevant & Handwritten & MathWriting & Textual & \\
\midrule
0 & -- & -- & -- & -- & -- & -- & 3,152 \\
\midrule
\multicolumn{8}{c}{\textit{Visual distractor}} \\
\midrule
1 & 2,684 & 7,676 & 3,152 & -- & -- & -- & 13,512 \\
2 & 2,684 & 7,676 & 3,152 & -- & -- & -- & 13,512 \\
3 & 2,684 & 7,676 & 3,152 & -- & -- & -- & 13,512 \\
4 & 2,684 & 7,676 & 3,152 & -- & -- & -- & 13,512 \\
\midrule
\multicolumn{8}{c}{\textit{Linguistic distractor: Typographic}} \\
\midrule
1 & -- & -- & -- & 3,152 & 3,152 & -- & 6,304 \\
2 & -- & -- & -- & 3,152 & 3,152 & -- & 6,304 \\
3 & -- & -- & -- & 3,152 & 3,152 & -- & 6,304 \\
4 & -- & -- & -- & 3,152 & 3,152 & -- & 6,304 \\
\midrule
\multicolumn{8}{c}{\textit{Linguistic distractor: Textual}} \\
\midrule
1 & -- & -- & -- & -- & -- & 3,152 & 3,152 \\
2 & -- & -- & -- & -- & -- & 3,152 & 3,152 \\
3 & -- & -- & -- & -- & -- & 3,152 & 3,152 \\
4 & -- & -- & -- & -- & -- & 3,152 & 3,152 \\
\bottomrule
\end{tabular}%
}
\caption{\textbf{Dataset statistics of \textit{Idis-math}}. The no-
distractor setting corresponds to the original MathVerse. Visual distractors are organized by their semantic relationship to the target class and distractor number. Textual distractors are inserted into the question prompt rather than the image.}
\label{tab:Idis-math_statistic}
\end{table}
\begin{table*}[ht]
\centering
\small
\renewcommand{\arraystretch}{1.2}
\setlength{\tabcolsep}{8pt}
\resizebox{0.95\linewidth}{!}{
\begin{tabular}{l cccc}
\toprule
& \multicolumn{4}{c}{\textbf{Objects}} \\
\cmidrule(lr){2-5}
\textbf{Class} & Object 1 & Object 2 & Object 3 & Object 4 \\
\midrule
\textbf{Dog} & dog bone chew toy & dog bowl & tennis ball & kennel \\
\textbf{Bird} & birdcage & nest & feather & bird feeder \\
\textbf{Wheeled Vehicle} & tire & steering wheel & license plate & bumper \\
\textbf{Reptile} & terrarium rock & heat lamp & log hideout & shed skin \\
\textbf{Carnivore} & toy fang & fake blood stain & fake chunk of meat & toy skeletal animal carcass \\
\textbf{Insect} & trash bag & empty net designed for catching insects & fruit peel & flower \\
\textbf{Musical Instrument} & chalkboard with music notes & music stand & metronome & sheet music \\
\textbf{Primate} & patch of jungle foliage & banana & coconut & vine \\
\textbf{Fish} & fishing rod & large empty nylon fishing net & life jacket & aquarium coral ornament \\
\midrule
\textbf{Irrelevant} & umbrella & clock & TV & suitcase \\
\bottomrule
\end{tabular}
}
\caption{\textbf{List of defined objects per class.} Each semantic class is associated with four representative objects used in distractor generation.}
\label{tab:class_object_mapping}
\end{table*}

\section{Pipeline details of Idis} \label{sec:suppl_A}
\subsection{Dataset statistic}

We summarize the composition of \textit{Idis} in \Cref{tab:Idis-perception_statistic,tab:Idis-math_statistic}. 
The benchmark consists of two task families, \textit{Idis-perception} and \textit{Idis-math}.

\noindent\textbf{Idis-perception.}
\textit{Idis-perception} is built from 4,050 images in the original split of ImageNet-9. There are a total of nine classes in the dataset (Precisely, the classes are: bird, carnivore, dog, fish, insect, instrument, primate, reptile, and vehicle.), with 450 samples for each class. The nine classes are the coarse-grained superclasses for the ImageNet classes determined based on the WordNet hierarchy. We use these images as the no-distractor baseline, where each image contains a single salient target object at a resolution of $224 \times 224$. 
For visual-object distractors, we vary the number of distractors from $N=1$ to $N=4$. 
For each distractor number and each semantic category---aligned, conflicting, and irrelevant---we generate one augmented image for every base image, resulting in 4,050 images per semantic category and 12,150 images per distractor-number setting. 
All distractor-augmented images are generated at a resolution of $1024 \times 1024$. 
The typographic distractor condition inserts non-target class-name text into the image, producing eight augmented variants for each sample, one for each non-target class.

\noindent\textbf{Idis-math.}
\textit{Idis-math} is built from MathVerse and contains both visual and linguistic distractor conditions. MathVerse consists of 2,612 high-quality mathematical questions, each paired with variants that provide different degrees of multimodal information. As our base dataset, we use the Test Mini split under four settings—Text Dominant, Text Lite, Vision Intensive, and Vision Dominant—which contain approximately 3,152 samples.
We consider six distractor conditions: aligned, conflicting, irrelevant, handwritten, mathwriting, and textual. The first three correspond to visual-object distractors, whereas handwritten, mathwriting, and textual distractors are treated as linguistic distractors.

For visual distractors, we vary the distractor density from $N=1$ to $N=4$. 
The resulting visual subset contains 40,536 augmented question-image pairs in total: 2,684 aligned, 7,676 conflicting, 3,152 irrelevant samples. 
To construct irrelevant visual distractors, we use LogicVista \citep{xiao2024logicvista} as an external source pool. LogicVista \citep{xiao2024logicvista} evaluates the logical reasoning abilities of VLMs across spatial, deductive, inductive, numeric, and mechanical reasoning, and contains 448 visual multiple-choice questions. We use the the table-capability subset as the seed pool. 

Linguistic distractors fall into two types: typographic and textual.
Typographic distractors are rendered directly into the image. Specifically, we insert handwritten mathematical expressions \citep{humynlabs_english_handwritten_math_notes, gervais2025mathwriting}. 
For textual distractors, we keep the original image unchanged and insert answer-irrelevant sentences into the question prompt. Following \citet{inverse_scaling_ttc}, we generate textual distractors with Sonnet 4.5 \citep{anthropic2025claudesonnet45} using the prompt in \Cref{tab:system_prompt_textual_distractor}, and insert them into the question prompt as natural-language statements.


\subsection{Distractor lists and pipeline} \label{sec:suppl_A_1}

\vspace{0.05in}
\noindent\textbf{Defined object.}
As shown in~\cref{tab:class_object_mapping}, we define a set of class-relevant objects for each class, as well as all class-irrelevant objects. For class-relevant objects, we first prompted GPT-5 to generate candidate items associated with each category using the following prompt.

\begin{tcolorbox}[boxsep=0pt,colback=black!5]
\textbf{\textit{Prompt}}: What objects are typically aligned with \{class\}?  
I'm curious about which objects commonly co-occur with \{class\} in images.  
Please focus on objects that are suitable for segmentation.
\end{tcolorbox}
Through human evaluation, these candidates were filtered and refined, and the four most appropriate objects were finally selected.
For the irrelevant category, we followed the same procedure, using object candidates from the MS-COCO (COCO 2017) dataset.

\subsection{Human validation and statistics} \label{sec:suppl_A_2}
Dataset construction was carried out with human-in-the-loop validation at every stage. Generated images were manually inspected, and images that did not satisfy our quality criteria were regenerated and re-evaluated by human annotators. We repeated this iterative validation process until all examples in the final dataset met the required standards. We summarize the validation criteria and the statistics of filtered cases below.

\vspace{0.05in}
\noindent\textbf{Idis-perception.}
(a) Criteria
\begin{itemize}[leftmargin=*,noitemsep, topsep=0pt]
\item The target object disappears or is not preserved.
\item The distractor differes from the intended concept.
\item An object from other class is generated.
\item The distractor occludes the target object.
\item Class-indicative text appears in the image and could serve as a hint.
\end{itemize}

(b) Statistics
\begin{itemize}[leftmargin=*,noitemsep, topsep=0pt]
\item Regeneration cases: 367/48,600 (0.76\%)
\end{itemize}

\vspace{0.05in}
\noindent\textbf{Idis-math.}
To construct aligned and conflicting visual distractors, we categorize each source diagram into a coarse mathematical concept family (e.g., triangle, quadrilateral, circle, or 3D shape). Labels are automatically generated and human-verified. For visual distractor placement, we use deterministic image compositing. The original image is kept unchanged, a blank canvas is appended to its right, and distractors are placed only within this added region. Thus, adding distractors does not remove or obscure information required to solve the problem.

\subsection{Prompts for dataset generation}
To generate each distractor-conditioned sample in the Idis-perception, we provide the image generation model with a structured text prompt describing the target class and the distractor type to be inserted. The generic prompt template is shown below:

\begin{tcolorbox}[boxsep=0pt,colback=black!5]
\textbf{\textit{Prompt}}: Using the provided image of a \{Class\},
please add \{Aligned / Conflicting / Irrelevant Distractors\}
to the scene. Ensure the changes are seamlessly integrated
into the natural setting. Do not modify or obscure the \{Class\}.
\end{tcolorbox}

For example, if the target class is \texttt{dog} and the distractor category is the conflicting class \texttt{bird}, the instantiated prompt becomes:

\begin{tcolorbox}[boxsep=0pt,colback=black!5]
\textbf{\textit{Prompt}}: Using the provided image of a \textbf{dog},
please add \textbf{birdcage}, \textbf{nest}, \textbf{feather}, \textbf{bird feeder}
to the scene. Ensure the changes are seamlessly integrated
into the natural setting. Do not modify or obscure the \textbf{dog}.
\end{tcolorbox}


\begin{table}[!ht]
\centering
\scriptsize
\setlength{\tabcolsep}{3pt}
\renewcommand{\arraystretch}{0.92}
\begin{tabular}{@{}p{0.54\columnwidth}p{0.14\columnwidth}p{0.25\columnwidth}@{}}
\toprule
\textbf{Comparison} & \textbf{Pearson $r$} & \textbf{Notes} \\
\midrule
Ours vs DeepSeek-V3.2-Exp\newline{}+ alternative prompt & 0.803 & Prompt sensitivity \\
Ours vs Qwen3.5-27B & 0.785 & Alternative LLM \\
Ours vs Strict lexical matcher & 0.476 & Non-LLM baseline \\
\bottomrule
\end{tabular}
\caption{\textbf{Robustness of attribute extraction across alternative extractors.}}
\label{tab:attribute_extractor_robustness}
\end{table}
\begin{table*}[ht]
\centering
\begin{tabular}{p{0.95\textwidth}}
\toprule
\textbf{System prompt for textual distractor generation}\\
\midrule

\begin{minipage}{\linewidth}
You are a math education expert. Your task is to insert distractor sentences into a math question to test whether a solver can ignore irrelevant information and focus on what actually matters.\\[6pt]

I have a math question with an accompanying figure. I want you to add \texttt{\{num\_distractors\}} distractor sentences into the question text. The distractors should be inserted between the setup/given conditions and the final question, blending naturally into the text.\\[6pt]

\texttt{<original\_question>}\\
\texttt{\{question\}}\\
\texttt{</original\_question>}\\[6pt]

\texttt{<correct\_answer>}\\
\texttt{\{answer\}}\\
\texttt{</correct\_answer>}\\[6pt]

\texttt{<subject>}\\
\texttt{\{subject\} / \{subfield\}}\\
\texttt{</subject>}\\[6pt]

\textbf{Rules:}
\begin{enumerate}
    \item Distractors MUST be mathematically related to the question's domain, such as geometry or algebra, but MUST NOT affect the correct answer.
    \item Distractors should introduce plausible-sounding information, such as extra angles, lengths, relationships, theorems, or observations, that seems relevant but is actually irrelevant to solving the problem.
    \item The distractors should reward careful thinking about what information is actually needed.
    \item The correct answer MUST remain exactly \texttt{\{answer\}} after inserting distractors.
    \item Distractors should blend naturally into the question text; they should read as if they were always part of the problem statement.
    \item Do NOT change, remove, or rephrase the original sentences. Only INSERT new distractor sentences.
    \item After generating, verify that the answer is still \texttt{\{answer\}} by mentally solving the augmented question. If it is not, regenerate.
    \item Output ONLY the complete augmented question with distractors inserted. Do not output the answer, explanation, tags, or anything else.
    \item Keep the Choices section exactly as-is at the end.
\end{enumerate}

\textbf{Instruction:} Only output the complete augmented question. No explanations or extra text.
\end{minipage}\\

\bottomrule
\end{tabular}
\caption{\textbf{System prompt for textual distractor generation in Idis-math.}}
\label{tab:system_prompt_textual_distractor}
\end{table*}

\section{Implementation details} \label{sec:suppl_B}

\subsection{Experimental protocol} \label{sec:suppl_B-1}

\vspace{0.05in}
\noindent\textbf{Extract area of each object.}
To quantify the spatial prominence of visual entities, we estimate the area of both target and distractor regions using text-conditioned segmentation. We employ LangSAM to produce binary masks conditioned on class-level textual prompts (e.g., ``feather'', ``birdcage''), allowing predicted mask labels to be matched against either the target class or the distractor object list. All masks whose labels belong to the target class are merged via pixelwise union into a single target mask, and likewise, all masks whose labels correspond to the distractor object list are merged into a single distractor mask. The area of each region is computed as the number of pixels in the resulting aggregated mask, which we use in downstream analyses such as distractor–target area ratios and their effects on accuracy and reasoning length.

\vspace{0.05in}
\noindent\textbf{Attribute extraction procedure.}
To characterize how visual evidence is used within the model's reasoning process, we extract fine-grained visual attributes directly from the generated reasoning traces. For each trace, we provide the full text together with structured, class-aware instructions to a specialized large language model (DeepSeek-V3.2-Exp) (See~\cref{tab:system_prompt_classwise,tab:system_prompt_mathverse_attribute_extraction}). The model is guided to operate purely as an evidence extractor: it must identify literal words or phrases in the reasoning text that denote observable attributes, or class-related features.
For Idis-perception, we define 10 attribute categories corresponding to the nine semantic classes plus an `other' category. For Idis-math, we instead define three attribute categories: target-related, distractor-related, and other.
To ensure consistency, the extractor follows a small set of rules. It only selects literal words or phrases that explicitly appear in the reasoning trace, without paraphrasing or inference. Each extracted phrase is then assigned to exactly one of the ten predefined categories. Multi-word expressions such as long tail are treated as a single attribute.

An analogous extraction procedure is conducted for predictions on the Waterbirds dataset. In this case, we use a separate set of instruction prompts tailored to the ecological and environmental cues that are central to Waterbirds, as shown in~\cref{tab:system_prompt_bio_env}. 

\vspace{0.05in}
\noindent\textbf{Attention blocking.} We first generate the complete reasoning trace without intervention and identify the tokens corresponding to target- or distractor-related attributes. We then directly generate the final answer without producing any additional reasoning tokens. During final-answer generation, we block attention from the answer tokens to the selected attribute-token positions in the preceding reasoning trace. The intervention is applied across all transformer layers and all attention heads, with the same mask broadcast across heads. This intervention blocks only direct attention to the selected tokens; attribute information may already be distributed across other token representations.

\subsection{Metrics.} \label{ssec:suppl_B_metric}
To quantify how visual distractors affect the model's reasoning behavior, we define two metrics as follows.

\textit{Distractor Attribute Ratio} measures the proportion of extracted attributes that are allocated to distractors in the reasoning trace:
\begin{equation}
r_{\mathrm{attr}}
=
\frac{\#\text{Distractor Attributes}}{\#\text{Total Attributes}}.
\end{equation}

\textit{Distractor Area Ratio} quantifies distractor spatial size relative to the target:
\begin{equation}
r_{\mathrm{area}}
=
\frac{A_{\mathrm{dist}}}{A_{\mathrm{dist}} + A_{\mathrm{target}}}.
\end{equation}

Here, $A_{\mathrm{dist}}$ denotes the distractor area, and $A_{\mathrm{target}}$ denotes the target object area.

\subsection{Hardware}
All inference experiments were conducted on a single NVIDIA RTX A6000 and RTX A6000 Ada GPU using BF16 precision.

\subsection{Prompts} \label{ssec:suppl_B-3}
For Idis-math evaluation, we follow the VLMEvalKit \citep{duan2024vlmevalkit} protocol and use Qwen3.5-27B \citep{qwen3.5} as the judge model. VLMEvalKit\citep{duan2024vlmevalkit} is a widely adopted evaluation framework for visual mathematical reasoning and serves as the de facto evaluation standard for MathVerse, on which Idis-Math is built~\citep{tian2026more,wang2026monet,zhan2026vision}

\vspace{0.05in}
\noindent\textbf{Inference prompt.}
We employ different inference VQA task prompts depending on the dataset:

\begin{prompt_for_data}{Prompts for Idis-perception}
\begin{itemize}[leftmargin=1.2em, itemsep=1pt, topsep=2pt]
    \item \textbf{VQA Task prompt:} 
    [Question] Which category best describes the main object in the image? 
    Choose exactly one from: Dog, Bird, Vehicle, Reptile, Carnivore, Insect, Instrument, Primate, Fish. 
    Use a thinking process to analyze the problem step-by-step. 
    At the end, provide your answer and clearly indicate it using 
    $<answer>X</answer>$ format.


    \item \textbf{Prompt strategy:} 
    [Question] Which category best describes the main object in the image? 
    Choose exactly one from: Dog, Bird, Vehicle, Reptile, Carnivore, Insect, Instrument, Primate, Fish. 
    First identify the main object. Base your reasoning only on that object's own visual attributes, 
    and determine which of the nine categories best matches the object. 
    Do not assume that the most visually dominant object is the main object. 
    Do not infer the category from surrounding objects or from items that are merely associated with another category. 
    At the end, provide your answer as $<answer>X</answer>$ format.
\end{itemize}
\end{prompt_for_data}


\begin{prompt_for_data}{Prompts for Waterbirds}
\begin{itemize}[leftmargin=1.2em, itemsep=1pt, topsep=2pt]
    \item \textbf{Reasoning model prompt:} 
    [Question] Is the bird in the image a waterbird or a landbird? 
    Use a thinking process to analyze the problem step-by-step. 
    At the end, provide your answer and clearly indicate it using 
    $<answer>X</answer>$ format.

    \item \textbf{Prompt strategy:} 
    [Question] Is the bird in the image a waterbird or a landbird? 
    Think step by step based on the foreground bird's attributes. 
    At the end, select your answer from the provided options and clearly indicate it using 
    $<answer>X</answer>$ format.
\end{itemize}
\end{prompt_for_data}

\vspace{0.05in}
\noindent\textbf{Attribute extraction prompt.}
We use two types of extraction prompts, depending on the dataset:

\vspace{-0.1in}
\begin{itemize}[leftmargin=*,noitemsep, topsep=0pt]
\item Idis attribute extraction: The class-aware attribute extraction prompt shown in~\cref{tab:system_prompt_classwise}.
\item Waterbirds attribute extraction: The biological/environmental attribute extraction prompt shown in~\cref{tab:system_prompt_bio_env}.
\end{itemize}

\subsection{Validation of the attribute-extraction pipeline.} \label{sec:suppl_B-parser}
we evaluated the robustness of our findings across alternative extraction pipelines. Our extracted scores show strong correlation with both (1) an alternative prompting strategy (Pearson r = 0.803; see \cref{tab:attribute_extractor_robustness} for the prompt) and (2) a different base LLM, Qwen3.5-27B \citep{qwen3.5} (r = 0.785). In contrast, a strict lexical matching baseline yields substantially weaker correlation (Pearson r = 0.476), suggesting that simple word matching is insufficient for this analysis. Taken together, these results indicate that our findings are robust to reasonable variations in the extraction pipeline while still requiring semantic parsing beyond a non-LLM baseline.

\begin{prompt_for_data}{Prompts for Waterbirds}
\begin{itemize}[leftmargin=1.2em, itemsep=1pt, topsep=2pt]
    \item \textbf{Alternative prompt:} 
    You are analyzing a chain-of-thought reasoning trace.
    
    Task: Extract every concrete noun phrase that literally appears in the text, and classify each into one of these 10 categories based on what the noun phrase typically refers to:
      dog, bird, vehicle, reptile, carnivore, insect, instrument, primate, fish, other.
    
    The "other" category is for noun phrases that don't naturally belong to any of the nine main classes.
    
    Rules:
    - Only use words/phrases that literally appear in the text. Do not paraphrase or infer.
    - Multi-word noun phrases count as one item (e.g., "long tail" -> one attribute).
    - "Dog", "bird", and other class words themselves count if they literally appear.
    - Each extracted item must be assigned to exactly one of the 10 categories.
    
    Expected JSON output: [SAME with original prompt]

\end{itemize}
\end{prompt_for_data}

\clearpage
\begin{table}[ht]
\centering
\begin{tabular}{p{0.95\textwidth}}
\toprule
\textbf{System prompt for class-wise attribute extraction}\\
\midrule

You are an expert in analyzing a model's chain-of-thought. Extract literal evidence words or phrases from the text and classify them into 10 categories: nine main classes (\texttt{dog}, \texttt{bird}, \texttt{vehicle}, \texttt{reptile}, \texttt{carnivore}, \texttt{insect}, \texttt{instrument}, \texttt{primate}, \texttt{fish}) and one ``\texttt{other}'' category for anything else. For each main class, include attributes or objects directly related to it, considering morphology, taxonomy, features, shape, size, or adaptations.\\[6pt]

\textbf{Representative related objects:}
\begin{itemize}
    \item \textbf{dog\_attributes:} dog bone chew toy, dog bowl, tennis ball, kennel
    \item \textbf{bird\_attributes:} birdcage, nest, feather, bird feeder
    \item \textbf{vehicle\_attributes:} tire, steering wheel, license plate, bumper
    \item \textbf{reptile\_attributes:} terrarium rock, heat lamp, log hideout, shed skin
    \item \textbf{carnivore\_attributes:} fang, blood stain, chunk of meat, skeletal animal carcass
    \item \textbf{insect\_attributes:} trash bag, insect net, fruit peel, flower
    \item \textbf{instrument\_attributes:} chalkboard with music notes, music stand, metronome, sheet music
    \item \textbf{primate\_attributes:} jungle foliage, banana, coconut, vine
    \item \textbf{fish\_attributes:} fishing rod, fishing net, life jacket, aquarium coral ornament
    \item \textbf{other\_attributes:} unrelated attributes or objects (e.g., umbrella, clock, TV, suitcase)
\end{itemize}

\textbf{Rules:}
\begin{enumerate}
    \item Use only literal words/phrases from the text (case-insensitive match for listed objects).
    \item Multi-word phrases (e.g., ``long tail'') count as one attribute.
    \item Do not infer or paraphrase.
    \item ``Taxonomic labels'' like ``bird'' or ``dog'' are valid only if they literally appear.
    \item Each extracted attribute must belong to exactly one of the ten categories.
\end{enumerate}

\textbf{Expected JSON output:}\\[-2pt]
\begin{minipage}{\linewidth}\ttfamily
\{\\
\ \ "dog\_attributes": [...],\\
\ \ "bird\_attributes": [...],\\
\ \ "vehicle\_attributes": [...],\\
\ \ "reptile\_attributes": [...],\\
\ \ "carnivore\_attributes": [...],\\
\ \ "insect\_attributes": [...],\\
\ \ "instrument\_attributes": [...],\\
\ \ "primate\_attributes": [...],\\
\ \ "fish\_attributes": [...],\\
\ \ "other\_attributes": [...],\\
\ \ "counts": \{ "dog": <int>, "bird": <int>, "vehicle": <int>, "reptile": <int>, "carnivore": <int>, "insect": <int>, "instrument": <int>, "primate": <int>, "fish": <int>, "other": <int> \}\\
\}
\end{minipage}

\vspace{8pt}
\textbf{Instruction:} Only output the JSON object. No explanations or extra text.\\
\bottomrule
\end{tabular}
\caption{\textbf{System prompt for Idis-perception.}}
\label{tab:system_prompt_classwise}
\end{table}

\clearpage
\begin{table}[ht]
\centering
\normalsize
\setlength{\tabcolsep}{4pt}
\renewcommand{\arraystretch}{1.0}
\begin{tabular}{p{0.95\textwidth}}
\toprule
\textbf{System prompt for target--distractor attribute extraction}\\
\midrule

You are analyzing a model's mathematical visual reasoning trace for the
MathVerse conflicting setting, where a TARGET geometric figure (the figure the
original question is actually about) coexists with one or more DISTRACTOR
figures (extra shapes inserted into the image that are NOT the subject of the
question).\\[3pt]

The user message provides four pieces of context for this sample:\\[2pt]
\begin{tabular}{@{}ll@{}}
\textbf{TARGET shapes:}     & \texttt{<comma-separated shape names>} \\
\textbf{DISTRACTOR shapes:} & \texttt{<comma-separated shape names>} \\
\textbf{QUESTION:}          & \texttt{<the original problem statement, text-only version>} \\
\textbf{Raw text:}          & \texttt{a single sentence from the model's reasoning trace.}
\end{tabular}
  
Use the QUESTION as the authoritative description of the TARGET. Any entity, point, line, segment, angle, length, value, relation, or option in the QUESTION belongs to the TARGET context; when the same entity or a clear anaphoric reference appears in Raw text, classify it as \texttt{target\_related}.\\[3pt]

\textbf{Task:} Extract every concrete mathematical, diagrammatic, or visual attribute phrase that literally appears in Raw text, and assign each phrase to exactly one of \texttt{target\_related}, \texttt{distractor\_related}, or \texttt{other}.\\[3pt]

\textbf{Definitions:}
\begin{itemize}[leftmargin=*,nosep]
    \item \textbf{\texttt{target\_related}:} phrases referring to the TARGET shape; any entity, property, measurement, relation, value, or option mentioned in the QUESTION; or a part/property/operation clearly attached to the TARGET.
    \item \textbf{\texttt{distractor\_related}:} phrases referring to a DISTRACTOR shape or to a part/property/measurement/relation/operation explicitly tied to a DISTRACTOR in Raw text.
    \item \textbf{\texttt{other}:} concrete attribute phrases that cannot be confidently assigned to either TARGET or DISTRACTOR, including generic math terms, unrelated entities, or ambiguous mentions.
\end{itemize}

\textbf{Rules:}
\begin{enumerate}[leftmargin=*,nosep]
    \item Use only words/phrases that literally appear in Raw text; do not infer, translate, or paraphrase. The QUESTION is for grounding only.
    \item Assign \texttt{target\_related} if the phrase names or clearly refers to anything introduced by the QUESTION or to the TARGET shape. Anaphora counts only when unambiguous.
    \item Assign \texttt{distractor\_related} only when Raw text explicitly links the phrase to a DISTRACTOR shape.
    \item If ambiguous, assign \texttt{other}. Multi-word phrases count as one item.
    \item Keep duplicates only when they have meaningfully different wording.
    \item Do not extract generic discourse words such as ``step,'' ``approach,'' ``question,'' or ``answer'' unless tied to a concrete option or shape.
    \item Respond strictly as one JSON object and nothing else.
\end{enumerate}

\textbf{Expected JSON output:}\\[-4pt]
\begin{minipage}{\linewidth}\ttfamily\normalsize
\{\\
\ \ "target\_related": [...],\\
\ \ "distractor\_related": [...],\\
\ \ "other": [...],\\
\ \ "counts": \{\\
\ \ \ \ "target\_related": <int>,\\
\ \ \ \ "distractor\_related": <int>,\\
\ \ \ \ "other": <int>\\
\ \ \}\\
\}
\end{minipage}

\vspace{4pt}
\textbf{Instruction:} Only output the JSON object. No explanations or extra text.\\
\bottomrule
\end{tabular}
\caption{\textbf{System prompt for Idis-math.}}
\label{tab:system_prompt_mathverse_attribute_extraction}
\end{table}
\clearpage
\begin{table}[ht]
\centering
\begin{tabular}{p{0.95\textwidth}}
\toprule
\textbf{System prompt for bio/env attribute extraction}\\
\midrule

You are an expert in analyzing a model's chain-of-thought. Your job is to pull out the concrete evidence words or phrases the model itself cites and sort them into two buckets.\\[4pt]

\textbf{Buckets}
\begin{itemize}
    \item \textbf{bio attribute}: many morphological part, taxonomic label, features, adaptations or size/shape descriptor of the foreground object (e.g., wings, webbed feet, long legs, body shape, long tail, petrel).
    \item \textbf{env attribute}: physical background or habitat terms that locate the scene (e.g., forest path, reeds, lake, ocean, coastal zone, sky, sand).
\end{itemize}

\textbf{Respond strictly in this JSON format:}\\[-2pt]
\begin{minipage}{\linewidth}\ttfamily
\{\\
\ \ "bio\_attributes":\ [ \ldots ],\\
\ \ "env\_attributes":\ [ \ldots ],\\
\ \ "bio\_count":\ \ \ \ \ \textless integer\textgreater,\\
\ \ "env\_count":\ \ \ \ \ \textless integer\textgreater\\
\}
\end{minipage}

\vspace{6pt}
\textbf{Rules for extracting attributes}
\begin{enumerate}
    \item A multi-word phrase like ``long neck'' counts as one attribute.
    \item Do not invent attributes; use only words or phrases literally present in the model output.
\end{enumerate}

\textbf{Examples}\\[2pt]

\textbf{Example Input 1:}\\
\emph{``The bird has a thick body, similar to a juvenile albatross, which are seabirds adapted to marine environments. They spend most of their time at sea and rely on oceanic ecosystems.''}\\
\textbf{Output:}
\begin{minipage}{\linewidth}\ttfamily
\{"bio\_attributes":\ ["thick body","juvenile albatross","seabirds","adapted to marine environments"],\ "env\_attributes":\ ["sea","oceanic ecosystems"],\ "bio\_count":\ 4,\ "env\_count":\ 2\}
\end{minipage}

\vspace{4pt}
\textbf{Example Input 2:}\\
\emph{``The image shows a small animal with a light-colored face, dark eyes, and a body that's mostly light brown or beige. It has a small head with pointed ears, and its front paws are visible.''}\\
\textbf{Output:}
\begin{minipage}{\linewidth}\ttfamily
\{"bio\_attributes":\ ["light-colored face","dark eyes","light brown","beige","small head","pointed ears","front paws"],\ "env\_attributes":\ [],\ "bio\_count":\ 7,\ "env\_count":\ 0\}
\end{minipage}

\vspace{4pt}
\textbf{Example Input 3:}\\
\emph{``The background has a body of water, like a pond or lake, and the bird is near that. Also, waterbirds often have adaptations for aquatic life, like webbed feet (though here it's a statue, but the context). The setting with water suggests it's a waterbird.''}\\
\textbf{Output:}
\begin{minipage}{\linewidth}\ttfamily
\{"bio\_attributes":\ ["adaptations for aquatic life","webbed feet","waterbird"],\ "env\_attributes":\ ["background","body of water","pond","lake","water"],\ "bio\_count":\ 3,\ "env\_count":\ 5\}
\end{minipage}

\vspace{8pt}
\textbf{Input:}\\
\emph{\{model output text here\}}\\[2pt]
\textbf{Output:} \{JSON object only; no additional text\}\\
\bottomrule
\end{tabular}
\caption{\textbf{System prompt for Waterbirds dataset.}}
\label{tab:system_prompt_bio_env}
\end{table}

\clearpage
\section{Qualitative examples}
\definecolor{primary-orange}{HTML}{FF6B35}
\definecolor{secondary-orange}{HTML}{F7931E}
\definecolor{light-gray}{HTML}{F0F0EB}
\definecolor{white-bg}{HTML}{FAFAF7}
\definecolor{text-secondary}{HTML}{4A4A4A}
\definecolor{reasoning-bg}{HTML}{FFF5F0}
\definecolor{answer-bg}{HTML}{daf0ff}
\definecolor{correct-blue}{HTML}{b5e2ff}
\definecolor{incorrect-red}{HTML}{F44336}
\definecolor{incorrect-bg}{HTML}{FFEBEE}
\definecolor{prompt-bg}{HTML}{F0F0EB}
\definecolor{secondary-brown}{HTML}{EBDBBC}

\newtcolorbox{qualitativeexample}[1]{
  breakable,
  colback=white-bg,
  fonttitle=\bfseries,
  title=#1,
  boxsep=4mm
}
\newcommand{\codeblock}[1]{\begin{verbatim}#1\end{verbatim}}

\newtcolorbox{promptbox}{
  colback=prompt-bg,
  colframe=secondary-orange!50,
  boxrule=0.5pt,
  arc=1mm,
  boxsep=2pt,
  before=\par\smallskip,
  after=\par\smallskip
}

\newtcolorbox{correctanswerbox}[1][answer-bg]{
  colback=#1,
  colframe=correct-blue,
  boxrule=0.5pt,
  arc=1mm,
  boxsep=2pt,
  before=\par\smallskip,
  after=\par\smallskip
}

\newtcolorbox{reasoningbox}[1][reasoning-bg]{
  colback=#1,
  colframe=secondary-brown,
  boxrule=0.5pt,
  arc=1mm,
  boxsep=2pt,
  before=\par\smallskip,
  after=\par\smallskip
}

\newtcolorbox{incorrectanswerbox}[1][incorrect-bg]{
  colback=#1,
  colframe=incorrect-red,
  boxrule=0.5pt,
  arc=1mm,
  boxsep=2pt,
  before=\par\smallskip,
  after=\par\smallskip
}

\subsection{Idis-perception: Visual distractors}
\begin{strip}
\begin{qualitativeexample}{}
All qualitative examples in this section are generated by Qwen3-VL-8B-Thinking.
    \subsection*{Prompt}
    \begin{promptbox}
  \noindent
  \begin{minipage}[t]{0.55\linewidth}
    \vspace{0pt}
    \small{
[Question] Which category best describes the main object in the image? Choose exactly one from: Dog, Bird, Vehicle, Reptile, Carnivore, Insect, Instrument, Primate, Fish.\\
Use a thinking process to analyze the problem step-by-step.\\
At the end, provide your answer and clearly indicate it using \textless answer\textgreater X\textless /answer\textgreater\ format.}
  \end{minipage}%
  \hfill
  \begin{minipage}[t]{0.4\linewidth}
    \vspace{0pt}

  \centering
  \begin{minipage}[t]{0.5\linewidth}
    \includegraphics[width=\linewidth]{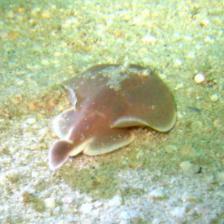}
  \end{minipage}%
  \hfill
  \begin{minipage}[t]{0.5\linewidth}
    \includegraphics[width=\linewidth]{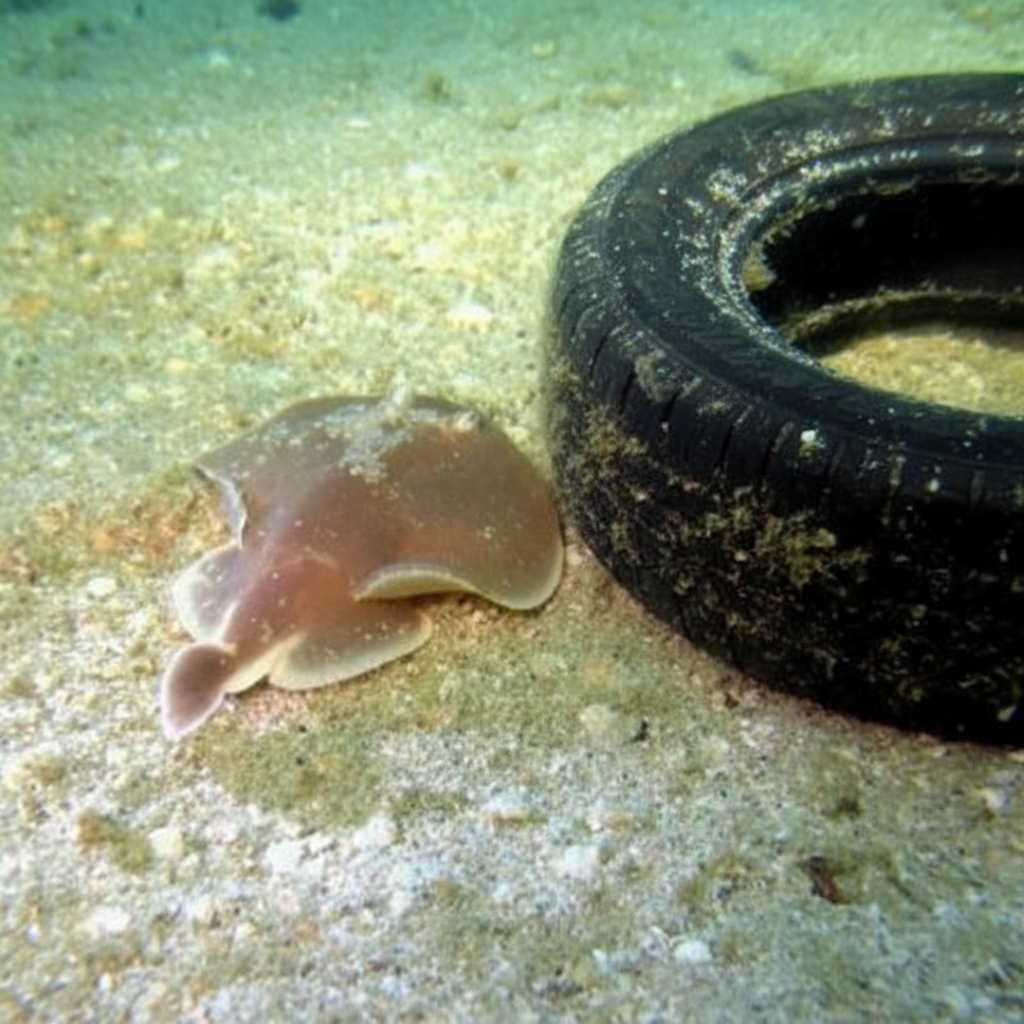}
  \end{minipage}

  \end{minipage}
    \end{promptbox}

\subsection*{Expected Answer}
Fish \\\\
(Left: no distractor. Right: adding one conflicting distractor from the vehicle class.)
    
    \hrulefill
    \vspace{1em}

    \subsubsection*{Response without distractor - Correct}
    \textbf{Model Output:}
    \begin{correctanswerbox}
    \small{So, let's try to figure out what the main object in the image is. ... \\
    It's \textbf{a flat, disc-shaped creature with a tail}. Let's look at the options: ... \\
     the creature in the image has a flat body, which is typical of rays. Rays are a type of fish, specifically cartilaginous fish (like sharks, but rays are different). \textbf{So the main object is a fish.} ...\\ 
     To determine the category of the main object in the image, follow this step - by - step analysis: ...\\
     \textbf{2. **Analyze the creature’s physical traits**}: The object has a flat, disc - shaped body with a tail, which is characteristic of **rays**. Rays are a type of **fish** (specifically, cartilaginous fish, related to sharks). ...\\
     **Confirm the best fit**: The creature’s morphology and habitat align with the definition of a fish, making this the only category that accurately describes the main object.
     \textbf{\textless answer\textgreater Fish\textless /answer\textgreater}
}
    \end{correctanswerbox}

    \hrulefill
    \vspace{1em}

    \subsubsection*{Response with one visual distractor - Incorrect}

    \textbf{Model Output:}
    \begin{incorrectanswerbox}
    \small{So, let's look at the image. The main objects are \textbf{a sea slug (or maybe a type of mollusk) and a tire.} ... \\
    First, identify the main object. \textbf{The tire is a vehicle part, but the other object is a sea creature.} ... \\
    Wait, maybe the tire is a vehicle part, but the question says main object. ...\\  So if the main object is the tire, then Vehicle. But is the tire the main object? \textbf{The image has both, but the tire is more prominent.} ...\\
   To determine the correct category, we analyze the image and the given options:  \\
   1. **Identify the main objects**: The image contains a sea slug (a marine invertebrate) and a tire.  ... \\
   so the tire (as a vehicle component) is the most relevant object for the given options. \\
         \textbf{\textless answer\textgreater Vehicle\textless /answer\textgreater}
}
    \end{incorrectanswerbox}
\end{qualitativeexample}
\end{strip}

\clearpage
\subsection{Attribute-level reasoning behavior}
\begin{strip}
\begin{qualitativeexample}{}
\small

\noindent
\begin{minipage}[t]{0.98\linewidth}
    \begin{promptbox}
        \centering
        \includegraphics[width=0.32\linewidth]{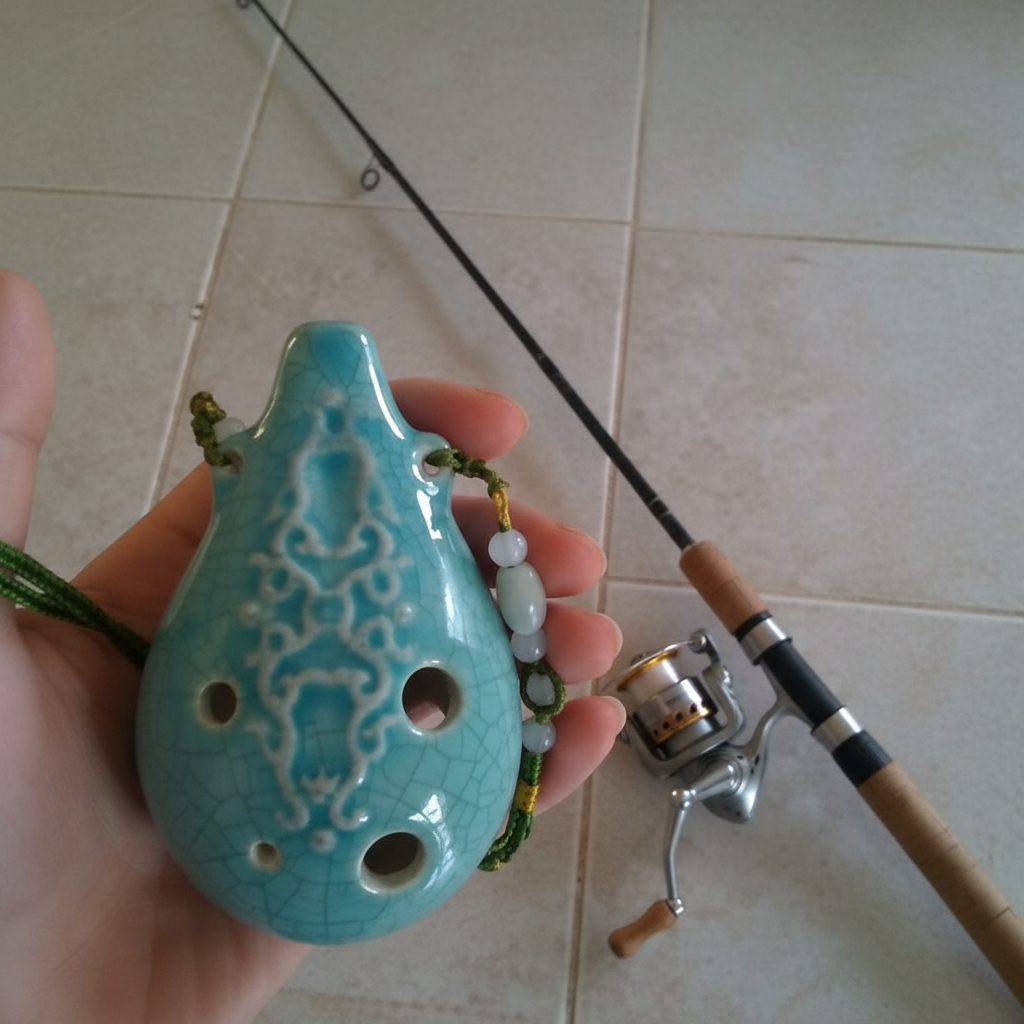}

        \vspace{0.5em}

        \raggedright
        \textbf{Example 1}

        \textbf{Target attributes.} instrument, ceramic object

        \textbf{Distractor attributes.} fishing rod, fishing, fish-shaped ornament

        \textbf{Correct answer.} Musical Instrument

        \textbf{Model output.}
        \itshape
        The image shows a hand holding a turquoise ceramic object with some holes and a design, and there's a fishing rod next to it. ... Wait, is the main object a fish? No ...But is it a real fish? No ... but maybe the question is considering the context of fishing. ... Instrument? Maybe ...

        \textbf{\textless answer\textgreater Fish\textless /answer\textgreater}

        \normalfont
        \textbf{Distractor.} fishing rod (from Fish class)
    \end{promptbox}
\end{minipage}

\vspace{0.8em}

\noindent
\begin{minipage}[t]{0.98\linewidth}
    \begin{promptbox}
        \centering
        \includegraphics[width=0.32\linewidth]{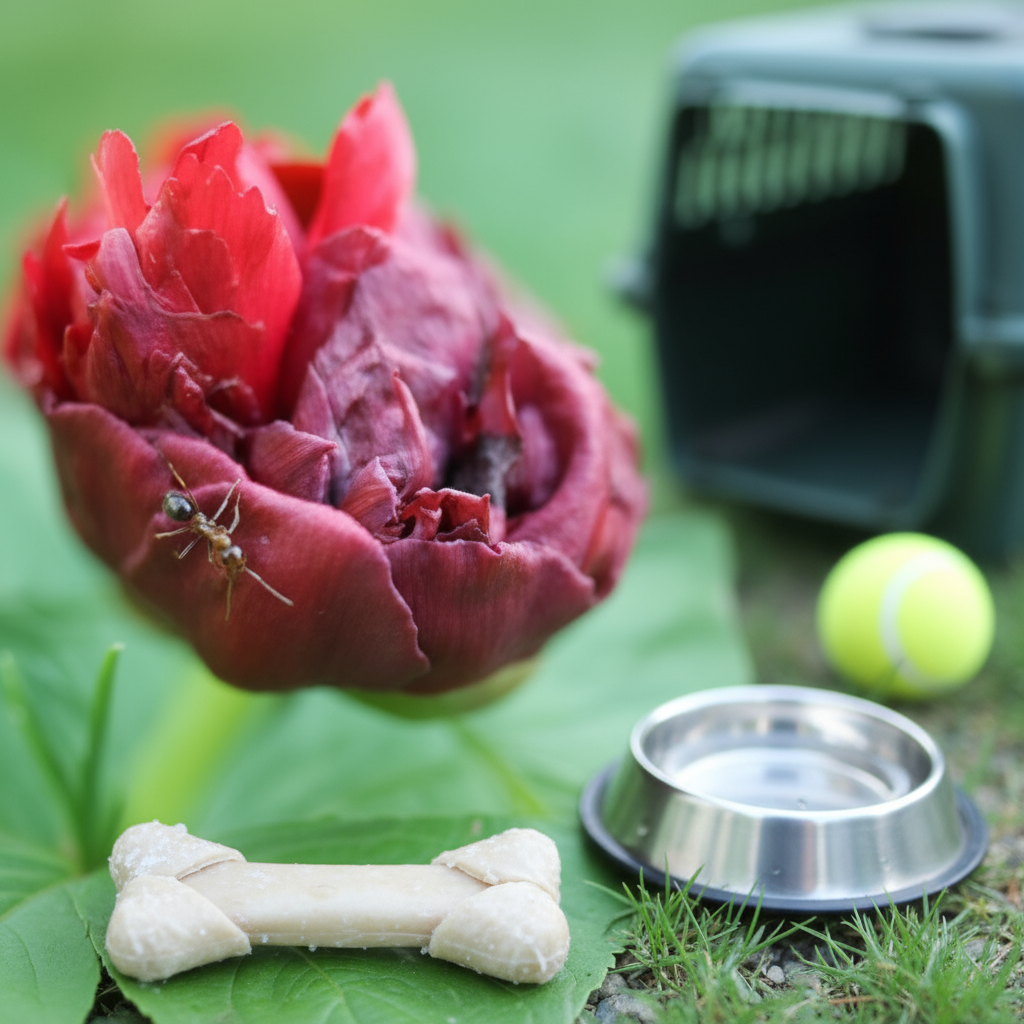}

        \vspace{0.5em}

        \raggedright
        \textbf{Example 2}

        \textbf{Target attributes.} flower, insect

        \textbf{Distractor attributes.} bone-shaped toy, dog bed, metal food bowl, tennis ball

        \textbf{Correct answer.} Insect

        \textbf{Model output.}
        \itshape
        So, let's look at the image. Wait, maybe the main object is the dog-related items, ... Wait, the image has a flower, ... Insect is unrelated ... The items in the image are all dog-related, so the category that best describes the primary subject (the context of the objects) is Dog.

        \textbf{\textless answer\textgreater Dog\textless /answer\textgreater}

        \normalfont
        \textbf{Distractor.} dog bone chew toy, dog bowl, tennis ball, kennel (from Dog class)
    \end{promptbox}
\end{minipage}

\vspace{0.8em}

\begin{promptbox}
These examples illustrate how Qwen3-VL-Thinking extracts attributes and reasons on Idis samples with visual distractors. When distractor-related attributes outnumber target-object attributes, the model's reasoning shifts toward distractor context and produces an incorrect answer. In particular, when distractors occupy a larger image area than the target object, the model sometimes predicts the distractor's class as the final label, highlighting biased attribute allocation toward distractors.
\end{promptbox}
\end{qualitativeexample}
\clearpage
\end{strip}

\clearpage
\noindent 

\begin{strip}
\subsection{Effect of prompt strategy}
\begin{qualitativeexample}{}
    \subsection*{Prompt}
    \begin{promptbox}
  \noindent
  \begin{minipage}[t]{0.55\linewidth}
    \vspace{0pt}
    \small{
[Question] Is the bird in the image a waterbird or a landbird? Use a thinking process to analyze the problem step-by-step. \\ 
At the end, provide your answer and clearly indicate it using \textless answer\textgreater X\textless /answer\textgreater\ format.}
  \end{minipage}%
  \hfill
  \begin{minipage}[t]{0.4\linewidth}
    \vspace{0pt}

  \centering
  \begin{minipage}[t]{0.5\linewidth}
    \includegraphics[width=\linewidth]{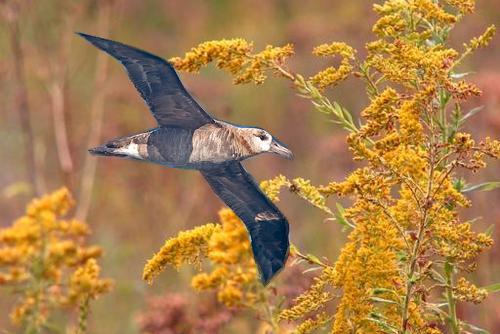}
  \end{minipage}%
  \end{minipage}
    \end{promptbox}

\subsection*{Expected Answer}
waterbird \\\\
(Bias-conflicting sample with land background)
    
    \hrulefill
    \vspace{1em}

    \subsubsection*{Response without prompt strategy - Incorrect}
    \textbf{Model Output:}
    \begin{incorrectanswerbox}
    \small{So, let's try to figure out if the bird in the image is a waterbird or a landbird. ... \\
the bird is flying near some yellow flowers, which are probably goldenrod (a common plant in meadows or fields). ... \\
the context of the habitat is important. Since the bird is in a field with flowers, not near water, that's a clue. ... \\
So maybe it's a landbird.  ... \\
 The bird looks like a Black-bellied Plover or maybe a similar shorebird, but wait, shorebirds are waterbirds. ...\\
 \textbf{Wait, the key is the habitat.} If the bird is in a terrestrial environment (land), then it's a landbird. ...\\
 \textbf{Since the bird is in a terrestrial habitat with no evidence of aquatic adaptation or proximity to water}, it is classified as a **landbird**.  \\
    \textbf{\textless answer\textgreater landbird\textless /answer\textgreater}

}
    \end{incorrectanswerbox}

    \hrulefill
    \vspace{1em}

    \subsubsection*{Response with prompt strategy - Correct}

    \textbf{Model Output:}
    \begin{correctanswerbox}
    \small{So, let's figure out if the bird is a waterbird or a landbird. ... \\
    Looking at the image: the bird is flying near yellow flowers (maybe goldenrod). ... \\
    Wait, maybe it's a tern? Terns are waterbirds. Wait, but the image shows it flying over land with flowers. Hmm. Wait, \textbf{maybe the key is the habitat.} ...\\
    Wait, \textbf{the question says ``based on the foreground bird's attributes".} ... \\
Wait, \textbf{let's check the bird's features.} The bird has a long, pointed beak, which is common in seabirds that catch fish. Also, the wings are long and narrow, which is typical for birds that fly over water. So maybe it's a waterbird. \\
Since the bird has the physical traits of a seabird (long wings, pointed beak for fishing), it's likely a waterbird.\\
So the answer should be waterbird. ...\\
Thus, \textbf{based on the bird’s physical attributes (long wings, pointed beak)} and typical ecological role, it is a waterbird. 

\textbf{\textless answer\textgreater waterbird\textless /answer\textgreater}
}
    \end{correctanswerbox}
\end{qualitativeexample}
\end{strip}


\begin{figure*}[t]
  \centering
  \begin{subfigure}{0.33\linewidth}
    \includegraphics[width=\linewidth]{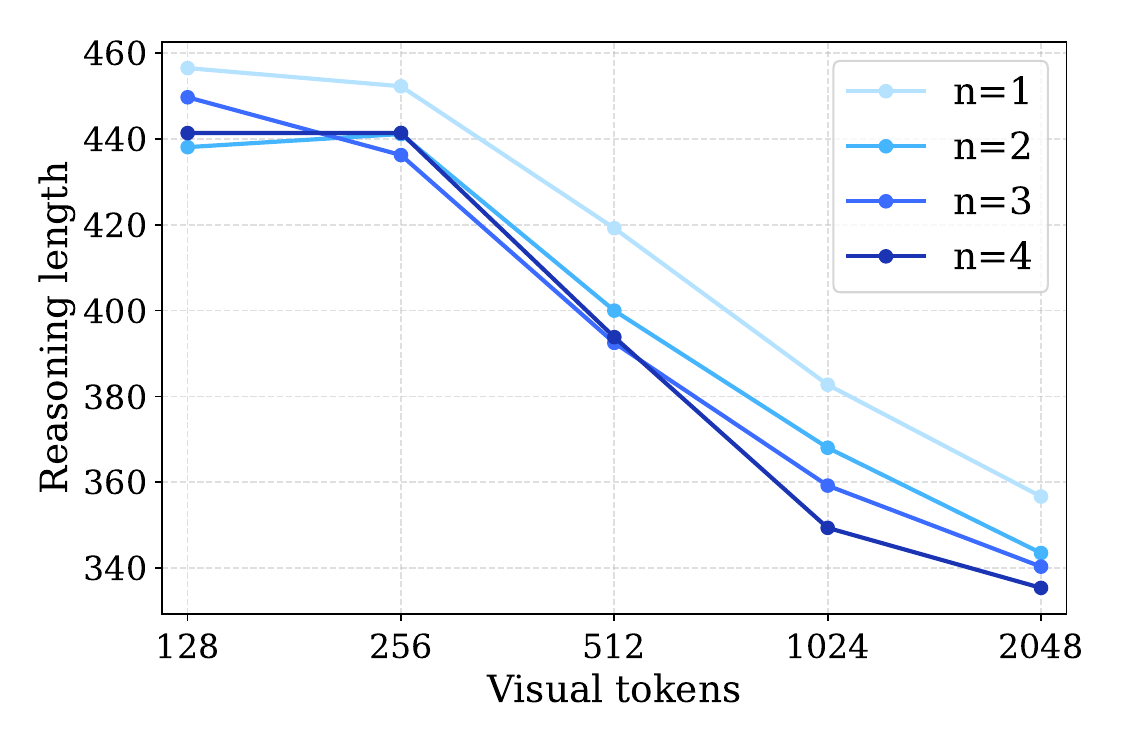}
    \subcaption{Visual tokens \& Reasoning Length}\label{fig:resolution-a}
  \end{subfigure}\hfill
  \begin{subfigure}{0.33\linewidth}
    \includegraphics[width=\linewidth]{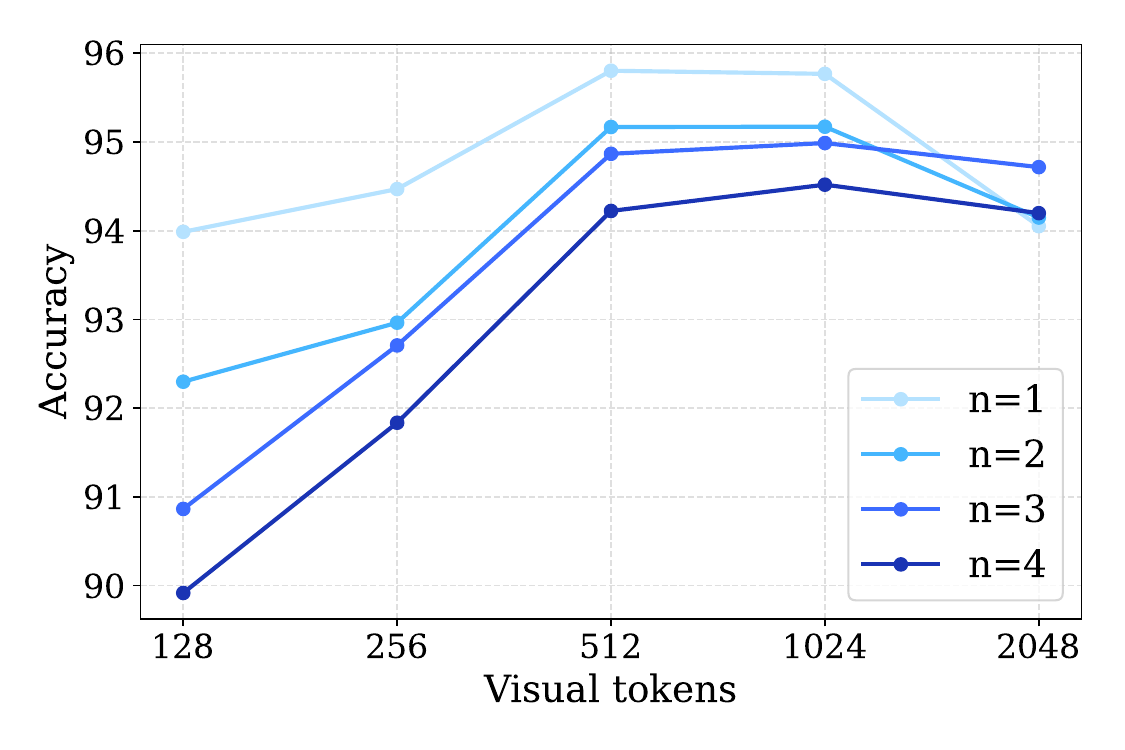}
    \subcaption{Visual tokens \& Accuracy}\label{fig:resolution-b}
  \end{subfigure}\hfill
  \begin{subfigure}{0.33\linewidth}
    \includegraphics[width=\linewidth]{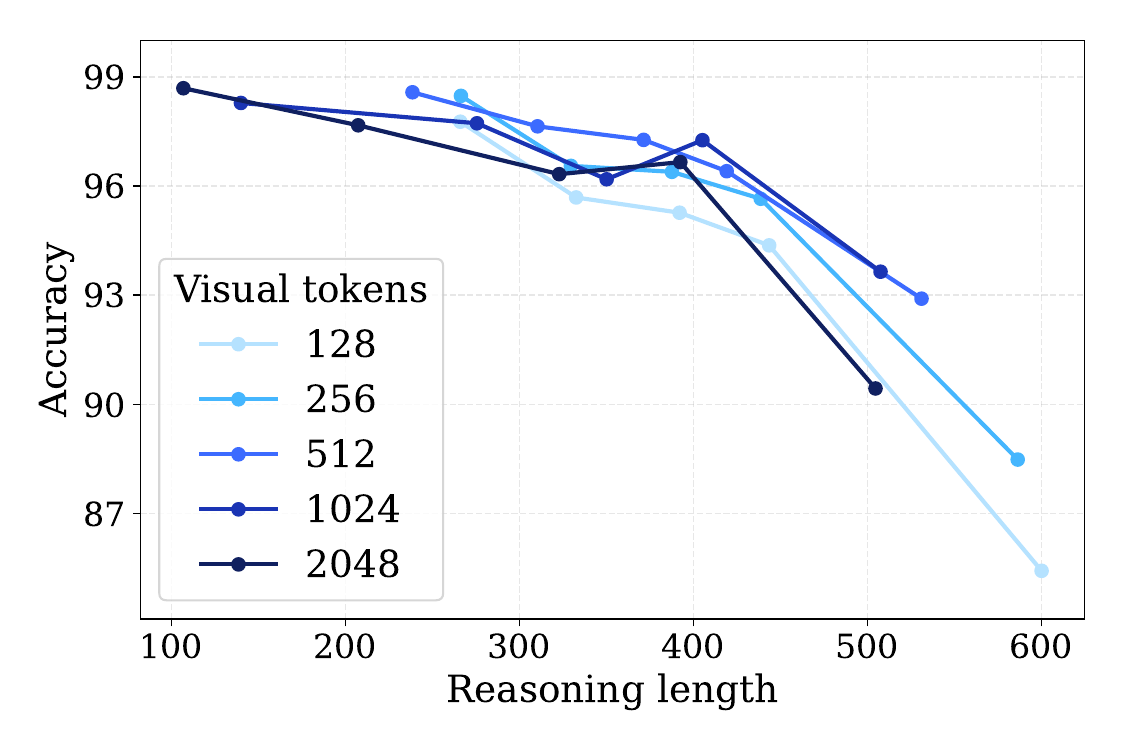}
    \subcaption{Reasoning length \& Accuracy}\label{fig:resolution-c}
  \end{subfigure}\hfill
  \caption{\textbf{Impact of number of visual tokens.} (a) Increasing the number of visual tokens consistently shortens the model’s reasoning traces, indicating that higher-resolution visual inputs reduce the need for long reasoning chains.
(b) Accuracy generally improves as the number of visual tokens increases, eventually reaching a plateau as additional visual detail yields diminishing returns.
(c) The degree of inverse-scaling with respect to reasoning length is similar across visual-token settings; however, the absolute reasoning lengths differ substantially, reflecting the effect of visual-token granularity on the model’s reasoning process.
}
  \label{fig:resolution}
\end{figure*}

\begin{figure}[ht]
  \centering
  \resizebox{0.8\linewidth}{!}{%
    \begin{minipage}{\linewidth}
      \begin{subfigure}{0.5\linewidth}
        \includegraphics[width=\linewidth]{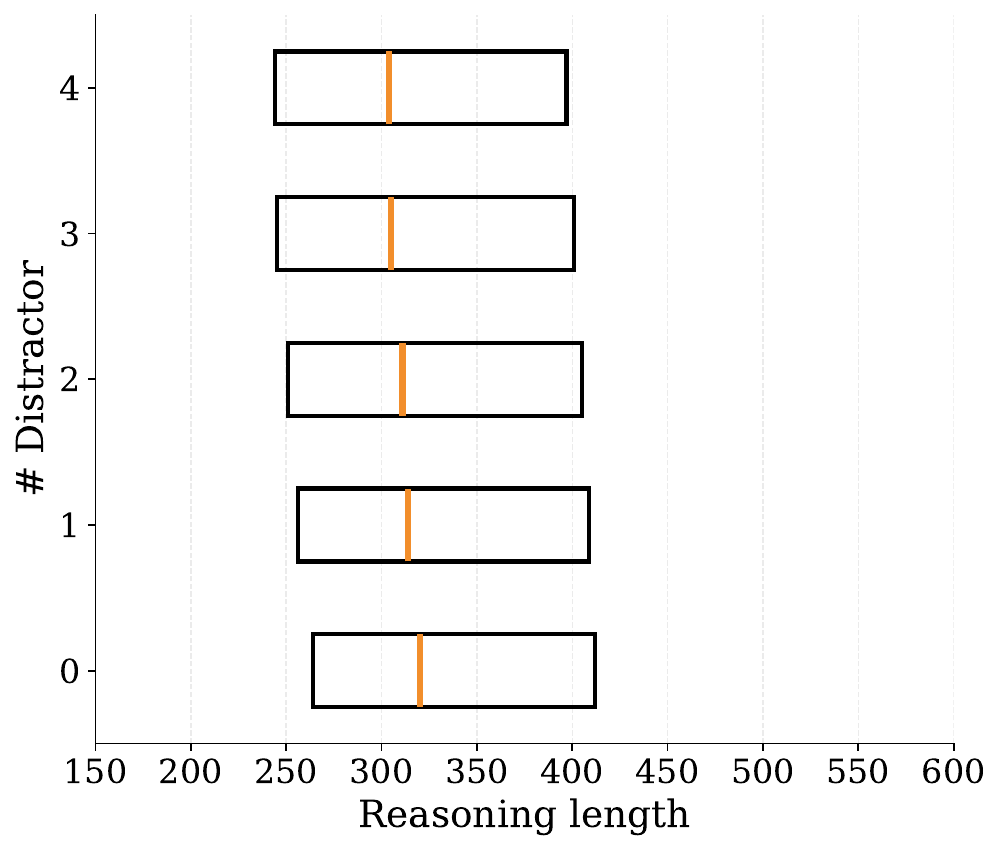}
        \subcaption{Number of distractors}\label{fig:sec8-ob1-a}
      \end{subfigure}\hfill
      \begin{subfigure}{0.5\linewidth}
        \includegraphics[width=\linewidth]{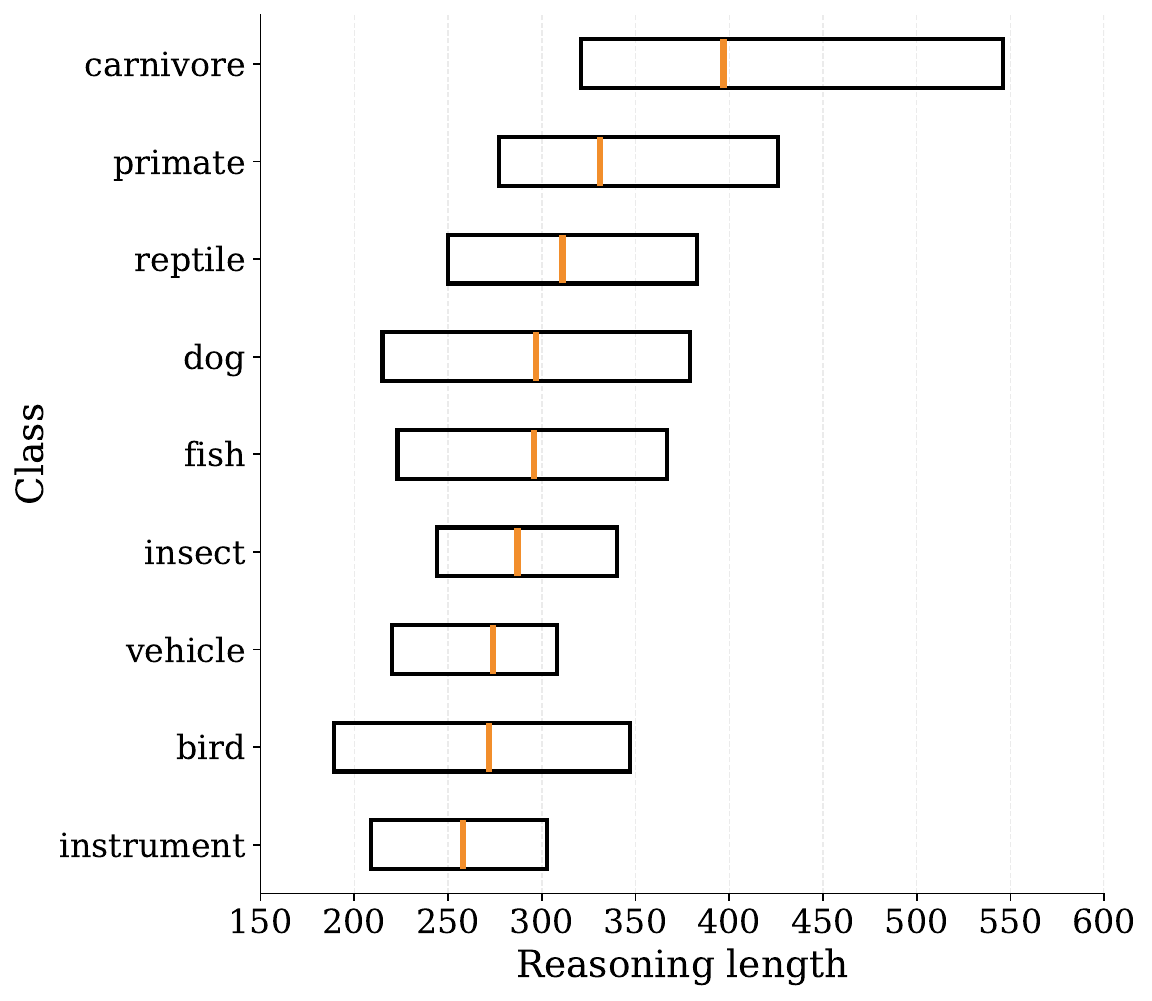}
        \subcaption{Target object class}\label{fig:sec8-ob1-b}
      \end{subfigure}
    \end{minipage}%
  }
  \caption{\textbf{Average reasoning length distribution across distractor counts and image classes.} (a) shows that reasoning lengths remain relatively consistent across different numbers of distractors, whereas (b) reveals substantial variation in length distributions across nine classes of the Idis dataset.}
  \label{fig:sec5-class_length}
\end{figure}

\section{Beyond distractors: Exploring reasoning length determinants}
\label{sec:suppl_D}

In this section, we conduct further investigation into the factors that influence reasoning length, given that the presence of distractors did not yield significant changes in this metric.

We further analyze the distributions of reasoning lengths across three dimensions: image class, number of visual tokens, and sampling variability.

\subsection{Impact of image class}
\label{sec:suppl_C-1}

We compare how reasoning length changes with the number of distractors to how it changes across object classes. We perform inference using Qwen3-VL-Thinking. As shown in~\cref{fig:sec5-class_length}, variations in reasoning length across object classes are substantially larger than those induced by additional distractors. This suggests that reasoning length is primarily governed by the intrinsic properties of the target object in the image, rather than by the presence of auxiliary visual elements.

\subsection{Impact of number of visual tokens}
\label{sec:suppl_C-2}

The Idis dataset comprises images at a native resolution of 1024$\times$1024 pixels. To analyze how varying the number of visual tokens affects model reasoning length, we resized images to 128$\times$128, 256$\times$256, 512$\times$512, 1024$\times$1024, and 2048$\times$2048, corresponding to visual-token counts of 128, 256, 512, 1024, and 2048, respectively. We perform inference using Qwen3-VL-Thinking across all token counts. 

Overall, we find that the number of visual tokens primarily determines a sample’s position along a shared inverse-scaling curve: fewer visual tokens induce longer, lower-accuracy reasoning sequences, whereas moderate visual token counts shorten chains and improve accuracy, with diminishing returns at very high token counts.

In detail, as shown in~\cref{fig:resolution-a}, reasoning length decreases consistently as the number of visual tokens increases, indicating a strong inverse relationship. Larger token budgets provide richer visual information, reducing the need for extended reasoning and mitigating overthinking.

\cref{fig:resolution-b} reveals a parallel trend in accuracy. Performance is lowest at 128--256 tokens, peaks around 512--1024 tokens, and slightly declines at 2048 tokens. When combined with the length–accuracy curves in~\cref{fig:resolution-c}, the underlying mechanism becomes clear: across all visual-token settings, accuracy monotonically decreases as reasoning length increases, and the curves for different token counts substantially overlap. This alignment suggests that the inverse-scaling relationship holds regardless of token count.

Consequently, the accuracy patterns in~\cref{fig:resolution-b} are largely mediated by where each sample lands on the shared length–accuracy curve. Low token counts push samples into a long-chain, low-accuracy region, while moderate token counts move them into shorter, higher-accuracy regimes. The slight degradation at 2048 tokens indicates diminishing returns, where additional visual tokens no longer meaningfully improve accuracy despite further shortening the reasoning chains.

These trends likely reflect a simple balance. With too few visual tokens, the model cannot fully perceive the scene, leading to overthinking with unnecessarily long reasoning. With too many tokens, the model shortens its chains, suggesting a diminishing effect on reasoning length as visual information becomes increasingly abundant. Thus, the number of visual tokens naturally modulates how much the model needs to reason.

\subsection{Impact of sampling}
\label{sec:suppl_C-3}
To quantify the variability inherent in autoregressive generation, we sampled each image in the Idis dataset 64 times using the Qwen3-VL-Thinking, with standard settings of temperature 0.7 and Top-$p$ (nucleus sampling) 0.95.

Across the four distractor conditions, the average response lengths were \(\{397.4, 378.5, 370.7, 361.0\}\) tokens for 1 through 4 distractors, respectively. At the single-image level, this corresponds to an average per-step slope of 11.8 tokens per additional distractor and an average reduction of 36.6 tokens when increasing the number of distractors. Overall, these extents indicate that the impact of the distractor count on reasoning length is relatively slight.

However, within-image sampling variability is substantial: the mean sampling standard deviation is 122.4~tokens. Consequently, sampling noise overwhelms the distractor effect at the single-image level, where stochastic decoding noise (SD \(\approx\) 122 tokens) is roughly \(10\times\) larger than the per-step effect and \(3\times\) larger than the full reduction when increasing from 1 to 4 distractors. Concretely, if we randomly sample one response from a 1-distractor image and one from a 4-distractor image, the first response is longer only 40\% of the time---barely better than chance.

While sampling noise dominates at the single-image level, its impact diminishes when aggregating over many images. When averaging across 4,050 images per condition, the standard error drops to \(3.4\) tokens (95\% CI: \(\pm 6.7\)), yielding an overall effect of \(\Delta_{4-1} = -36.3 \pm 9.5\) tokens. Thus, the random variability that obscures distractor effects in individual images largely cancels out in aggregate, allowing prior analyses to reliably measure average trends across the whole dataset.

\section{Controlled reasoning budgets} \label{sec:suppl_E}

\begin{figure*}
  \centering
  \begin{subfigure}{0.3\linewidth}
    \includegraphics[width=\linewidth]{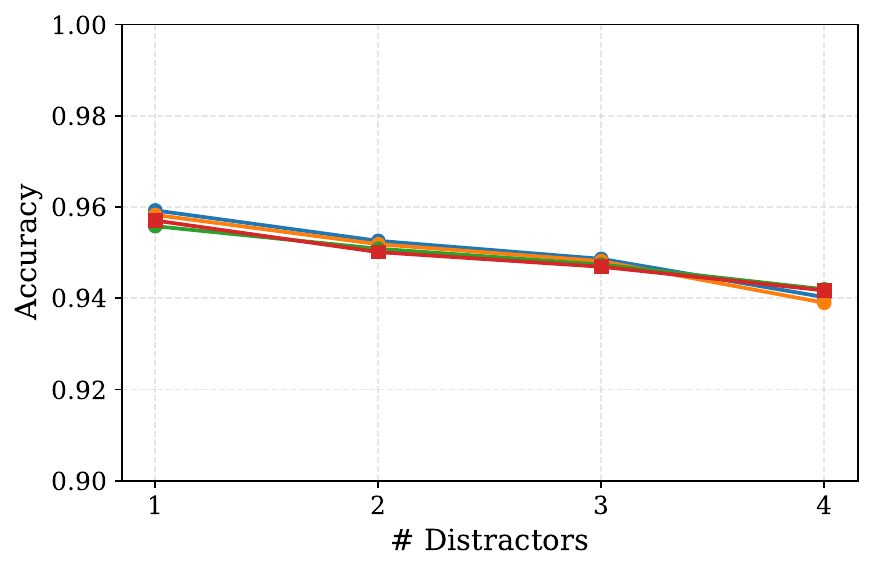}
    \subcaption{Conflicting}\label{fig:a}
  \end{subfigure}\hfill
  \begin{subfigure}{0.3\linewidth}
    \includegraphics[width=\linewidth]{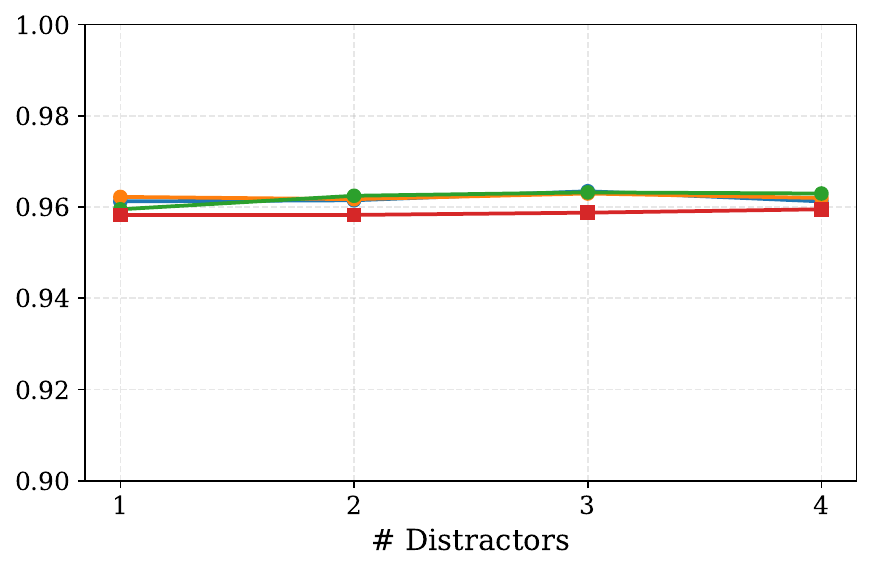}
    \subcaption{Irrelevant}\label{fig:b}
  \end{subfigure}\hfill
  \begin{subfigure}{0.3\linewidth}
    \includegraphics[width=\linewidth]{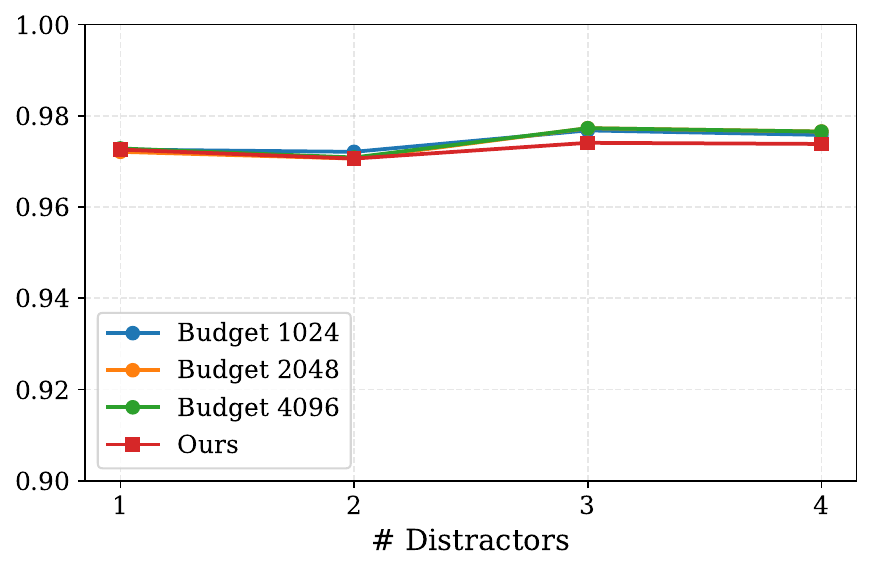}
    \subcaption{Aligned}\label{fig:c}
  \end{subfigure}\hfill
  \caption{\textbf{Reasoning budgets do not meaningfully affect accuracy in reasoning VLMs.} Across all distractor counts (from 1 to 4), both the controlled overthinking setting that adjusts the thinking budget via prompting (1024, 2048, 4096 tokens) and the natural overthinking setting (“Ours”) yield nearly identical performance. This contrasts with reasoning language models, where longer budgets typically alter the scaling curve.
Accuracy decreases most noticeably only in the conflicting distractor condition, whereas aligned and irrelevant distractors maintain stable accuracy regardless of distractor count, demonstrating that semantic conflict—not budget size—drives performance degradation.}
  \label{fig:suppl-d-acc}
\end{figure*}

\begin{figure*}
  \centering
  \begin{subfigure}{0.3\linewidth}
    \includegraphics[width=\linewidth]{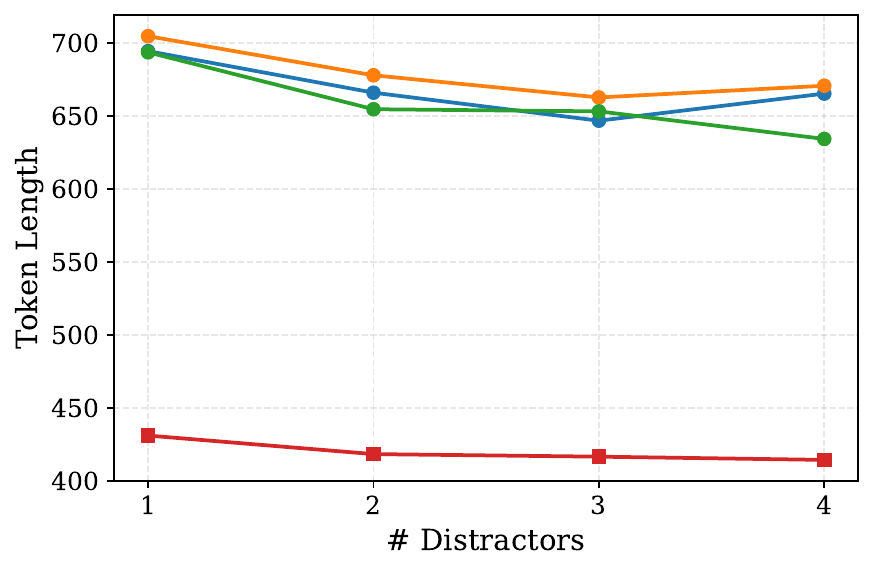}
    \subcaption{Conflicting}\label{fig:a}
  \end{subfigure}\hfill
  \begin{subfigure}{0.3\linewidth}
    \includegraphics[width=\linewidth]{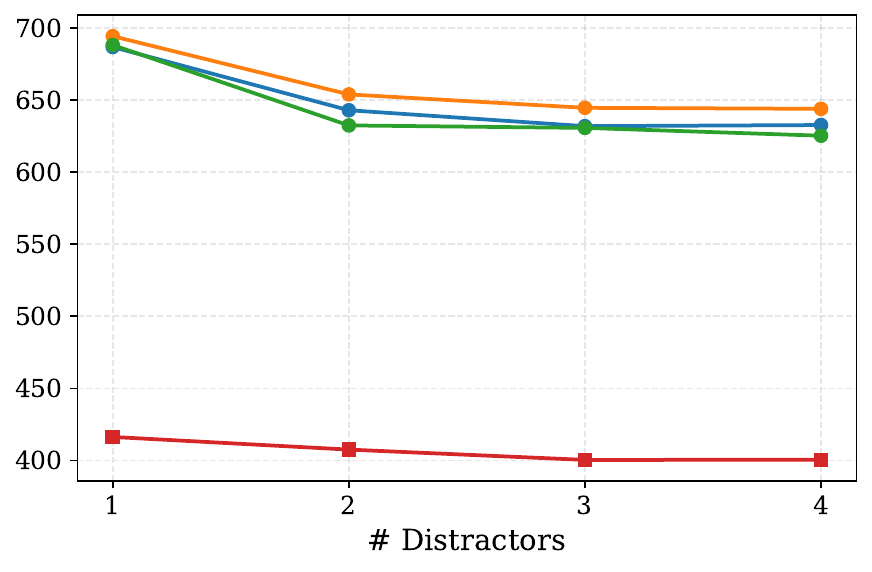}
    \subcaption{Irrelevant}\label{fig:b}
  \end{subfigure}\hfill
  \begin{subfigure}{0.3\linewidth}
    \includegraphics[width=\linewidth]{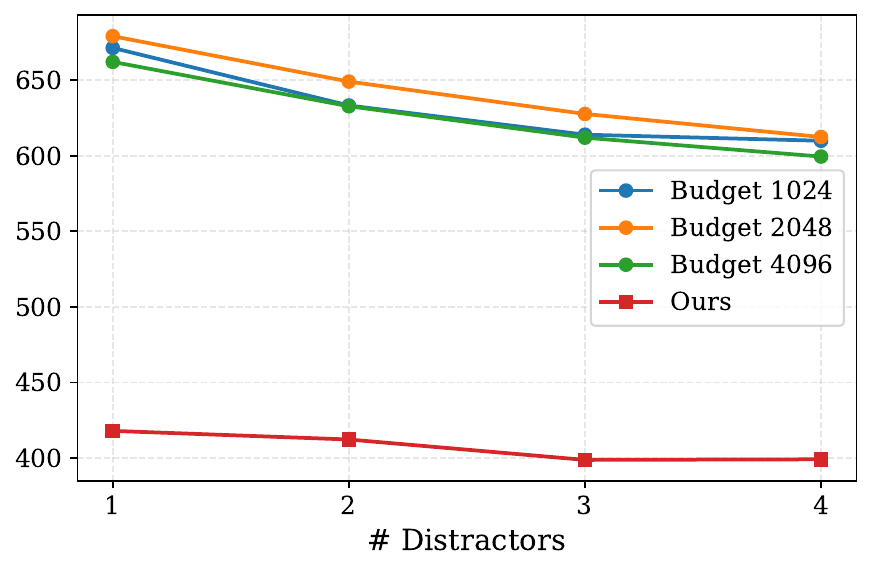}
    \subcaption{Aligned}\label{fig:c}
  \end{subfigure}\hfill
  \caption{\textbf{Reasoning budgets do not meaningfully affect reasoning length in reasoning VLMs.} Across all distractor conditions---i.e., conflicting, irrelevant, and aligned---the controlled overthinking settings with prompting-based budgets (1024, 2048, 4096 tokens) produce nearly identical reasoning lengths, mirroring the stability observed in accuracy.
However, these controlled settings consistently generate substantially longer reasoning traces than the natural overthinking setting (“Ours”).}
  \label{fig:suppl-d-length}
\end{figure*}

We additionally consider a controlled overthinking setting where we explicitly cap the thinking length via prompting, in contrast to the natural overthinking setting used in our main experiments. Concretely, we prepend an instruction that fixes the maximum number of reasoning tokens to 1024, 2048, or 4096 and evaluate the model on the Idis dataset. As shown in \cref{fig:suppl-d-acc} and \cref{fig:suppl-d-length}, varying this budget barely changes either accuracy or reasoning length across all distractor counts and semantic types, and the three budgeted variants almost overlap in both metrics. Consistent with our main results, \cref{fig:suppl-d-acc} also shows that accuracy drops the most under conflicting distractors. These results suggest that, unlike in reasoning LMs, the test-time behavior of reasoning VLMs is largely insensitive to such prompt-level budget control, so we conduct all main analyses under the natural overthinking setting.




\clearpage
\section{Additional experimental results}\label{sec:suppl_F}

\subsection{Full quantitative results on Idis dataset} \label{sec:suppl_F-1}

\begin{figure*}[t]
  \centering

  \begin{subfigure}{0.24\linewidth}
    \includegraphics[width=\linewidth]{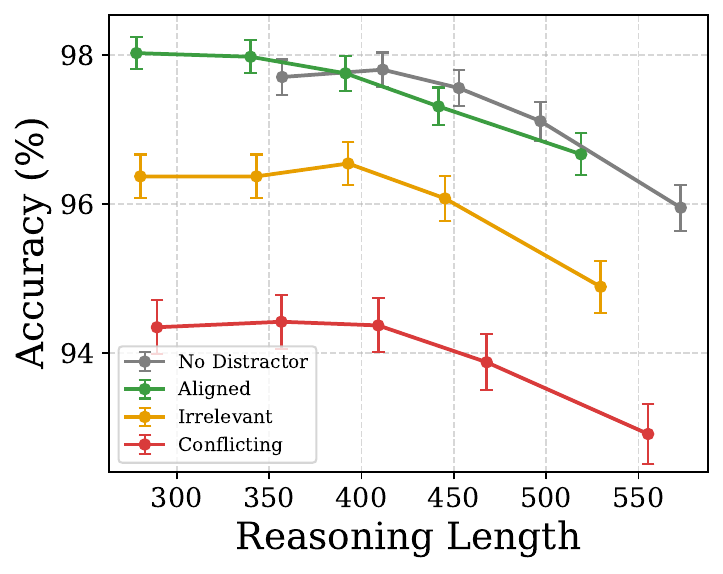}
  \end{subfigure}\hfill
  \begin{subfigure}{0.24\linewidth}
    \includegraphics[width=\linewidth]{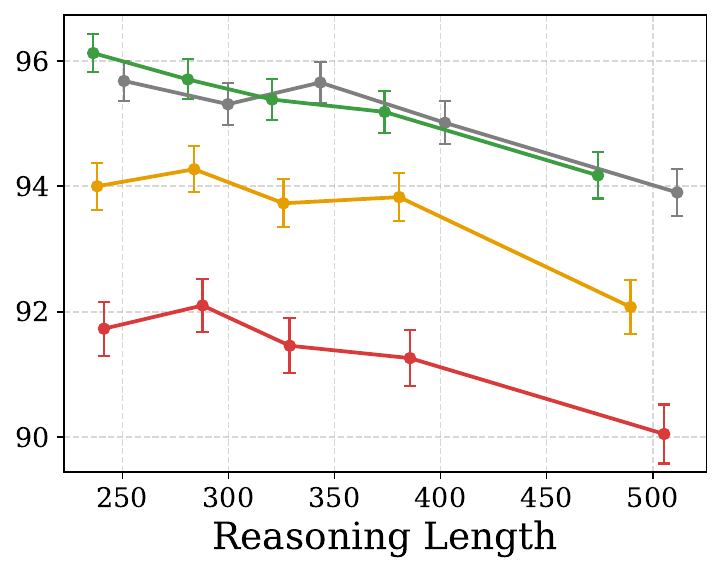}
  \end{subfigure}\hfill
  \begin{subfigure}{0.24\linewidth}
    \includegraphics[width=\linewidth]{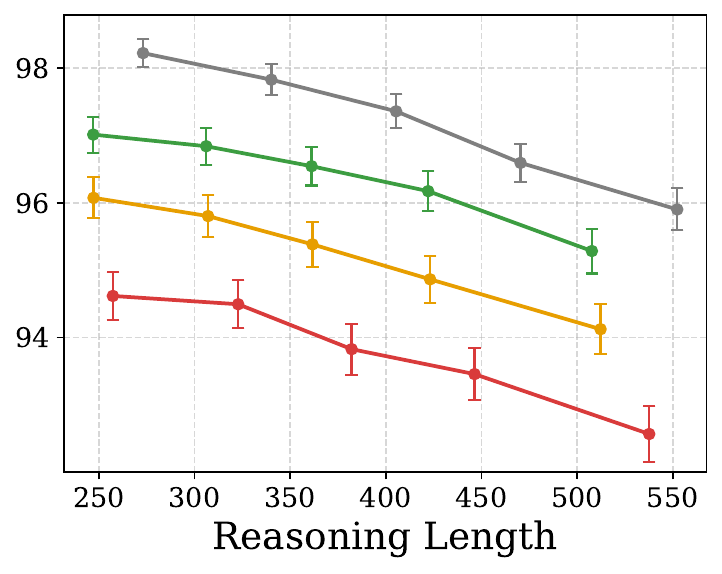}
  \end{subfigure}\hfill
  \begin{subfigure}{0.24\linewidth}
    \includegraphics[width=\linewidth]{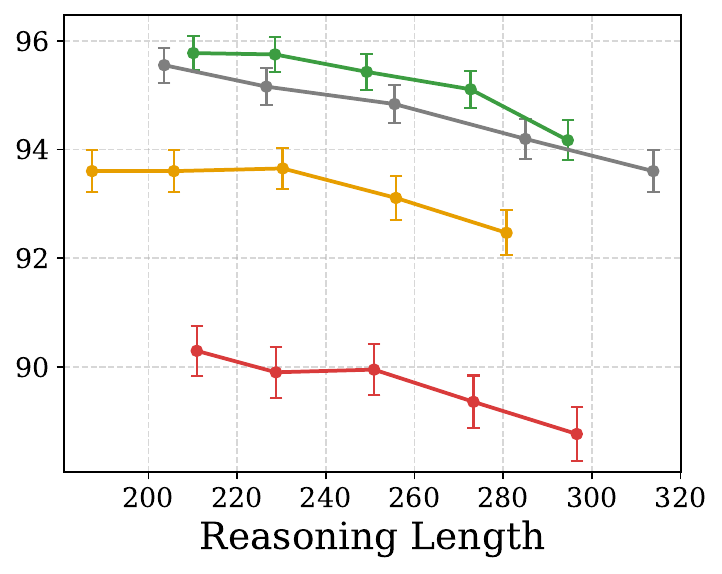}
  \end{subfigure}

  \vspace{0.5em}

  \begin{subfigure}{0.24\linewidth}
    \includegraphics[width=\linewidth]{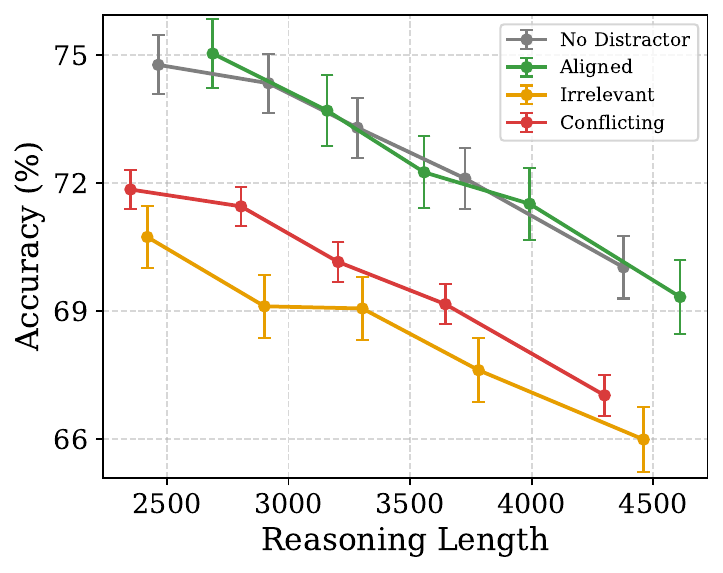}
    \subcaption{Qwen3-VL-Thinking}\label{fig:semantic_e}
  \end{subfigure}\hfill
  \begin{subfigure}{0.24\linewidth}
    \includegraphics[width=\linewidth]{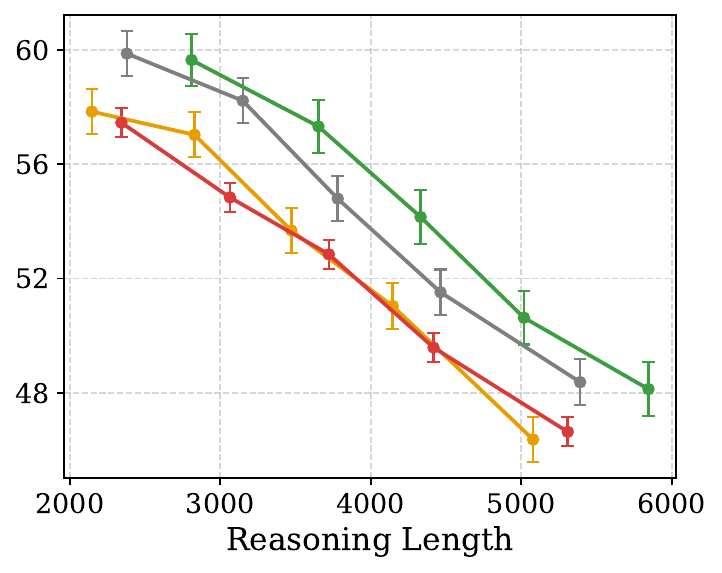}
    \subcaption{Intern-S1-mini}\label{fig:semantic_f}
  \end{subfigure}\hfill
  \begin{subfigure}{0.24\linewidth}
    \includegraphics[width=\linewidth]{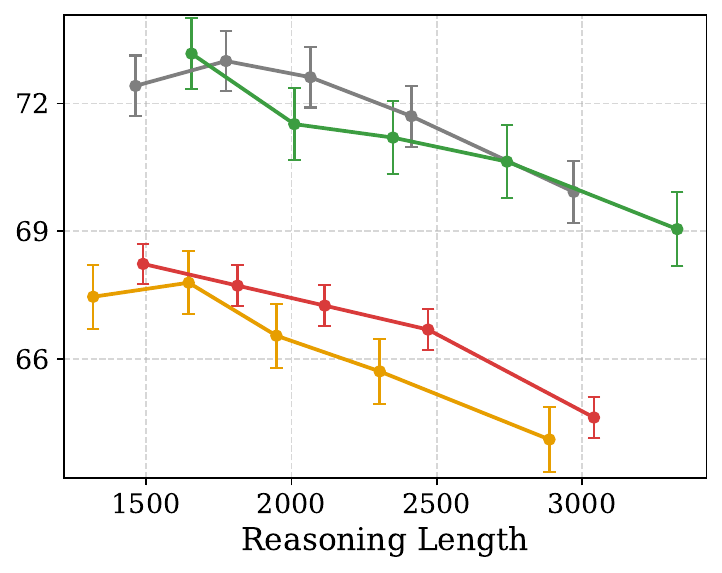}
    \subcaption{GLM-4.1V-Thinking}\label{fig:semantic_g}
  \end{subfigure}\hfill
  \begin{subfigure}{0.24\linewidth}
    \includegraphics[width=\linewidth]{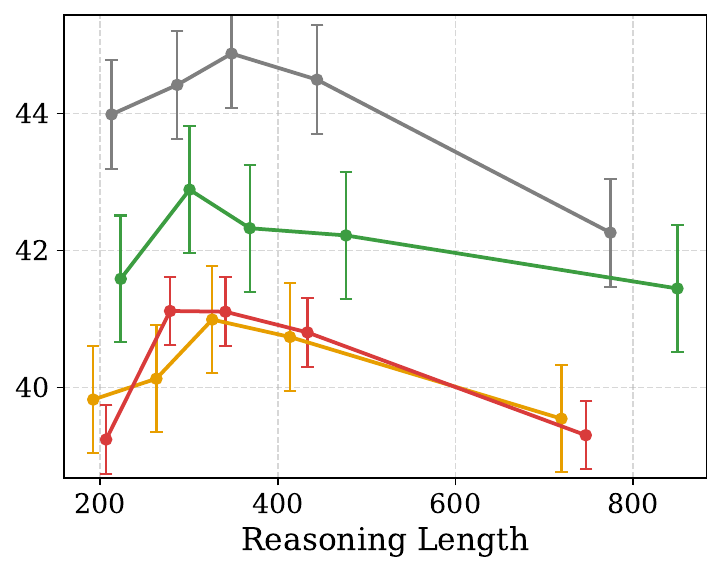}
    \subcaption{R1-OneVision}\label{fig:semantic_h}
  \end{subfigure}

  \caption{\textbf{Scaling behavior in reasoning VLMs, with various types of semantic relationship of visual distractors to the target object.}
  }
  \label{fig:full_semantic}
\end{figure*}

\begin{figure*}[!t]
  \centering
  \begin{subfigure}{0.25\linewidth}
        \includegraphics[width=\linewidth]{figure/qwen3-thinking_outval_vs_original_bin5.pdf}
    \subcaption{Qwen3-VL-Thinking}\label{fig:attribute_a}
  \end{subfigure}\hfill
  \begin{subfigure}{0.25\linewidth}
    \includegraphics[width=\linewidth]{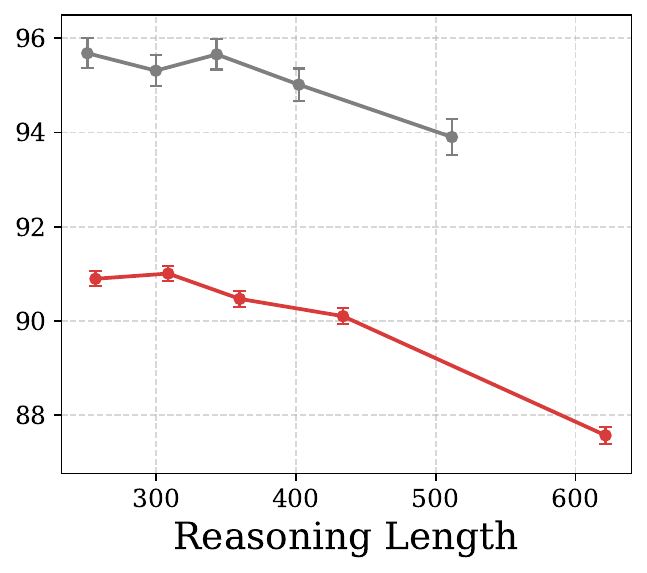}
    \subcaption{Intern-S1-mini}\label{fig:attribute_b}
  \end{subfigure}\hfill
  \begin{subfigure}{0.25\linewidth}
    \includegraphics[width=\linewidth]{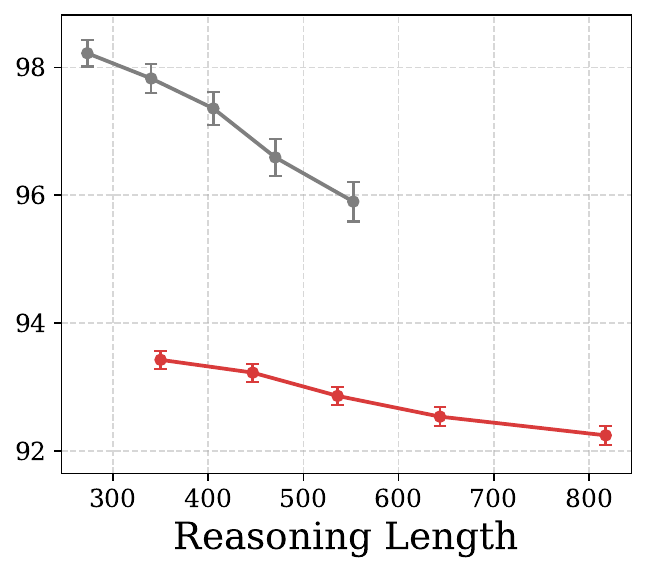}
    \subcaption{GLM-4.1V-Thinking}\label{fig:attribute_c}
  \end{subfigure}\hfill
  \begin{subfigure}{0.25\linewidth}
    \includegraphics[width=\linewidth]{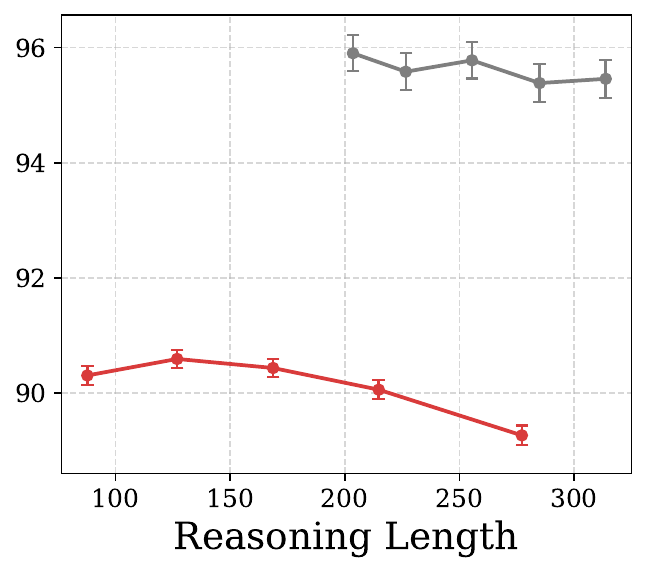}
    \subcaption{R1-OneVision}\label{fig:attribute_d}
  \end{subfigure}
\caption{\textbf{The effect of typographic distractors on Idis-perception.}}
\label{fig:idis_textual_full}
\end{figure*}

\begin{figure*}[!t]
  \centering
  \begin{subfigure}{0.25\linewidth}
        \includegraphics[width=\linewidth]{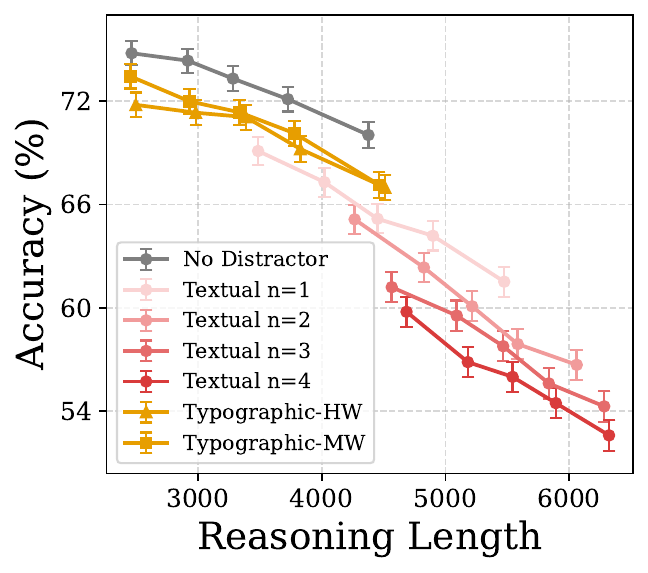}
    \subcaption{Qwen3-VL-Thinking}\label{fig:attribute_a}
  \end{subfigure}\hfill
  \begin{subfigure}{0.25\linewidth}
    \includegraphics[width=\linewidth]{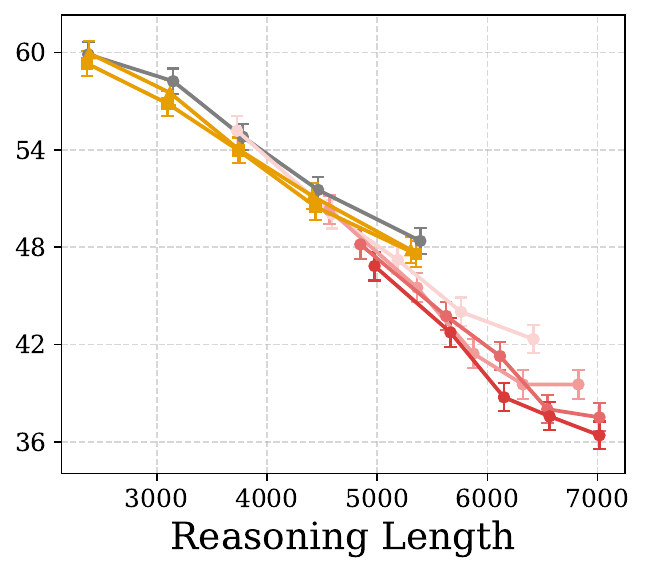}
    \subcaption{Intern-S1-mini}\label{fig:attribute_b}
  \end{subfigure}\hfill
  \begin{subfigure}{0.25\linewidth}
    \includegraphics[width=\linewidth]{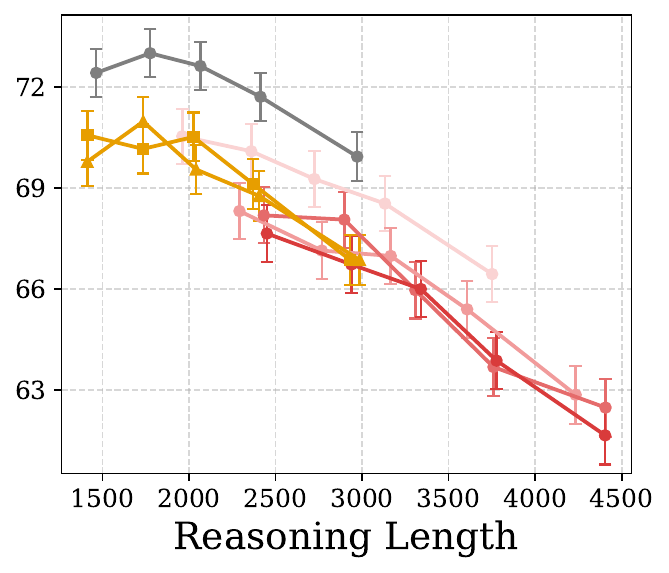}
    \subcaption{GLM-4.1V-Thinking}\label{fig:attribute_c}
  \end{subfigure}\hfill
  \begin{subfigure}{0.25\linewidth}
    \includegraphics[width=\linewidth]{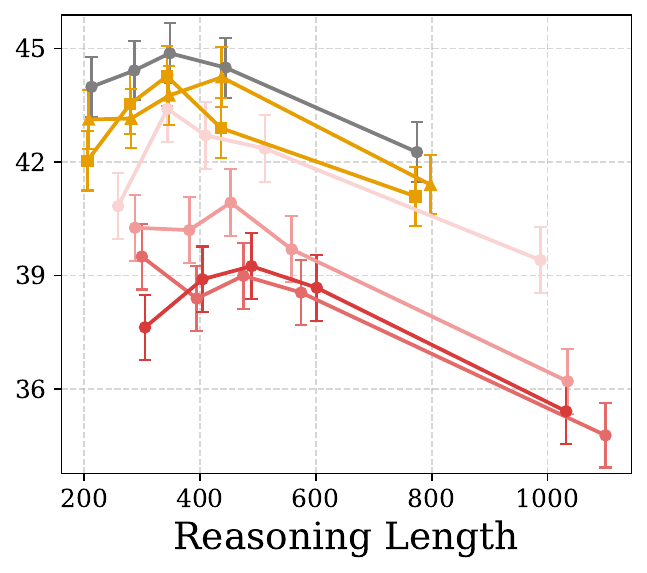}
    \subcaption{R1-OneVision}\label{fig:attribute_d}
  \end{subfigure}
\caption{\textbf{ The effect of typographic and textual distractors on Idis-math.}}
\label{fig:math_textual_full}
\end{figure*}

\begin{figure*}[!t]
  \centering
  \begin{subfigure}{0.25\linewidth}
    \includegraphics[width=\linewidth]{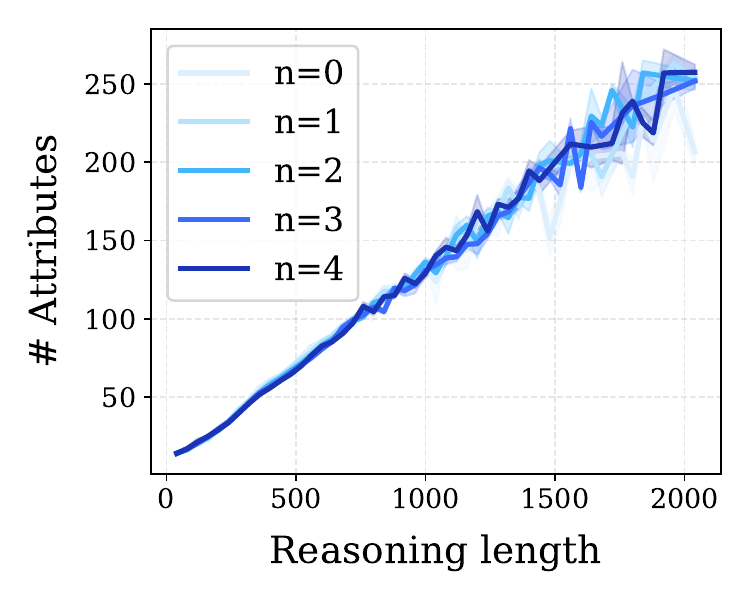}
    \subcaption{Qwen3-VL-Thinking}\label{fig:attribute_a}
  \end{subfigure}\hfill
  \begin{subfigure}{0.25\linewidth}
    \includegraphics[width=\linewidth]{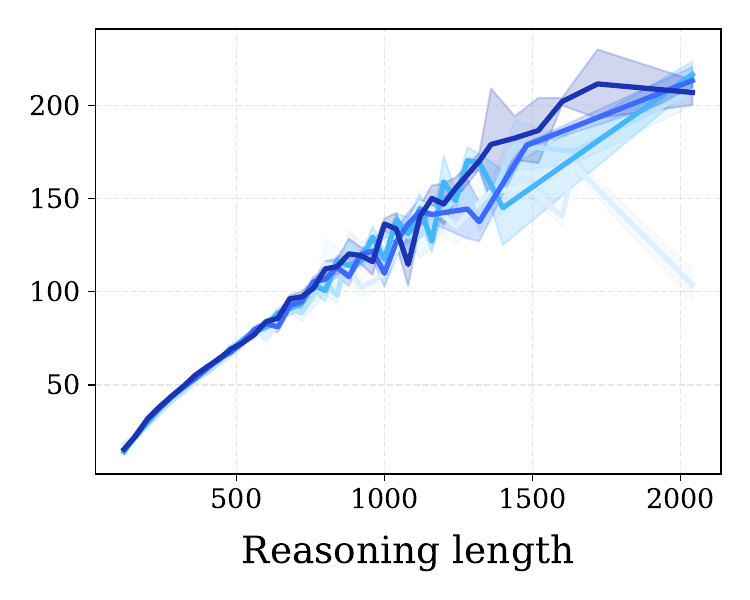}
    \subcaption{Intern-S1-mini}\label{fig:attribute_b}
  \end{subfigure}\hfill
  \begin{subfigure}{0.25\linewidth}
    \includegraphics[width=\linewidth]{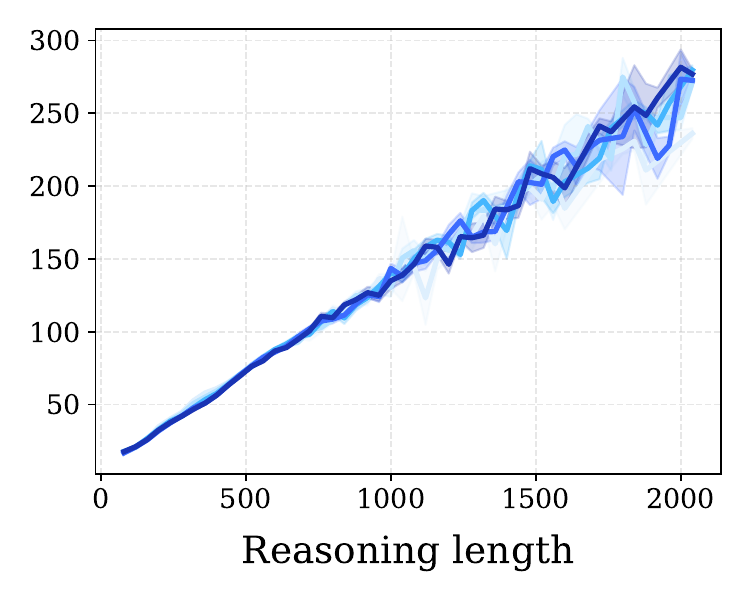}
    \subcaption{GLM-4.1V-Thinking}\label{fig:attribute_c}
  \end{subfigure}\hfill
  \begin{subfigure}{0.25\linewidth}
    \includegraphics[width=\linewidth]{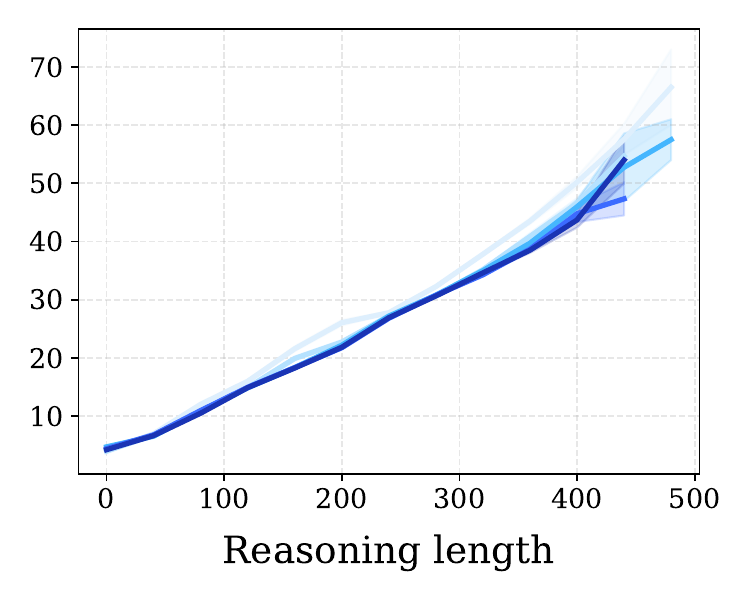}
    \subcaption{R1-OneVision}\label{fig:attribute_d}
  \end{subfigure}
\caption{\textbf{ A strong linear correlation between reasoning length and the number of attributes.}}
\label{fig:attribute_appendix}
\end{figure*}

\begin{figure*}[!t]
  \centering
  \begin{subfigure}{0.25\linewidth}
        \includegraphics[width=\linewidth]{figure/qwen3-thinking_dist-att.pdf}
    \subcaption{Qwen3-VL-Thinking}\label{fig:attribute_a}
  \end{subfigure}\hfill
  \begin{subfigure}{0.25\linewidth}
    \includegraphics[width=\linewidth]{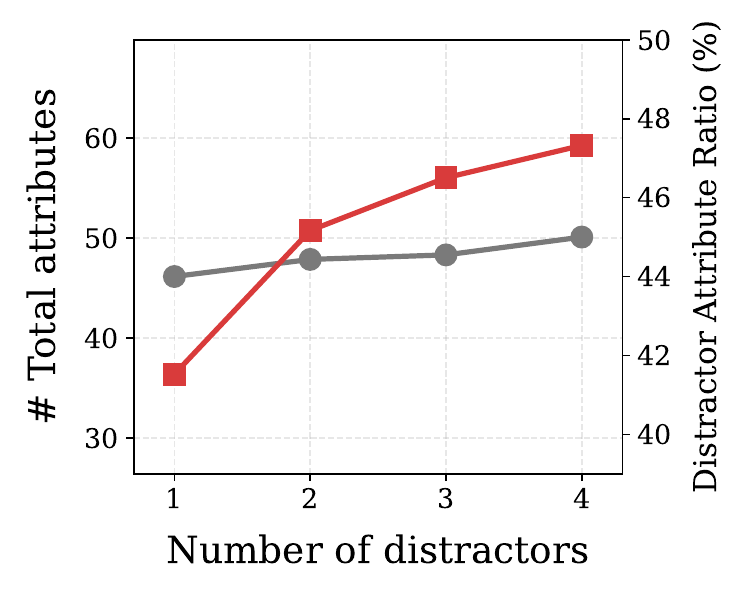}
    \subcaption{Intern-S1-mini}\label{fig:attribute_b}
  \end{subfigure}\hfill
  \begin{subfigure}{0.25\linewidth}
    \includegraphics[width=\linewidth]{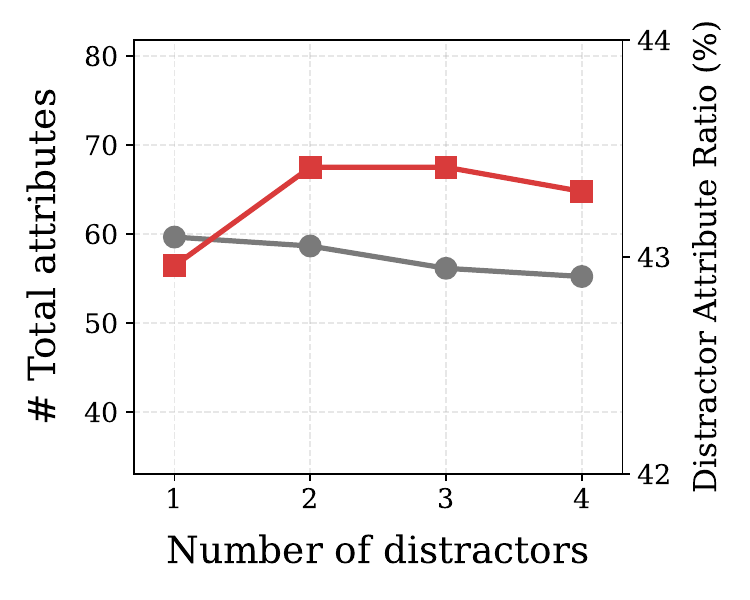}
    \subcaption{GLM-4.1V-Thinking}\label{fig:attribute_c}
  \end{subfigure}\hfill
  \begin{subfigure}{0.25\linewidth}
    \includegraphics[width=\linewidth]{figure/onevision_dist-att.pdf}
    \subcaption{R1-OneVision}\label{fig:attribute_d}
  \end{subfigure}
\caption{\textbf{ The effect of typographic and textual distractors on Idis-math.}}
\label{fig:att_full}
\end{figure*}

\begin{figure*}[!t]
  \centering
  \begin{subfigure}{0.25\linewidth}
    \includegraphics[width=\linewidth]{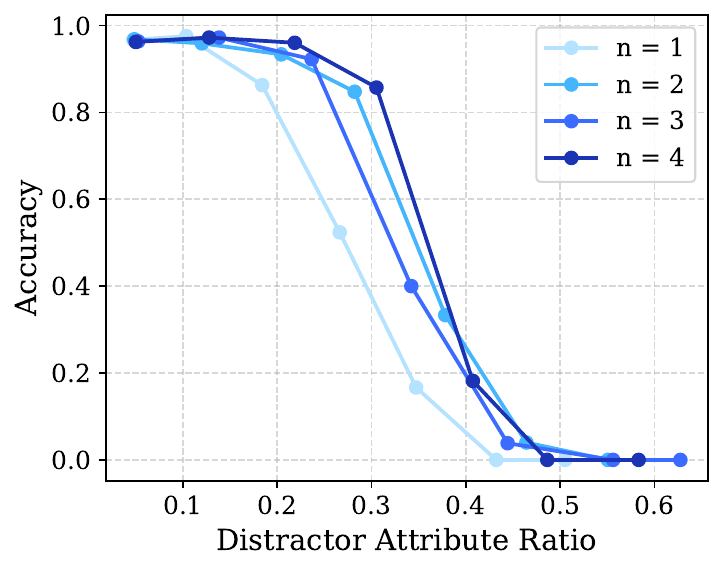}
    \subcaption{Qwen3-VL-Thinking}\label{fig:attribute_a}
  \end{subfigure}\hfill
  \begin{subfigure}{0.25\linewidth}
    \includegraphics[width=\linewidth]{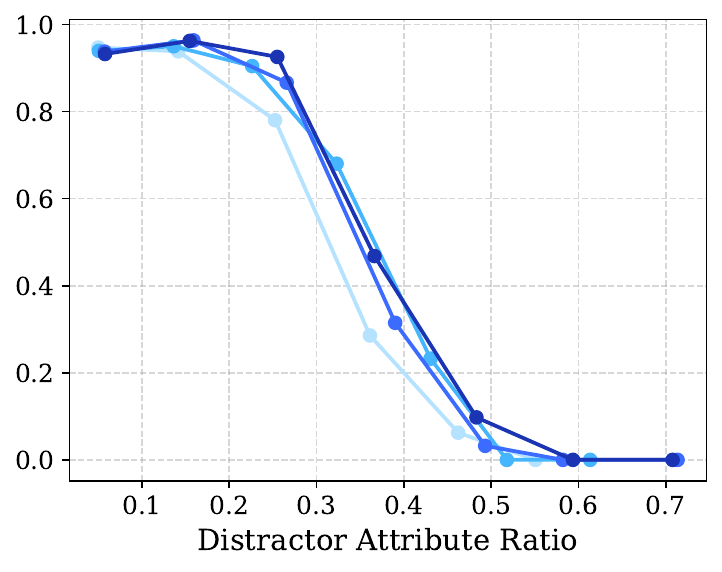}
    \subcaption{Intern-S1-mini}\label{fig:attribute_b}
  \end{subfigure}\hfill
  \begin{subfigure}{0.25\linewidth}
    \includegraphics[width=\linewidth]{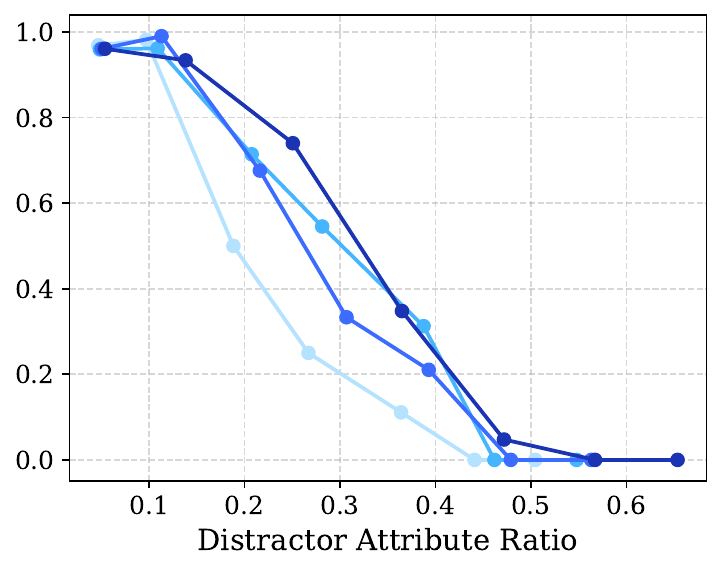}
    \subcaption{GLM-4.1V-Thinking}\label{fig:attribute_c}
  \end{subfigure}\hfill
  \begin{subfigure}{0.25\linewidth}
    \includegraphics[width=\linewidth]{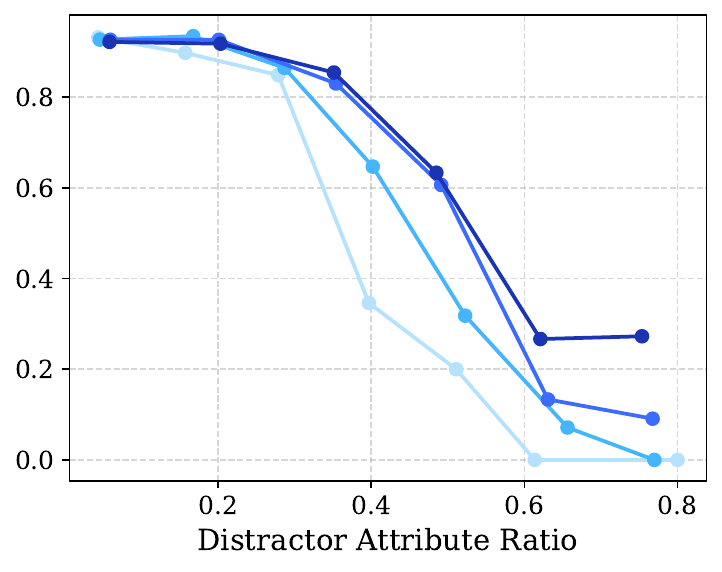}
    \subcaption{R1-OneVision}\label{fig:attribute_d}
  \end{subfigure}
\caption{\textbf{The distractor attribute ratio is negatively correlated with the accuracy.}}
\label{fig:attribute_acc_appendix}
\end{figure*}

In this section, we provide additional quantitative results on the Idis dataset across all four reasoning VLMs (Qwen3-VL-Thinking, Intern-S1-mini, GLM-4.1V-Thinking, and R1-OneVision). 
\cref{fig:full_semantic} shows that, across all models, conflicting distractors cause the largest declines with downward shifts, indicating that reasoning VLMs are most vulnerable to distractors that semantically conflict with the target object. 
As illustrated in \cref{fig:idis_textual_full}, typographic distractors rendered as images exhibit a similar trend to visual distractors across all four models, shifting the length--accuracy curve downward with only a limited increase in reasoning length.
As shown in \cref{fig:math_textual_full}, textual distractors inserted into the question prompt reduce accuracy across all four models by inducing longer reasoning traces. 
\cref{fig:attribute_appendix} shows the relationship between reasoning length and the number of generated visual attributes. All models exhibit a strong linear positive correlation: longer traces systematically produce more attributes, and this trend holds regardless of the number of distractors. 
\cref{fig:att_full} presents that increasing the number of distractors does not substantially change the total number of attributes, but consistently increases the fraction of distractor-related attributes.
Finally, \cref{fig:attribute_acc_appendix} illustrates that the distractor-related attribute ratio is negatively correlated with the accuracy. As the fraction of attributes assigned to distractors increases, accuracy monotonically decreases, and when distractor attributes dominate, accuracy effectively collapses. Taken together, these full quantitative results support our main finding that performance degradation on Idis is driven by how attributes are allocated to visual distractors rather than the target object, not by an overall expansion of the reasoning length.

\begin{table}
\centering
\renewcommand{\arraystretch}{1.2}
\setlength{\tabcolsep}{12pt}
\resizebox{1\linewidth}{!}{
\begin{tabular}{c cccc}
\toprule
& \textbf{Qwen3-VL-Thinking} & \textbf{Intern-S1-mini} & \textbf{GLM-4.1V-Thinking} & \textbf{R1-OneVision} \\
\cmidrule(lr){2-5}
\textbf{Size} & Acc. & Acc. & Acc. & Acc. \\
\midrule
Small  & 96.1 & 94.4 & 96.3 & 92.86 \\
Medium & 95.5 & 93.1 & 95.6 & 91.68 \\
Large  & 94.0 & 90.9 & 94.7 & 88.25 \\
\bottomrule
\end{tabular}
}
\caption{\textbf{Accuracy results on the Idis-manual dataset.} Accuracy across three distractor-size conditions (Small, Medium, Large) for four reasoning VLMs.}
\label{tab:model_size_acc}
\end{table}

\subsection{Results on Idis-manual dataset}\label{ssec:idis_manual}
While Gemini 2.5 Flash Image enables flexible image editing, it does not allow precise control over the size of generated distractors. To enable fine-grained manipulation of distractor size, we construct Idis-manual by directly over-laying background-masked segments of distracting objects onto the original images.
Specifically, we use Language Segment Anything (LangSAM) \citep{langsam} to extract image segments from the textual description of each distractor. The extracted segments are then resized according to predefined small, medium, or large configurations and overlaid onto the target image.

\begin{figure*}[ht]
  \centering
  \begin{subfigure}{0.25\linewidth}
    \includegraphics[width=\linewidth]{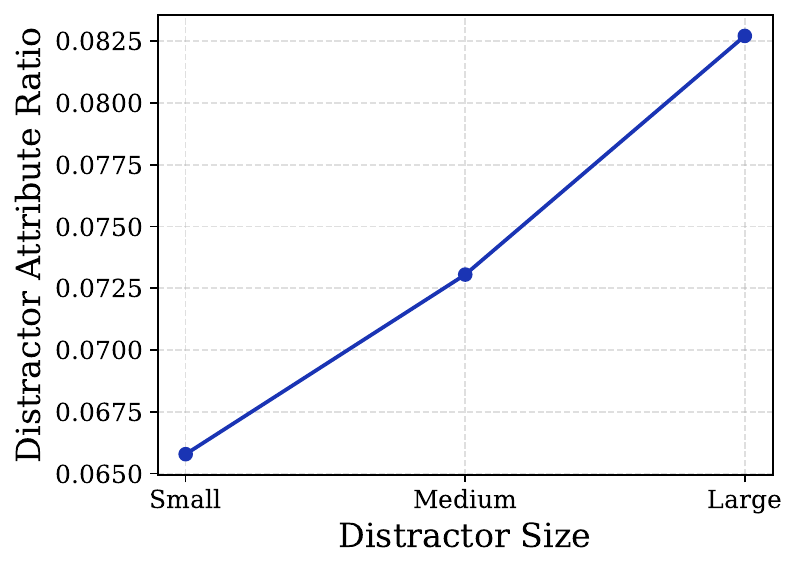}
    \subcaption{Qwen3-VL-Thinking}\label{fig:attribute_a}
  \end{subfigure}\hfill
  \begin{subfigure}{0.25\linewidth}
    \includegraphics[width=\linewidth]{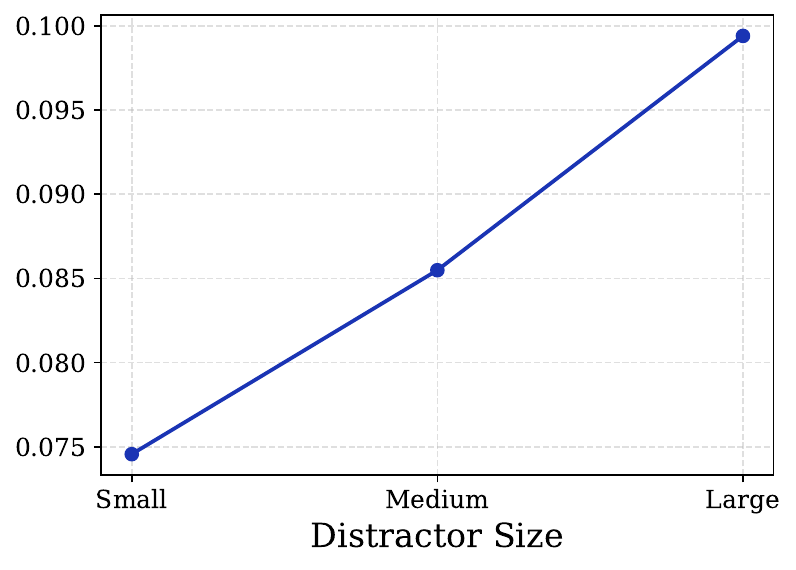}
    \subcaption{Intern-S1-mini}\label{fig:attribute_b}
  \end{subfigure}\hfill
  \begin{subfigure}{0.25\linewidth}
    \includegraphics[width=\linewidth]{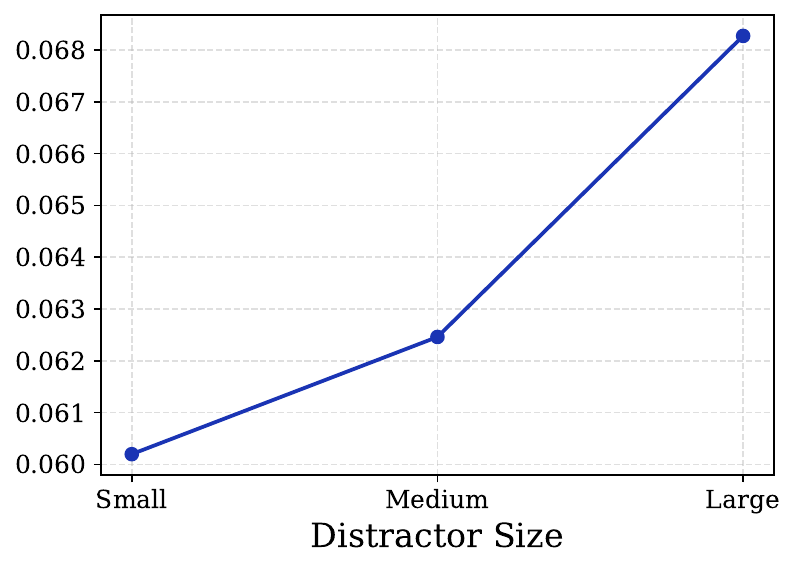}
    \subcaption{GLM-4.1V-Thinking}\label{fig:attribute_c}
  \end{subfigure}\hfill
  \begin{subfigure}{0.25\linewidth}
    \includegraphics[width=\linewidth]{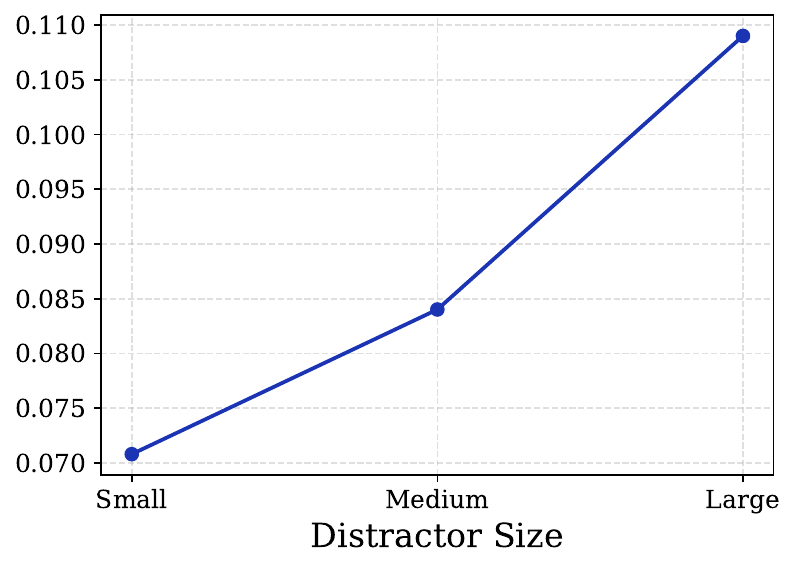}
    \subcaption{R1-OneVision}\label{fig:attribute_d}
  \end{subfigure}
\caption{\textbf{Larger distractors lead to higher distractor-attribute ratios.} Across all four reasoning VLMs, the proportion of distractor-related attributes increases as distractor size grows from small to medium to large. This indicates that larger distractors capture more of the model’s attention and contribute more heavily to the attribute composition.}
\label{fig:manual-attribute_appendix}
\end{figure*}

As shown in \cref{fig:manual-attribute_appendix}, increasing the distractor size from Small to Large consistently raises the distractor-related attribute ratio for all four reasoning VLMs, indicating that larger distractors capture more of the model’s attention and receive a greater portion of the generated attributes. \cref{tab:model_size_acc} further shows that this shift in attribute allocation is accompanied by a clear drop in accuracy as distractor size grows. Taken together, these results show that increasing distractor size is accompanied by both a higher distractor-related attribute ratio and lower accuracy, consistent with an association between distractor-oriented attribute allocation and performance degradation.

\begin{table}[ht]
\centering
\scriptsize 
\renewcommand{\arraystretch}{1.1} 
\setlength{\tabcolsep}{3pt} 
\begin{tabular}{l cc cc cc}
\toprule
\multirow{2}{*}{\textbf{Model}} & \multicolumn{2}{c}{\textbf{Aligned}} & \multicolumn{2}{c}{\textbf{Conflicting}} & \multicolumn{2}{c}{\textbf{Overall}} \\
\cmidrule(lr){2-3} \cmidrule(lr){4-5} \cmidrule(lr){6-7}
& Acc. & Len. & Acc. & Len. & Acc. & Len. \\
\midrule
\textbf{Qwen3-VL-Thinking} & 93.6 & 635.9 & 76.4 & 829.9 & 88.3 & 695.8 \\
\textbf{+ Prompt Strategy} & 94.1 & 474.7 & 78.2 & 652.1 & 89.2 & 529.5 \\
\midrule
\textbf{Intern-S1-mini} & 93.2 & 493.2 & 56.7 & 672.5 & 81.9 & 548.6 \\
\textbf{+ Prompt Strategy} & 92.4 & 574.2 & 58.7 & 711.9 & 82.0 & 616.7 \\
\midrule
\textbf{GLM-4.1V} & 94.4 & 340.9 & 80.7 & 498.2 & 90.1 & 389.5 \\
\textbf{+ Prompt Strategy} & 92.9 & 253.9 & 84.7 & 358.7 & 90.4 & 286.3 \\
\midrule
\textbf{R1-OneVision} & 93.9 & 300.5 & 63.0 & 310.5 & 84.4 & 303.6 \\
\textbf{+ Prompt Strategy} & 93.9 & 233.0 & 66.3 & 244.7 & 85.4 & 236.6 \\
\bottomrule
\end{tabular}
\caption{\textbf{Table results of four vision-language models on the Waterbirds dataset.} 
We report accuracy and average reasoning length for aligned, conflicting, and overall conditions across non-reasoning VLMs, reasoning VLMs, and prompt-strategy settings.}
\label{tab:waterbirds_columnwise_extended_full}
\end{table}

\subsection{Detailed results for debiasing experiments}
\cref{tab:waterbirds_columnwise_extended_full} presents the full results of our debiasing experiments on the Waterbirds dataset, extending the summary trends shown in 
\cref{fig:prompt_waterbirds}. All four reasoning VLMs with the prompt strategy show better performance on the conflicting. These detailed results confirm that our debiasing prompt can mitigate spurious-correlation failures on Waterbirds by steering the reasoning VLMs to reason primarily based on attributes of the target object rather than spurious cues.

\section{Experiments on proprietary model}
To examine whether test-time inverse scaling generalizes beyond open-weight models, we conduct additional experiments on Gemini 3 Flash, a proprietary SOTA model, using the Idis dataset. 
\begingroup
\setlength{\columnsep}{5pt}%
\begin{wrapfigure}{l}{0.45\linewidth}
  \centering
  \vspace{-15pt} 
  \includegraphics[width=\linewidth]{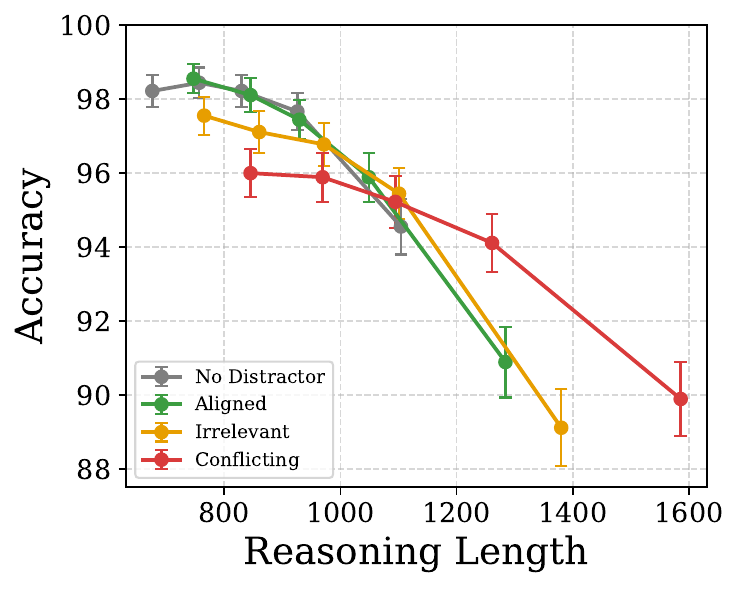}
  \vspace{-20pt} 
  \label{fig:additional_task}
  \vspace{-8pt}
\end{wrapfigure}
\noindent We observe similar inverse scaling trends, with accuracy degrading as reasoning length increases. Notably, the degree of performance degradation varies depending on distractor semantics, consistent with our findings on open-weight models.

\endgroup

\setlength{\columnsep}{5pt}%
\begin{wrapfigure}{l}{0.45\linewidth}
  \centering
  \vspace{-15pt} 
  \includegraphics[width=\linewidth]{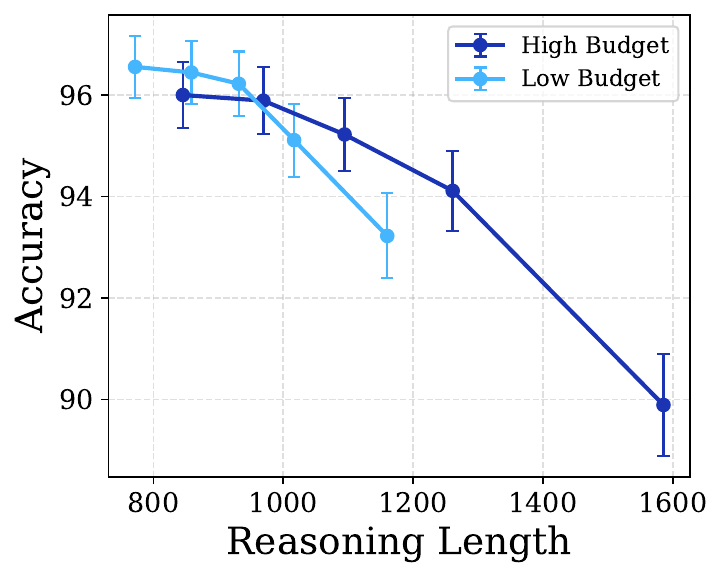}
  \vspace{-20pt} 
  \label{fig:additional_task}
  \vspace{-10pt}
\end{wrapfigure} 
\noindent Furthermore, by varying Gemini 3 Flash's native thinking budget, we find that the "High budget" setting, which encourages longer reasoning, consistently yields lower accuracy than the "Low budget" setup across all reasoning lengths. 

\section{Why do visual and textual distractors induce different scaling patterns}
\label{app:hypothesis}

Visual distractors lower accuracy without lengthening the trace, whereas textual distractors lengthen it (Sec.~\ref{sec:analysis}). We offer a hypothesis and two pieces of preliminary evidence consistent with it, measured on Idis-math with Qwen3-VL-8B-Thinking.

\paragraph{Hypothesis.}
The two modalities differ in whether additional distractors create \emph{cumulative} competition during autoregressive reasoning. A textual distractor lives in the same token space as the reasoning: it stays reachable in the prompt throughout generation, can be restated into the trace, and once restated becomes part of the context later steps attend to. Each additional distractor may thus add material that is processed and re-processed, lengthening the trace. A visual distractor is encoded into a fixed set of image tokens that the model appears to read once. Additional visual distractors then change \emph{what} is extracted, shifting verbalized attributes toward the distractor (Sec.~\ref{sec:6}), without adding new material to reason about, so accuracy drops but the trace does not grow.

\paragraph{Where the model attends.}
After generating each trace, we run a teacher-forced forward pass and sum each generated token's attention over image tokens, prompt tokens, and previously generated tokens, averaging over layers, heads, positions, and samples. Since adding distractors changes segment lengths, we also report a uniform baseline (segment length over context length). Both modalities are evaluated on the same Idis-math questions, so values are directly comparable across them.

\begin{table}[t]
\centering
\footnotesize
\setlength{\tabcolsep}{3.2pt}
\resizebox{\columnwidth}{!}{%
\begin{tabular}{@{}lccccc@{}}
\toprule
& \textbf{Orig.} & $n{=}1$ & $n{=}2$ & $n{=}3$ & $n{=}4$ \\
\midrule
\multicolumn{6}{@{}l}{\textbf{Visual distractors}} \\
\multicolumn{6}{@{}l}{\textit{Attention to image}} \\
\quad Observed (\%) & 0.96 & 2.04 & 1.65 & 1.75 & 1.76 \\
\quad Uniform (\%) & 8.84 & 13.31 & 13.30 & 13.31 & 13.18 \\
\quad Obs./Unif. & \textbf{0.11} & \textbf{0.15} & \textbf{0.12} & \textbf{0.13} & \textbf{0.13} \\
\multicolumn{6}{@{}l}{\textit{Attention by segment (\%)}} \\
\quad Image & 0.96 & 2.04 & 1.65 & 1.75 & 1.76 \\
\quad Prompt & 8.74 & 8.34 & 8.35 & 8.47 & 8.39 \\
\quad Trace & 90.30 & 89.62 & 90.00 & 89.77 & 89.85 \\
\midrule
\multicolumn{6}{@{}l}{\textbf{Textual distractors}} \\
\multicolumn{6}{@{}l}{\textit{Attention to input prompt}} \\
\quad Observed (\%) & 8.79 & 7.97 & 7.38 & 7.33 & 7.45 \\
\quad Uniform (\%) & 4.25 & 4.24 & 4.66 & 5.37 & 6.12 \\
\quad Obs./Unif. & \textbf{2.07} & \textbf{1.88} & \textbf{1.58} & \textbf{1.37} & \textbf{1.22} \\
\multicolumn{6}{@{}l}{\textit{Attention by segment (\%)}} \\
\quad Image & 0.97 & 0.98 & 0.81 & 0.82 & 0.86 \\
\quad Prompt & 8.79 & 7.97 & 7.38 & 7.33 & 7.45 \\
\quad Trace & \textbf{90.24} & \textbf{91.05} & \textbf{91.81} & \textbf{91.84} & \textbf{91.70} \\
\bottomrule
\end{tabular}}
\caption{\textbf{Attention allocation under visual and textual distractors.} For visual distractors, the first distractor raises image attention, but attention to the image and the overall composition stay flat from $n{=}1$ to $n{=}4$. For textual distractors, the prompt receives less attention despite becoming longer, its concentration relative to a uniform baseline falls monotonically, and the freed attention moves onto the trace. Question-level standard errors of the observed masses range from $0.08$ to $0.25$.}
\label{tab:attn}
\end{table}

Table~\ref{tab:attn} shows the asymmetry the hypothesis predicts. For visual distractors, adding the first distractor raises image attention ($0.96\%$ to $2.04\%$). This jump largely reflects the larger image: the canvas appended for distractors raises the image-token share from $8.8\%$ to $13.3\%$, after which the image size stays fixed. Accordingly, going from one to four distractors produces no further increase: raw image attention stays between $1.65\%$ and $2.04\%$, its ratio to the uniform baseline between $0.12$ and $0.15$, and the image/prompt/trace composition is stable. For textual distractors, each added sentence further dilutes the prompt, with its concentration relative to the baseline falling monotonically from $2.07$ to $1.22$, while attention shifts toward the generated trace. Textual distractors thus keep competing for attention as $n$ grows, whereas visual distractors stop competing after the first.

\paragraph{What the model verbalizes.}
If textual distractors are restated and revisited while visual ones are not, this should be visible in the trace. For each modality we identify sentences of the trace that mention the inserted distractor.

\begin{table}[t]
\centering
\footnotesize
\setlength{\tabcolsep}{4pt}
\resizebox{\columnwidth}{!}{%
\begin{tabular}{@{}lccc@{}}
\toprule
\textbf{Distractor} & \textbf{Mentions} & \textbf{Traces w/} & \textbf{Word} \\
 & \textbf{/ trace} & \textbf{${\geq}1$ mention} & \textbf{share} \\
\midrule
Visual, $n{=}1$ & 4.74 & 24.7\% & 2.51\% \\
Visual, $n{=}4$ & 4.81 & 25.6\% & 2.46\% \\
\midrule
Textual, $n{=}1$ & 48.6 & 94.6\% & 22.8\% \\
Textual, $n{=}4$ & \textbf{105.6} & 97.3\% & \textbf{35.7\%} \\
\bottomrule
\end{tabular}}
\caption{\textbf{Verbalization of distractors.} Visual distractors are mentioned in only a quarter of traces and their footprint is flat in $n$. Textual distractors are mentioned in nearly every trace, and both mentions and word share grow substantially with $n$.}
\label{tab:verbalization}
\end{table}

The contrast is stark (Table~\ref{tab:verbalization}). Visual distractors are only sparsely verbalized: about a quarter of traces mention them at all, and mentions per trace and word share are essentially unchanged from $n{=}1$ to $n{=}4$. Textual distractors are heavily incorporated: nearly every trace mentions them, mentions per trace more than double, and their word share rises from $22.8\%$ to $35.7\%$. This growth accounts for most of the lengthening: mean trace length rises from 2,558 to 3,181 words as $n$ goes from 1 to 4, and 89\% of the increase (553 of 623 words) falls in sentences that mention a distractor,\footnote{This attributes whole sentences that mention a distractor; a stricter phrase-level attribution yields 41\%, so between two fifths and nine tenths of the added length is distractor-related depending on the granularity. Some $n{=}4$ traces are truncated at the 8,192-token budget, so the length increase, and hence this share, is if anything underestimated.} consistent with the added reasoning being spent largely on processing the distractors rather than the problem. The visual pattern also matches the attribute-level analysis, where the total number of attributes per trace is unchanged as distractors are added (Fig.~\ref{fig:sec5-ob1}) and only their composition shifts.

\paragraph{Discussion.}
Taken together, the two analyses point to the same asymmetry. Textual distractors remain accessible in the prompt, are restated into the trace, and are repeatedly revisited, so each additional distractor adds reasoning that is largely spent on the distractor itself. Visual distractors, in contrast, are read once from a fixed set of image tokens and are rarely verbalized, so additional distractors change which attributes are extracted without adding reasoning.
This account also fits two earlier findings. Reasoning length is insensitive to the number of visual distractors but varies substantially with the object class and image resolution (Appendix~\ref{sec:suppl_D}), as expected if visual distractors do not add material to reason about. And typographic distractors behave like visual ones (Fig.~\ref{fig:textual}), suggesting that what matters is whether the content resides in the prompt or in image tokens. The main limitation is that both analyses observe only direct attention and explicit verbalization. Information that has already propagated into intermediate token representations is invisible to either measure. So we cannot rule out that visual distractors also act through such indirect channels, and a constant image-attention share does not imply that no further image information is used. 

\section{Artifact Licenses and Intended Use}
\label{sec:license}

In this section, we provide the licenses for all the pretrained models and datasets utilized in our study. Our use of all artifacts strictly adheres to their original licenses and is entirely consistent with their intended use for academic research and evaluation.

\subsection{Pretrained Models}
\begin{itemize}[noitemsep]
    \item \textbf{Qwen3-VL-8B-Thinking.} Apache License Version 2.0.
    \item \textbf{Qwen3.5-27B.} Apache License Version 2.0.
    \item \textbf{GLM-4.1V-9B-Thinking.} MIT.
    \item \textbf{Intern-S1-mini.} Apache License Version 2.0.
    \item \textbf{R1-OneVision-7B-RL.} Apache License Version 2.0.
    \item \textbf{Gemini 2.5 Flash Image.} Google Gemini API Additional Terms of Service and Google APIs Terms of Service.
    \item \textbf{Sonnet 4.5.} Anthropic Commercial Terms of Service.
\end{itemize}

\subsection{Datasets and Benchmarks}
\begin{itemize}[noitemsep]
\item \textbf{ImageNet-9.} ImageNet Terms of Access.
    \item \textbf{MathVerse.} MIT.
    \item \textbf{LogicVista.} Apache License Version 2.0.
    \item \textbf{Waterbirds.} MIT.
\end{itemize}


\label{sec:appendix}

\end{document}